\documentclass[]{article}
\usepackage{graphicx}
\usepackage{amsmath,amssymb} %
\usepackage{color}
\usepackage{fullpage}
\usepackage{url}
\usepackage{subfigure}
\usepackage[titletoc,title]{appendix}

\def\eg{\textit{e.g.}}
\def\ie{\textit{i.e.}}

\begin{document}

\title{Fully Convolutional Networks for Dense Semantic Labelling of High-Resolution Aerial Imagery}

\author{Jamie Sherrah\\
Defence Science \& Technology Group\\
Edinburgh, South Australia\\
{\tt\small email:  \url{https://au.linkedin.com/jsherrah}}}

\maketitle

\begin{abstract}
  The trend towards higher resolution remote sensing imagery facilitates a
  transition from land-use classification to object-level scene understanding.
  Rather than relying purely on spectral content, appearance-based image
  features come into play.  In this work, deep convolutional neural networks
  (CNNs) are applied to semantic labelling of high-resolution remote sensing
  data.  Recent advances in fully convolutional networks (FCNs) are adapted to
  overhead data and shown to be as effective as in other domains.  A
  full-resolution labelling is inferred using a deep FCN with {\em no
    downsampling}, obviating the need for deconvolution or interpolation.  To
  make better use of image features, a pre-trained CNN is fine-tuned on remote
  sensing data in a hybrid network context, resulting in superior results
  compared to a network trained from scratch.  The proposed approach is applied
  to the problem of labelling high-resolution aerial imagery, where fine
  boundary detail is important.  The dense labelling yields state-of-the-art
  accuracy for the ISPRS Vaihingen and Potsdam benchmark data sets.
\end{abstract}

\section{Introduction}

Land use classification has been a long-standing research problem in remote
sensing, and has historically been applied to coarse resolution multi-spectral
imagery (for example, LANDSAT has 30m x 30m ground sampling distance (GSD),
Quickbird 2.2m GSD).  More recently, high-resolution aerial imagery has become
available with a GSD of 5-10 cm, so that objects such as cars and buildings are
distinguishable.  Pixel labelling becomes a richer semantic representation, but
is more difficult.  Now instead of classifying a spectral signature averaged
over a large area (one pixel to many objects), individual objects must be
segmented (one object to many pixels).  Spectral properties alone may not be
sufficient to distinguish objects (e.g. grass from trees, road from roof), and
discriminative appearance-based features are needed.  The fine-grained
classification of image pixels is termed {\em semantic labelling}.

For such high-resolution imagery, computer vision techniques for object
segmentation and semantic labelling are eminently applicable.  Recently deep
convolutional neural networks (CNNs) have become the dominant method for visual
recognition, achieving state-of-the-art results on a number of problems
including semantic
labelling~\cite{linEtAl:cvpr2016,chenEtAl:iclr2015,longEtAl:CVPR2015}.  CNNs
have also been applied to remote sensing
data~\cite{paisitkriangkraiEtAl:cvprw2015}, but usually on a patch level.  For
classification, this involves classifying a cropped out patch of imagery (\eg
airport, forest).  In the case of semantic labelling, the aim is usually to
classify the pixel at the centre of the patch, and this classification is
applied to overlapping patches densely over the image, resulting in a fully
labelled output.

In this work we apply the recently-developed {\em fully convolutional
  network}~\cite{longEtAl:CVPR2015} to semantic labelling of aerial imagery,
achieving higher accuracy than the patch-based approach.  By exploiting the
convolutional nature of the CNN, the classifier can be treated like a
convolutional filter both during training and classification.  The result is
improved accuracy and computational efficiency.  However the FCN produces output
at a significantly lower resolution than the input imagery due to
pooling-and-downsampling layers in the network.  This is undesirable for
complete labelling of remote sensing data because the fine boundary details are
important.  Here a novel {\em no-downsampling} network is presented to maintain
the full resolution of the imagery at every layer in the FCN.  The
no-downsampling approach boosts the effective training sample size and achieves
higher accuracy than a downsampling FCN, especially when the downsampling factor
of the network is large.

For aerial imagery, a semantic labelling pixel accuracy of around 75\% can be
achieved simply using the spectral and elevation information at each individual
pixel~\cite{paisitkriangkraiEtAl:cvprw2015}.  To achieve higher accuracy on
higher resolution images, the scene appearance must be exploited using
discriminative texture features.  We use pre-trained convolutional features
derived from ImageNet data to improve overhead semantic labelling.  Pre-trained
networks only take 3-band data as input.  To make use of elevation data such as
a digital surface model (DSM) we propose a hybrid network that combines the
pre-trained image features with DSM features that are trained from scratch.  The
hybrid network improves the labelling accuracy on the highest-resolution
imagery.

The novel contributions of this work are:
\begin{enumerate}
\item the use of fully-convolutional networks to semantically label aerial imagery; 
\item a no-downsampling approach to FCNs to preserve output resolution; 
\item a hybrid FCN architecture to combine imagery with elevation data;
\item the first published results on the ISPRS Potsdam benchmark; and
\item state-of-the-art semantic labelling accuracy for high-resolution aerial imagery.
\end{enumerate}

The remainder of the paper is arranged as follows.  Related work on semantic
labelling using CNNs on remote sensing data and FCNs is reviewed in
Section~\ref{sec:related}.
The characteristics of high-resolution aerial imagery are discussed in
Section~\ref{sec:data} and the data sets used for experimentation are
introduced.
Section~\ref{sec:fcn} explains how FCNs are applied to remote sensing data and
compares the approach experimentally to patch-based training.
The no-downsampling FCN is introduced in Section~\ref{sec:downUpSample} and
compared with FCNs.
Section~\ref{sec:pretrain} shows how pre-trained image features can be combined
with a custom FCN with DSM input in a hybrid architecture to improve
segmentation of the very-high-resolution Potsdam data.
The paper concludes with Section~\ref{sec:conclusion}.  Details of the
experimental parameters are listed in the Appendix.

\section{Related Work}\label{sec:related}

There has been a significant amount of past work on classification and
segmentation of remote sensing imagery, for a recent review
see~\cite{Bruzzone2014,Ghamisi2015}.  For classification of very high resolution
imagery (GSD on the order of 10cm or less) the typical approach is to densely
extract features which are passed to a classifier to generate an image of class
labels.  \cite{Porway2008} combined low-level colour and edge features with
object-level features in a hierarchical contextual model for scene parsing of
aerial imagery.  \cite{daiEtAl:2009URSJE} used colour descriptors to label
Ikonos satellite imagery (1m GSD).  \cite{klucknerBischof:isprs2010} performed
unsupervised clustering of super-pixel features and CRF inference to segment
buildings in aerial imagery (10cm GSD).  \cite{Gerke2015Use} applied AdaBoost
classification to super-pixel appearance features for dense labelling of aerial
images (9cm GSD).  \cite{tokarczykEtAl:2015IGRS} used boosting to select
appropriate features from a large pool of local differences of primitives for
4-class classification of aerial (20cm GSD) and satellite images.  More methods
are comprehensively surveyed in the related work section
of~\cite{castelluccioEtAl:2015arxiv}.

Since 2012~\cite{Krizhevsky2012Imagenet} deep convolutional neural networks
(CNNs) have dominated computer vision due to their relatively high accuracy at
image recognition tasks.  Moreover it has been found that when trained on very
large image sets such as the ImageNet challenge, CNNs learn re-usable image
features that are directly applicable to other domains, achieving
state-of-the-art accuracy versus hand-crafted solutions to the same
problem~\cite{Razavian2014CNN}.  CNNs have been previously applied to classification of overhead imagery.
\cite{Mnih2010Learning,Minh2013Machine} trained CNNs to detect roads and buildings from automatically-generated noisy ground truth labels.
\cite{Firat2014} trained convolutional sparse auto-encoders on unlabelled data to detect specific targets such as planes and dry docks.
Both \cite{castelluccioEtAl:2015arxiv} and \cite{marmanisEtAl:grsl2016} performed fine-tuning on pre-trained CNNs to classify overhead image tiles (30cm GSD).
\cite{penattiEtAl:cvprw2015} investigated the generalisation ability of pre-trained CNN features for classification of aerial and satellite image tiles.
\cite{Vakalopoulou2015} used pre-trained CNNs on RGB and CIR satellite images (50cm GSD) to detect buildings.
\cite{paisitkriangkraiEtAl:cvprw2015} trained CNNs from scratch to learn dense features for semantic labelling of aerial CIR imagery and DSM data (9cm GSD).
None of these approaches employed fully-convolutional training which is the main topic of this paper.

In the field of computer vision CNNs have achieved the highest accuracy on
semantic labelling problems.  This has been demonstrated on social media images
in the Pascal VOC challenge~\cite{pascal-voc-2012} and on road scenes in the
Cityscapes benchmark~\cite{Cordts2016Cityscapes}.  More recently
the best results have come from the so-called {\em fully convolutional
  network (FCN) approach}~\cite{longEtAl:CVPR2015}.  An FCN is trained on all
overlapping patches in an image at once by applying it like a convolutional
filter.  Due to the down-sampling in the CNN the output label map has a lower
spatial resolution than the input imagery.  To compensate,
\cite{longEtAl:CVPR2015} used bilinear interpolation of the class probabilities
and a multi-scale fusion approach named {\em deepJet} to restore the output to
full resolution.  
\cite{chenEtAl:iclr2015} used the hole or {\em atrous} method to expand the
support of CNN filters and reduce the down-sampling factor of the network.
Interpolation was still required to upsample by the now-lesser factor.  In this
work we take this approach to its extreme and use the atrous method to avoid
downsampling altogether.
\cite{nohEtAl:iccv2015} trained a deconvolutional network consisting of
deconvolution and unpooling layers to restore the labelling output to full
resolution.  The pooling locations are copied from the corresponding encoding
layers.
State-of-the-art results are achieved by~\cite{linEtAl:cvpr2016} using CNNs in
both the unary and binary terms of a conditional random fields (CRF).  The
binary term network allows patch-to-patch context to be learned by a CNN.  The
output resolution is still degraded and is upsampled using interpolation and a
dense CRF.

\section{Overhead Imagery and Data Sets}\label{sec:data}

The goal of this work is to take semantic labelling methods from computer vision 
and apply them to high-resolution geospatial imagery.  
Processing overhead imagery has some unique challenges compared to the
multimedia images commonly encountered in the computer vision literature.  These
differences are summed up in Table~\ref{tab:mediaAerialDiffs}.  Overhead images
are typically much larger, containing tens to hundreds of megapixels.  In fact
overhead images could be stitched together to cover the entire earth.  The
implication is that multimedia images tend to have a bounded context, both
spatially and in subject matter, whereas overhead images do not.  The simple
approach taken here is to apply processing to a sliding window which provides
spatial context.  In most cases this is a valid approach but in some cases a
much larger spatial context is required for interpretation, for example very
large buildings.  In the case of semantic labelling, both ``things and
stuff''~\cite{heitzKoller:eccv2008} (that is, objects and extended background regions) need to be
identified, in contrast with multimedia data such as Pascal VOC that only
considers ``things'' (objects)~\cite{pascal-voc-2012}.  Regarding invariances, overhead
images do not usually contain significant out-of-plane transformations which
simplifies recognition.  However scenes can be seen from any azimuth so in-plane
rotation invariance is essential, in contrast to multimedia images that usually
assume an upright camera with respect to the ground.  Overhead imagery tends to
be taken at near-nadir angles or ortho-rectified in post-processing so that
occlusion does not usually pose a challenge.

\begin{table*}[htbp]
\centering
\caption{Major differences between ground-based multimedia images and aerial imagery}\label{tab:mediaAerialDiffs}
\begin{tabular}{p{0.475\textwidth}|p{0.475\textwidth}}\hline
Ground-based images                           & Aerial images                                         \\ \hline \hline
Well-defined scene context                    & A potentially-unlimited continuous scene              \\ \hline
Objects are upright                           & Objects can appear at any azimuth (in-plane rotation) \\ \hline
Object scale/distance varies considerably     & Object scale is known                                 \\ \hline
Arbitrary out-of-plane rotation and occlusion & Limited occlusion and close-to-nadir viewpoint        \\ \hline
\end{tabular}
\end{table*}

The overhead imagery used in this work comes from the ISPRS 2D Semantic
Labelling Challenge~\cite{ISPRS} Vaihingen and Potsdam data sets.  Both data
sets consist of near infra-red, red, green ortho-rectified imagery (or {\em
  colour infra-red}, CIR) with corresponding digital surface models (DSMs).  The
Potsdam data set also has the blue channel for imagery.  The DSM is an array the
same size as the input imagery and provides an elevation value at each pixel.
Each training image comes with ground truth labels from the following set:
impervious surface, building, tree, low vegetation, car, unknown.  The task is
to automatically generate labels for the unlabelled test imagery.  Statistics on
the data sets are shown in Table~\ref{tab:dataSets}.  We have also made used of
normalised DSMs made available by the authors of~\cite{Gerke2015Use}.
Contributed results can be seen on the leaderboard at~\cite{ISPRS}.

\begin{table}[htbp]
\centering
\caption{Statistics on the ISPRS 2D Semantic Labelling
Challenge data sets\label{tab:dataSets}}
\begin{tabular}{l|l|l}\hline
Property        & Vaihingen   & Potsdam       \\ \hline \hline
Training images & 16          & 24            \\ \hline
Test images     & 17          & 14            \\ \hline
Total Pixels    & 168,287,871 & 1,368,000,000 \\ \hline
GSD             & 9 cm        & 5 cm          \\ \hline
Bands           & IR,R,G,DSM  & IR,R,G,B,DSM  \\ \hline
\end{tabular}

\end{table}

In experimentation the labelled training images are broken into two subsets,
referred to as the training and validation sets.  The images used for these
subsets are listed in Table~\ref{tab:subTrainVal}.  Experimental
comparisons are based on the validation set results, and the validation set was
not used for training.  In some experiments, results are included for the test
set, which is the unlabelled data.  These accuracies are generated by the
challenge website.

\begin{table*}[htbp] 
\centering 
\caption{Partition of labelled data into training and validation tests\label{tab:subTrainVal}}
\begin{tabular}{l|p{.25\textwidth}|p{.45\textwidth}}\hline
                & Vaihingen & Potsdam \\ \hline \hline
Training set   & 13, 17, 1, 21, 23, 26, 32, 37, 3, 5, 7 & 
2\_10, 3\_10, 3\_11, 3\_12, 4\_11, 4\_12, 5\_10, 5\_12, 6\_10, 6\_11, 6\_12, 6\_8, 6\_9, 7\_11, 7\_12, 7\_7, 7\_9 \\ \hline
Validation set & 11, 15, 28, 30, 34                     & 
2\_11, 2\_12, 4\_10, 5\_11, 6\_7, 7\_10, 7\_8                                                                     \\ \hline
\end{tabular}
\end{table*}

\section{FCN Versus Patch-Based Methods}\label{sec:fcn}

CNNs became popular when applied to individual multimedia images with a defined
spatial context.  When CNNs are applied to remote sensing problems with large
image extent and ambiguous spatial context, the common approach is to crop out
image patches and process these with a pre-trained
CNN~\cite{nogueiraEtAl:arxiv2016,marmanisEtAl:grsl2016,napoletano:arxiv2016}.
The CNN processing is applied in the context of a single image patch, and gives
a single classification for that patch.  To process a whole image it is divided
into patches, which are processed in batches for efficiency, and the CNN outputs
are turned back into an image of now lower resolution than the input.  When
training a patch-based CNN, there are so many candidate patches in the source
image that only a subset is typically used.

Patch-based classification is wasteful, since redundant
operations are performed on neighbouring patches.  It was realised that the CNN
could be applied more efficiently to all overlapping patches by interpreting it as
an image filter: the first layer convolutions are applied to the entire input
image rather than just a patch; the second layer convolutions are applied to the
outputs of the first, and so on resulting in a 2D output matrix of class labels
rather than a single
classification~\cite{youEtAl:bmvc2014,sermanetEtAl:iclr2014}.  The convolutional
approach was taken one step further in~\cite{longEtAl:CVPR2015} where it was
 applied also to training of the CNN, making it fully convolutional.
Now rather than having to select some patches for training, effectively all
overlapping patches are used in a manner that is efficient computationally.
Fully convolutional training is also efficient in terms of GPU RAM, since the
alternative is to load a batch of overlapping patches that share the same
pixels.  These efficiencies mean that FCNs can be trained on more data resulting
in higher accuracy, as was demonstrated on three public data sets
in~\cite{longEtAl:CVPR2015}.  Here we show that these benefits extend to remote
sensing data, and how FCNs can be applied to large images.

\subsection{The CNN as an Image Filter}

\begin{figure}[htbp]
\centering
\subfigure[Example 1-dimensional slice of a CNN classifier with two convolution-and-pooling layers, two fully-connected layers and five output classes.]{\includegraphics[width=0.4\textwidth]{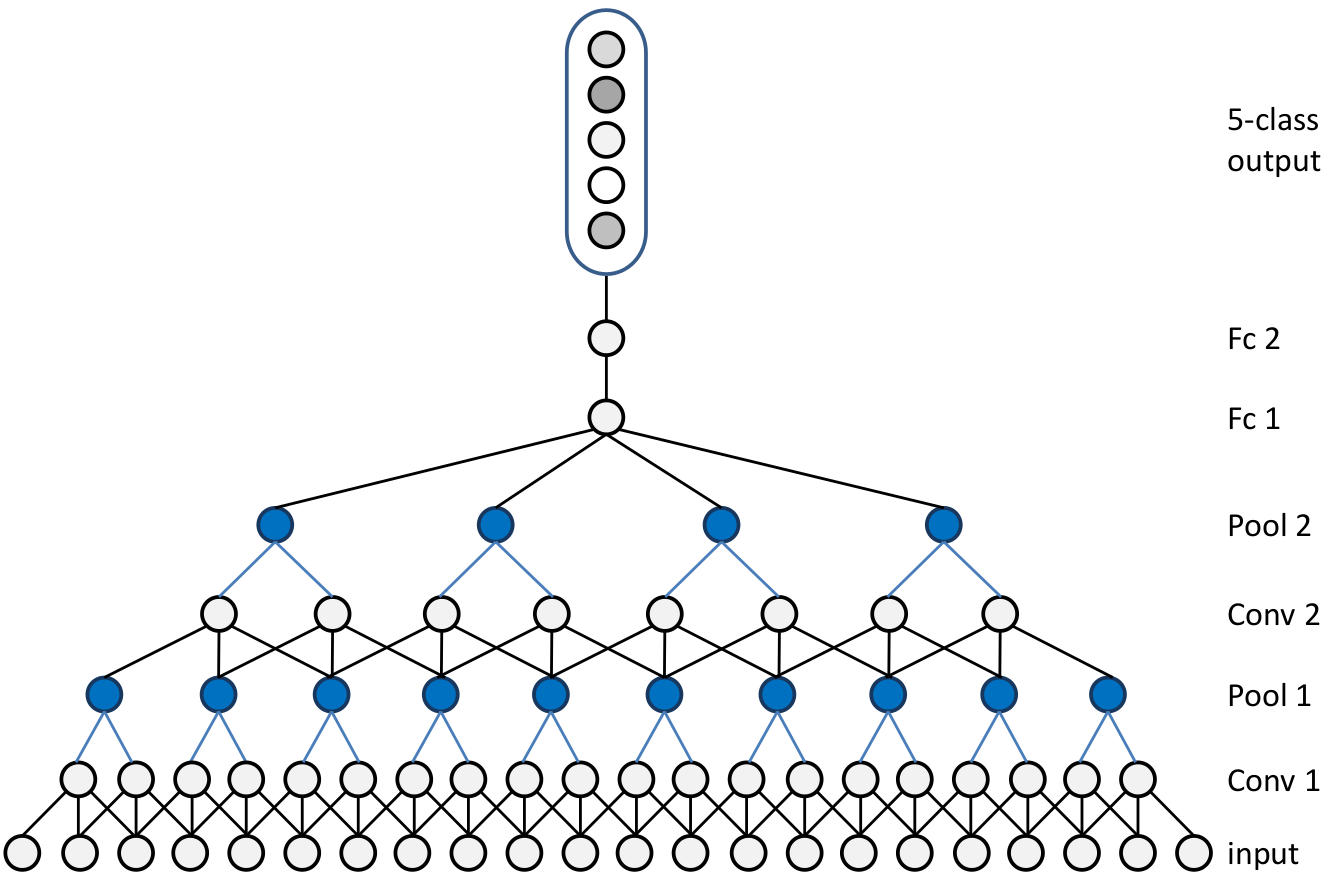}\label{subfig:cnnClassifier}} \hfill
\subfigure[Example of a CNN filter with a downsampling factor of 4, due to two pooling layers.  Therefore the CNN filter stride is $s=4$ pixels.]{\includegraphics[width=0.48\textwidth]{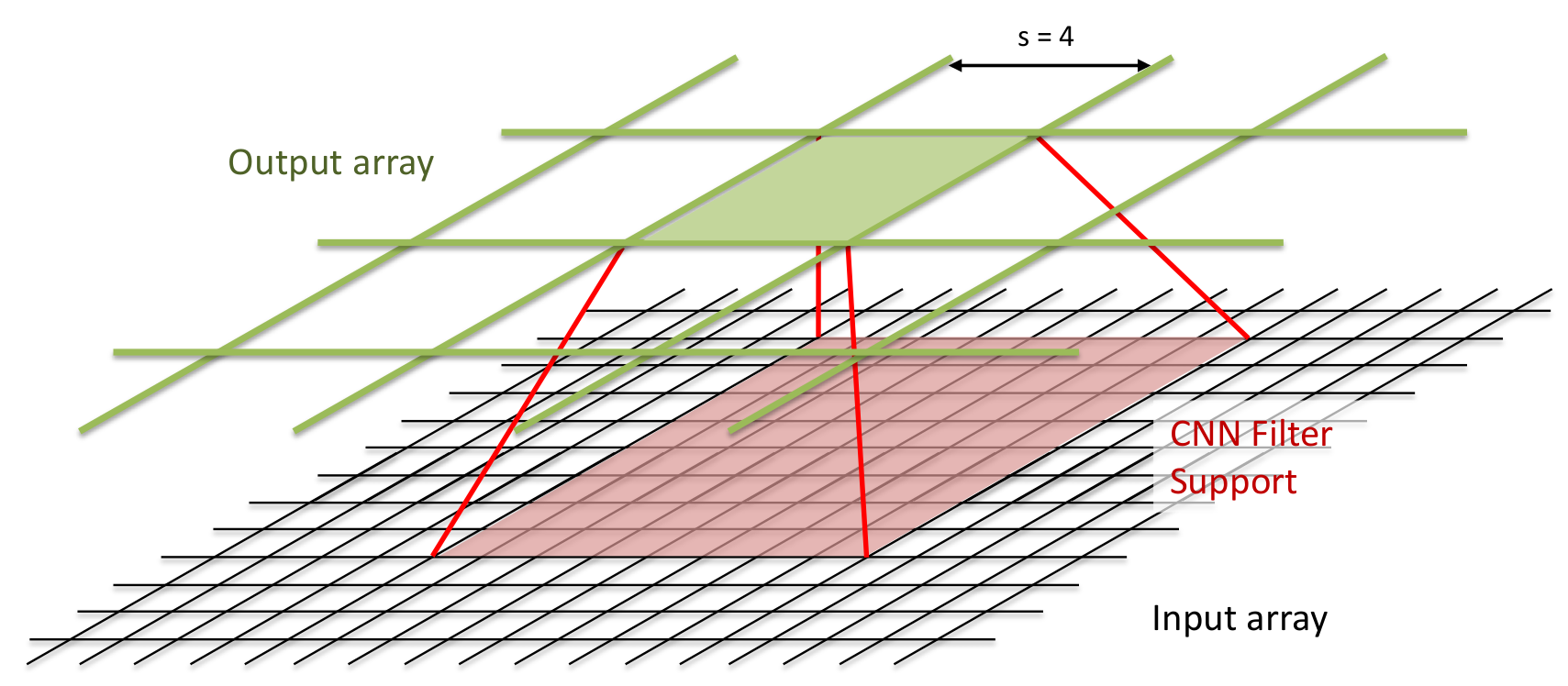}\label{subfig:cnnFilter}} 
\caption{Diagrams showing the difference between a CNN applied as an image classifier and an image filter.\label{fig:cnnClassifierFilter}}
\end{figure}

Initially CNNs were considered as a classifier, generating a single class label
given an input image (for example Figure~\ref{subfig:cnnClassifier}).  For
clarity let us refer to this modus opperandi as a {\em CNN classifier}.  In the
FCN framework the CNN can be interpreted as an image filter, here referred to as
a {\em CNN filter}.  Suppose a CNN classifier is trained on images of fixed size
$n \times n$ pixels.  It can then be applied as a CNN filter on images larger
than $n \times n$, and the spatial support of the CNN filter is $n \times n$.
This is illustrated in Figure~\ref{subfig:cnnFilter}.  To make the CNN a filter,
the fully-connected layers must be turned into convolutional
layers~\cite{longEtAl:CVPR2015}.  If the last convolutional layer of the CNN
classifier has spatial extent $w_c \times h_c$, then the first fully-connected
layer becomes convolutional with filter size $w_c \times h_c$.  All subsequent
fully-connected layers become convolutional with $1 \times 1$ filters.

\begin{figure}[htbp]
\centering
\subfigure[Padding of convolutional layers in a CNN classifier.  Zero padded regions (in green) expand the input array to the next layer.]{\includegraphics[width=0.4\textwidth]{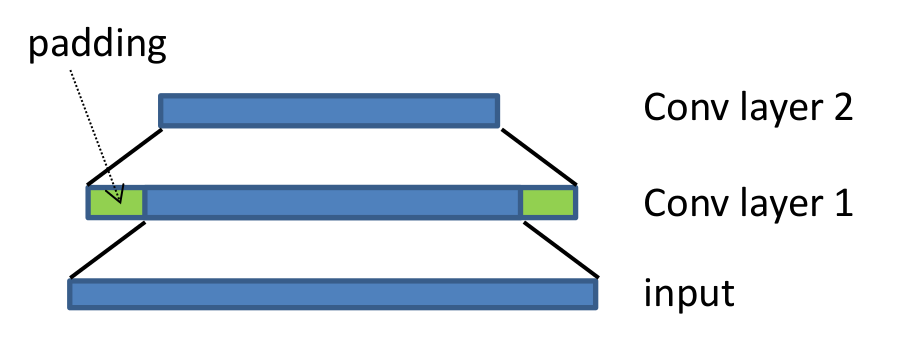}\label{subfig:cnnClassifierPadding}} \hfill
\subfigure[Padding in a CNN filter increases the effective input image size.  Instead of zeros, the padding region now contains features computed from the previous layer. This increases the support of the CNN filter.]{\includegraphics[width=0.48\textwidth]{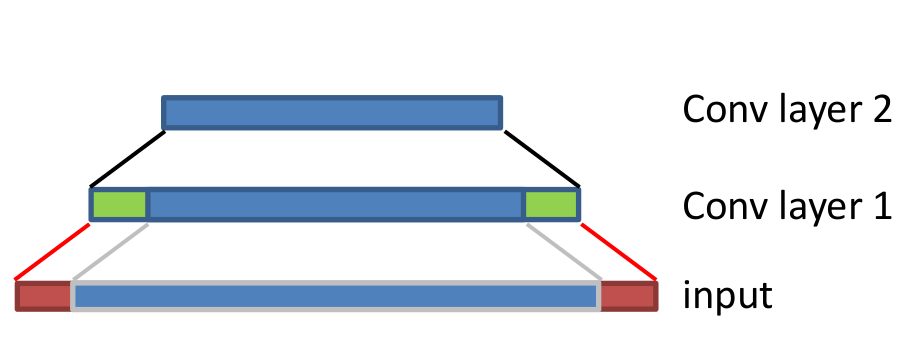}\label{subfig:cnnFilterPadding}} 
\caption{Diagrams showing the effect padding of convolutional layers has on FCNs.\label{fig:cnnClassifierFilterPadding}}
\end{figure}

Using the CNN as a filter introduces design considerations on the network
architecture.  Typically CNN classifiers are designed with zero padding in some
layers in order to arrive at the desired values of $w_c, h_c$ at the last
convolutional layer (see Figure~\ref{subfig:cnnClassifierPadding}).  In the case
of CNN filters, padded areas do not contain zeros (except at the image
boundaries), rather they contain adjacent features.  This could be problematic
if the CNN was trained as a classifier and zeros are expected at the filter
support boundary.  When padding is used in the CNN filter architecture it
increases the effective spatial support of the filter
(Figure~\ref{subfig:cnnFilterPadding}).  Alternatively if the first
fully-connected layer becomes convolutional with a filter smaller than $w_c
\times h_c$, the effective spatial support of the CNN filter is reduced.

In general the output of the CNN filter has a lower spatial resolution than the
input, which is undesirable for applications like semantic labelling.  The
downsampling factor from input to output is the product of the strides from all
layers.  For example if there are 4 max-pooling layers each with a stride of 2
(\ie downsampling factor of 2), the overall downsampling factor of the CNN
filter is $s = 2^4 = 16$.  The downsampling factor $s$ could be viewed as the
stride of the CNN filter, as demonstrated in Figure~\ref{subfig:cnnFilter}.
Methods for restoring the output to the same resolution as the input include
interpolation, shift-and-stitch, and learned multi-resolution deconvolution
(deepJet)~\cite{longEtAl:CVPR2015}.

\subsection{FCN Training on Tiles}

In~\cite{longEtAl:CVPR2015} the FCN was trained on images of modest size using a
minibatch size of 20.  While such multimedia data fits comfortably into GPU
RAM, remote sensing images are typically much larger and the whole image cannot
be accommodated on the GPU.  In our experiments input images and corresponding
target label images are divided into non-overlapping tiles.  During training the
tiles are treated as separate images and processed in minibatches.  During
classification each tile is processed separately to generate a tile of predicted
class labels.  These prediction tiles are mosaiced to form a single output
image.  To avoid artefacts at tile boundaries they are made to overlap by half
the CNN filter support.

FCN training is superior to patch-based training for two reasons.  First, it is
more memory efficient.  The patch-based approach crops out the patches and puts
them in a minibatch.  Since the patches overlap, the minibatch might contain the
same pixel multiple times.  With FCN training each pixel only appears in a
minibatch once.  Second, the FCN effectively trains on more data and so should be
more accurate.  In the patch-based case not all patches are used for training,
rather a random subset is selected.  In FCN training all possible overlapping
patches are used for training.  For a given tile, all overlapping patches form a
minibatch and contribute to the weight update.  The downside is that in natural
imagery neighbouring patches are highly correlated.  In patch-based training the
samples in a minibatch are randomised and the stochasticity of the weight
updates acts as a regulariser.  In contrast, care must be taken with FCN
training because although the tiles in a minibatch can be shuffled, the patches
within the tile cannot and so minibatch samples are not so random.  In our
experiments it was found that smaller batches of tiles produced better results.

\subsection{Experiments} \label{sec:netDesc}

Patch-based training is compared with FCN training on the Vaihingen data set.
In the patch-based case the CNN is trained on patches, but applied
convolutionally at test time.  In both cases the following standard network architecture is:
5x5x32 convolution, Relu, 2x2 max pooling; 5x5x64 convolution, Relu, 2x2 max
pooling; 3x3x96 convolution, Relu, 2x2 max pooling; 3x3x128 convolution, Relu,
2x2 max pooling; fully-connected layer as 3x3x128 convolutional layer, Relu,
dropout; 1x1x128 convolution, Relu, dropout; 1x1x5 softmax output layer.  With 4
pooling layers the downsampling factor is $s=16$.  During training the softmax layer
outputs are upsampled by a factor of 16 to full resolution with bilinear
interpolation.  Further details of the experimental setup are given in
Appendix~\ref{apdx:experiments}.

\begin{figure}[htbp]
\centering \hfill
\includegraphics[width=0.24\textwidth]{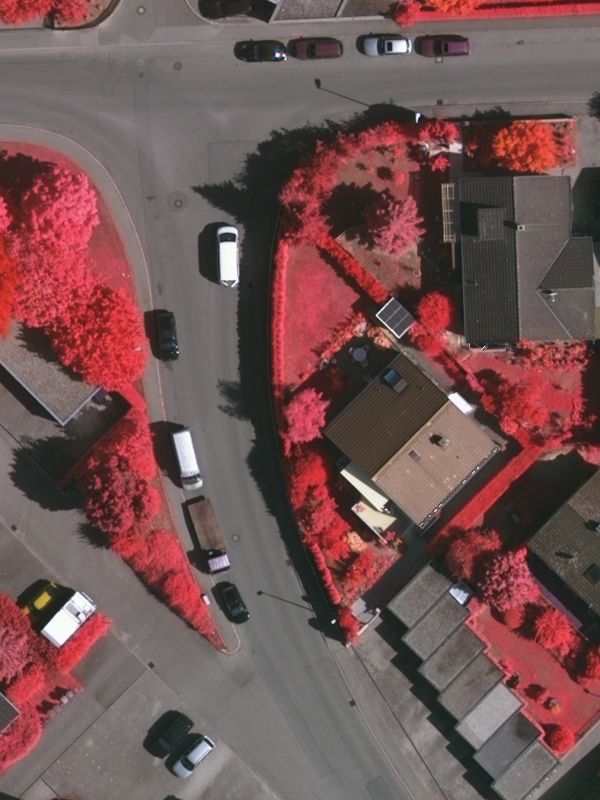} \hfill
   \includegraphics[width=0.24\textwidth]{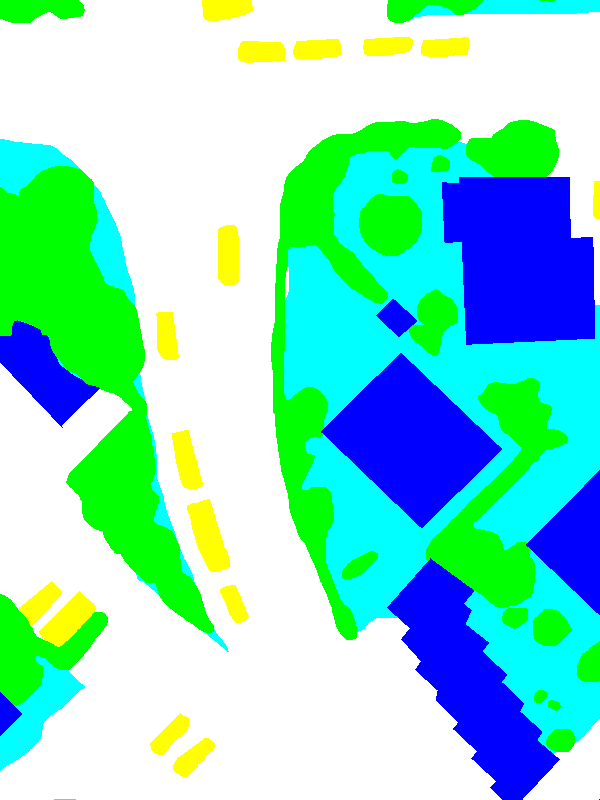} \hfill
\includegraphics[width=0.24\textwidth]{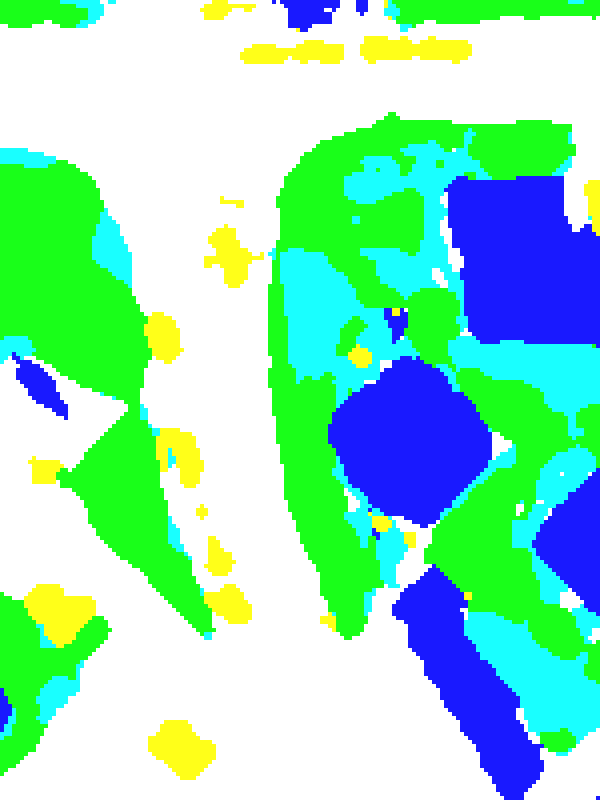} \hfill
  \includegraphics[width=0.24\textwidth]{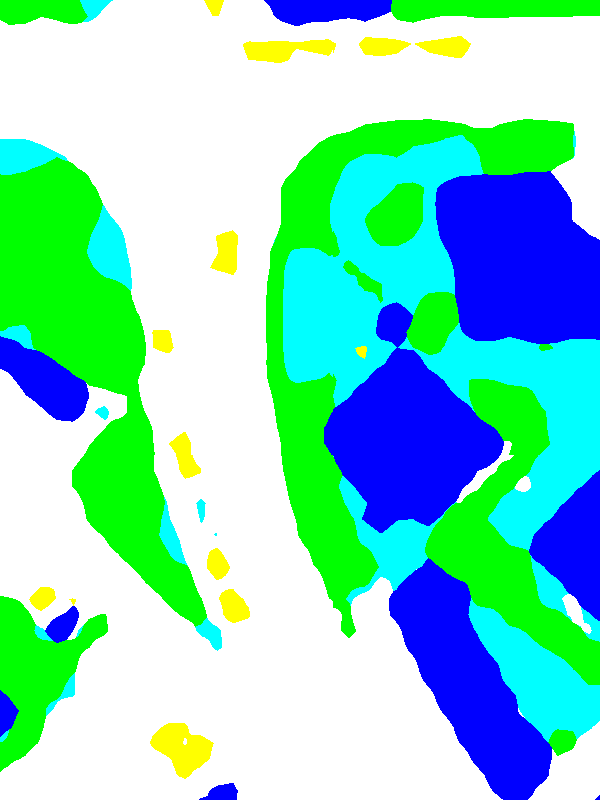} \hfill
\includegraphics[width=0.24\textwidth]{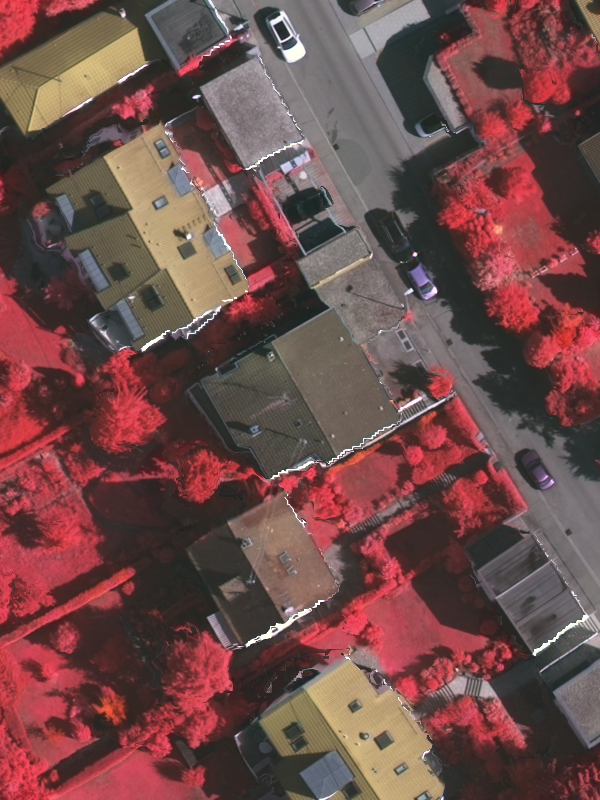} \hfill
\includegraphics[width=0.24\textwidth]{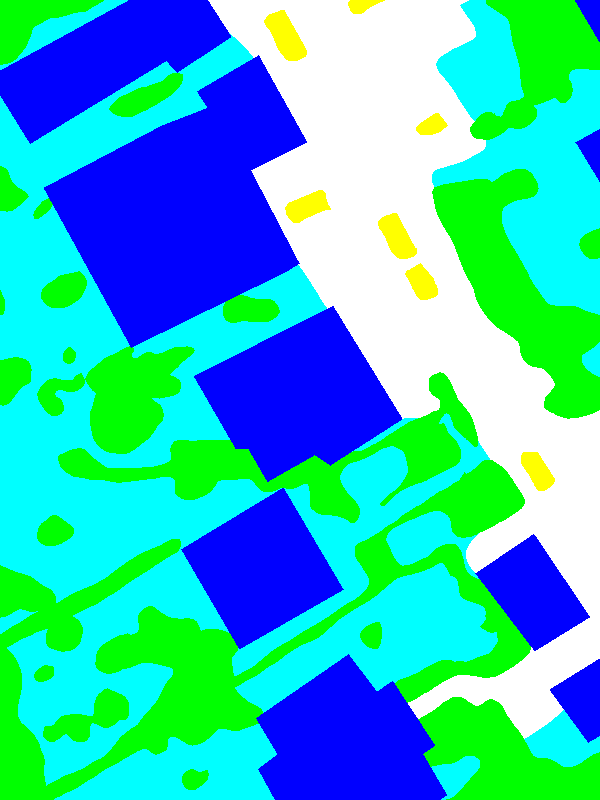} \hfill
\includegraphics[width=0.24\textwidth]{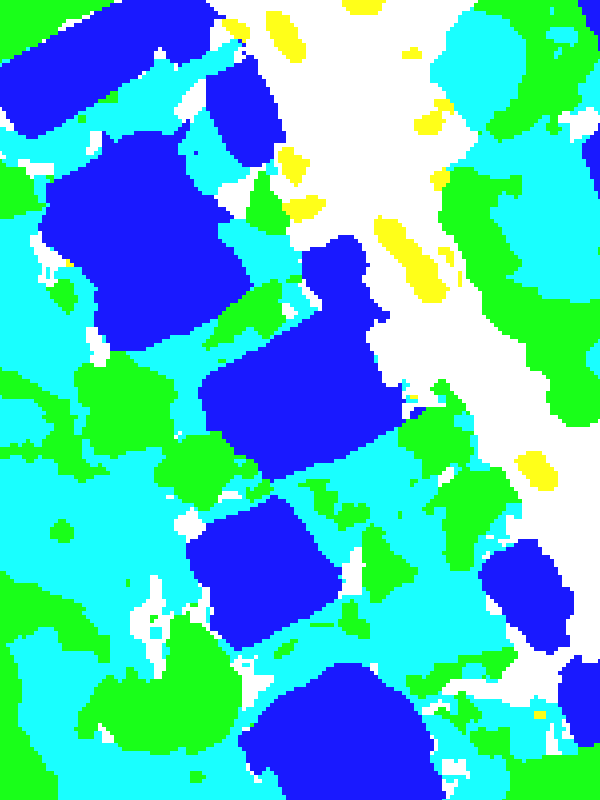} \hfill
\includegraphics[width=0.24\textwidth]{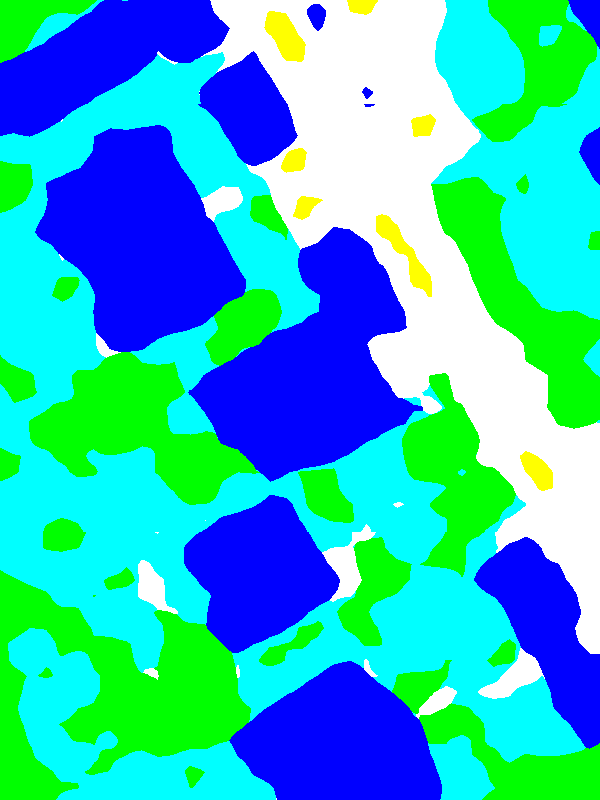} \hfill
\caption{Example semantic labelling output for Vaihingen data set.  From left to
  right: input CIR image, ground truth labels, patch-based training, FCN
  training. Classes are impervious surface (white), building (blue), low
  vegetation (cyan), tree (green), car
  (yellow).  \label{fig:vFcnLabellingResults}}
\end{figure}

The results are shown in Table~\ref{tab:fcnResultsVaihingen}.  The FCN is
trained with two versions of the data, one with only 90 degree augmentations of
the data, and the other with all 36 ten degree rotations.  The 90-degree version
was included to make the results comparable
with~\cite{paisitkriangkraiEtAl:cvprw2015}. The improvement from patch-based to
fully-convolutional training is quite significant, showing the importance of the
increased effective training set size brought about by FCNs.  Examples of
labelling in the two cases are shown in Figure~\ref{fig:vFcnLabellingResults}.
In the third row it can be seen that increasing the amount of rotation
augmentation on the data only produced an incremental improvement to accuracy,
indicating that the original data contains a good diversity of structure
orientations.  Nevertheless the 10-degree rotated data was used in all
subsequent experiments.

\begin{table*} [tb]
\scriptsize
  \centering
  \caption{ FCN semantic labelling results on Vaihingen data set, validation set results. }
  \label{tab:fcnResultsVaihingen}
  {
  \begin{tabular}{p{.28\textwidth}|c|c|c|c|c|c|c}
  \hline
   & \scriptsize{Imp. surf.} & \scriptsize{Building} & \scriptsize{Low veg.} & \scriptsize{Tree} & \scriptsize{Car} & \scriptsize{Overall F1} & \scriptsize{Overall Acc.}\\
  \hline
  \hline  

  Patch-based Training~\cite{paisitkriangkraiEtAl:cvprw2015}
                       & $86.07\%$ & $92.79\%$ & $72.85\%$ & $82.85\%$ & $54.63\%$ & $77.84\%$ & $83.46\%$ \\

  FCN with 4 rotation augmentations  & $89.46\%$  & $93.93\%$  & $76.57\%$  & $86.88\%$  & $67.35\%$  & $82.84\%$  & $86.98\%$ \\

  FCN with 36 rotation augmentations & $89.49\%$  & $94.20\%$  & $77.03\%$  & $86.99\%$  & $66.54\%$  & $82.85\%$  & $\bf 87.17\%$ \\

  \hline
  \end{tabular}
  }
\end{table*}

\section{Downsampling and Upsampling in CNNs}\label{sec:downUpSample}

Overhead images contain both things and stuff (objects and extended regions),
and the accuracy at region boundaries is crucial.  Therefore the output labels
are required at the same spatial resolution as the input image.  As we have seen
an FCN can be viewed as an image filter with a stride or downsampling factor
that is dictated by the number of pooling layers.  The typical solution is to
upsample, either with bilinear interpolation or with learned
deconvolution~\cite{longEtAl:CVPR2015}.  This approach is sub-optimal because
the starting point is a sub-sampled function of the input pixels.  Intuitively
if the FCN is applied to shifted versions of the input, it generates sub-sampled
outputs on intervening points on the pixel grid that can be interleaved to
produce a full-resolution output.  This is the so-called shift-and-stitch
approach.  Here we show that the same result can be achieved more efficiently by
pooling without downsampling.  This approach requires the filter size to
increase with layer depth.  To avoid very large convolutional filters, the
atrous method is used to introduce holes in the filters.  Another way to address
the loss of resolution is to combine features from multiple resolutions, as
in~\cite{linEtAl:cvpr2016} and the deepJet method of~\cite{longEtAl:CVPR2015}.

\subsection{No-Downsampling FCN}

\begin{figure}[htbp]
\centering
\subfigure[Regular downsampling FCN.  The downsampling factor after 2 layers is $s=4$.]{\includegraphics[width=0.75\textwidth]{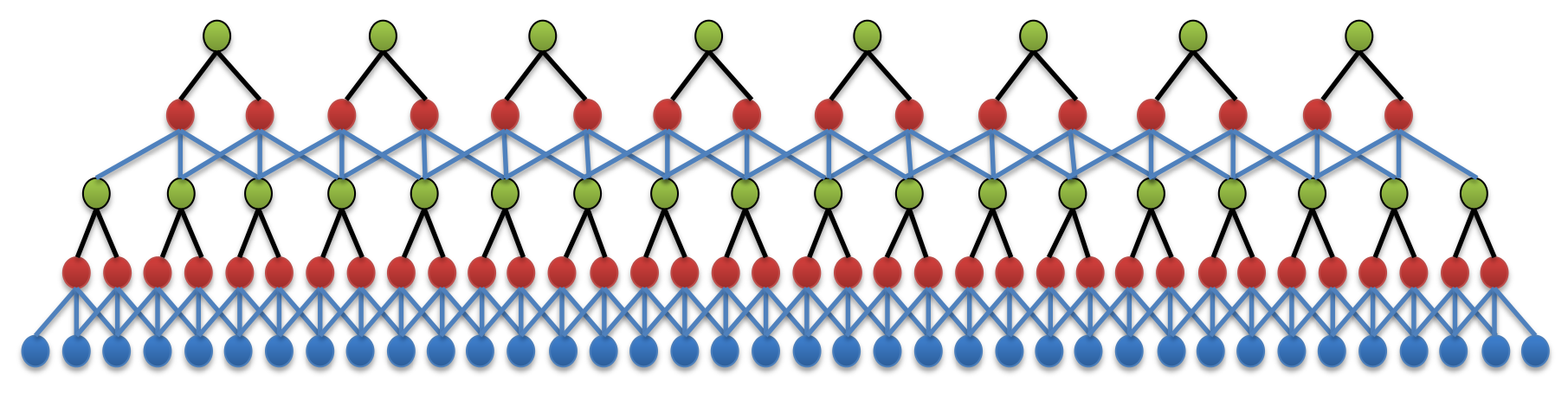}}
\subfigure[A no-downsampling FCN.  The stride at each layer is 1.  The output is equivalent to interleaving the responses to $(s-1)$ shifted versions of the input.]{\includegraphics[width=0.75\textwidth]{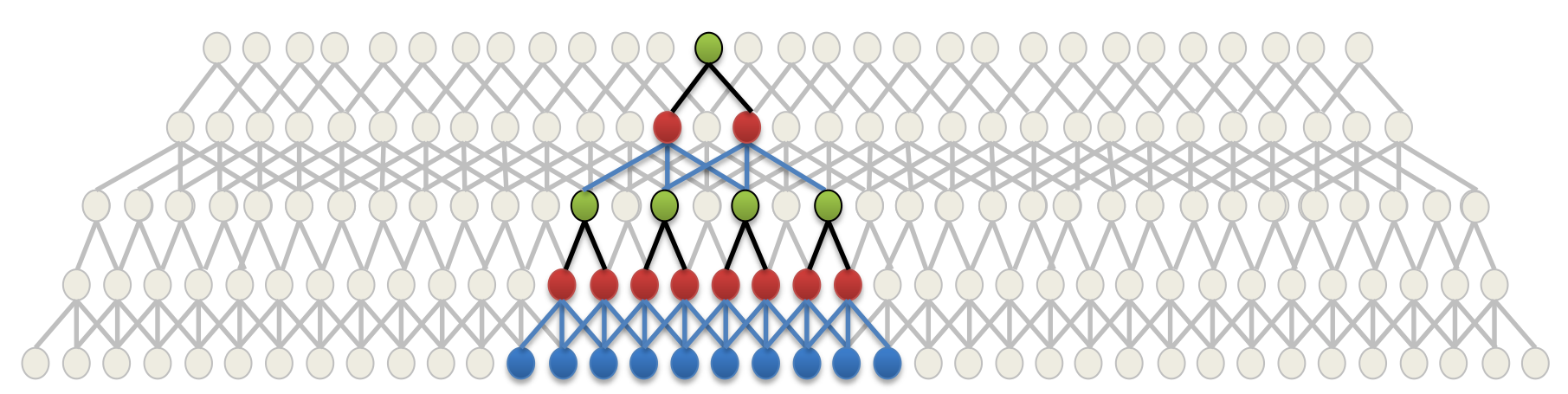}}
\caption{Comparison of convolutional layers with and with out downsampling (1-D
  slice).  In both cases there are two layers of convolution, pooling and $2
  \times$ downsampling.  The bottom-most layer is the input to the CNN.}
\label{fig:fcnNoDownsampleDiags}
\end{figure}

If pooling and downsampling reduces resolution, which is undesirable, then why
downsample at all?  Pooling has been found to be an essential component of CNNs
in summarising feature responses and introducing shift tolerance at multiple
scales~\cite{Krizhevsky2012Imagenet}.  In our proposed approach we retain pooling, but do not
downsample.  As layer depth increases, the size of the convolution and pooling
filters must increase exponentially (see Figure~\ref{fig:fcnNoDownsampleDiags}).  This would increase the
number of parameters in the model severely and cause over-fitting.  As suggested
in~\cite{chenEtAl:iclr2015} the atrous method is used to interleave the convolutional filter
coefficients with zeros so that the number of weights remains the same but the
spatial support increases.  For the pooling layers holes are not introduced.
The architecture of the no-downsampling network used in our experiments is shown
in Table~\ref{tab:noDownsampleArch}.

\begin{table}[htbp]
\centering
\caption{Architecture of the no-downsampling FCN.  In all layers the filtering stride is 1 pixel.}
\label{tab:noDownsampleArch}
{
\begin{tabular}{l|l|l|l|l}
\hline
layer & filter size   & number of filters & filter dilation & padding \\ \hline \hline
conv1 & $5 \times 5$  & 32                & 1               & 2       \\ 
pool1 & $3 \times 3$  & -                 & -               & 1       \\ 
&&&&\\                
conv2 & $5 \times 5$  & 96                & 2               & 4       \\ 
pool2 & $5 \times 5$  & -                 & -               & 2       \\ 
&&&&\\                
conv3 & $3 \times 3$  & 128               & 4               & 4       \\ 
pool3 & $9 \times 9$  & -                 & -               & 4       \\ 
&&&&\\                
conv4 & $3 \times 3$  & 196               & 8               & 8       \\ 
pool4 & $17 \times 17$& -                 & -               & 8       \\ 
&&&&\\
fc5   & $3 \times 3$  & 1024              & 16              & 16       \\ 
fc6   & $1 \times 1$  & 1024              & -               & -        \\ 
softmax
      & $1 \times 1$  & 5                 & -               & -        \\  \hline
\end{tabular}
}
\end{table}

At test time, the no-downsampling FCN is equivalent to a downsampling FCN
applied with shift-and-stitch, but more computationally efficient since the
convolution at each layer is performed only once.  In contrast, the downsampling
FCN would have to be applied to the image $s^2$ times using the shift-and-stitch
approach.  At training time, no-downsampling increases the effective training
set size, because all shifted-and-interleaved versions of the input image are
used in backpropagation.  However this comes at a computational cost: now the
outputs at each intermediate layer of the network are the same spatial
resolution as the input image.  Training requires much more GPU memory and
computation time.  The effective increase in training sample size is expected to
be of modest benefit because shifted versions of the input are highly correlated
with one another.

\subsection{Testing with No Downsampling}

An interesting aspect of the no-downsampling approach is that a regular
downsampling FCN can be turned into a no-downsampling FCN at test time.  The
output is equivalent to the shift-and-stitch approach, where shifted versions of
the input are processed and the results interleaved into a full resolution
output.  The advantage is that computation is much more efficient.  Suppose
there are $L$ repetitions of convolution, pooling and $2\times$ downsampling layers in a
CNN.  The stride of the CNN filter is $s = 2^L$.  Shift-and-stitch would require
processing the input $s^2$ times.  To give an example, the popular VGG-16
pre-trained network~\cite{simonyan2014very} has $s=32$, requiring the input to be processed
1024 times!  In contrast, the no-downsampling network processes the input image
only once.  

The shift-and-stitch approach uses a downsampling FCN, so there are fewer
computations per layer.  However this does not compensate for the $s^2$ factor
described above.  For example in the first layer, the whole input image must be
convolved $s^2$ times in comparison to once for the no-downsampling FCN.  Let us
estimate the reduction in the number of computations required for the
convolutional filters of the no-downsampling FCN compared with shift-and-stitch.
For simplicity, we assume all convolutional layers have a downsampling factor of
2 in the traditional architecture.  Define the following:
\begin{eqnarray*}
L         & : &  \mbox{ number of convolutional layers              }  \\
W, H      & : &  \mbox{ dimensions of input image                   }  \\
w_l, h_l  & : &  \mbox{ size of convolutional filters in layer} \  l      \\
n_l       & : &  \mbox{ number of convolutional filters in layer} \  l    \\
\lambda_0 & : &  \mbox{ number of computations for shift-and-stitch }  \\
\lambda   & : &  \mbox{ number of computations for no-downsampling  }  \\
\end{eqnarray*}

\begin{eqnarray*}
\lambda_0 &=& 2^{2L}  \sum_{l=1}^L 2^{-2(l-1)} W. H. w_l. h_l. n_l. n_{l-1} \\
          &=& W. H    \sum_{l=1}^L 4^{L-l+1}       w_l. h_l. n_l. n_{l-1} \\
\end{eqnarray*}

\begin{eqnarray*}
\lambda   &=& W. H    \sum_{l=1}^L               w_l. h_l. n_l. n_{l-1} \\
\end{eqnarray*}

The speed-up ratio is:

\begin{equation}
\eta = \frac{\sum_{l=1}^L 4^{L-l+1} w_l. h_l. n_l. n_{l-1} }
{\sum_{l=1}^L w_l. h_l. n_l. n_{l-1}}
\end{equation}

Note that although the filters have larger spatial support in higher layers of
the no-downsampling FCN, the computational cost is the same due to the holes
placed in the filter.  Here the computational cost of the pooling operation is
ignored because max pooling can be performed in $\mathcal{O}(W\times H)$
independent of the filter size.%

$\eta$ was computed for some popular pre-trained CNNs.  For AlexNet~\cite{Krizhevsky2012Imagenet},
$\eta = 73.24$.  For VGG-16~\cite{simonyan2014very}, $\eta = 21.29$.  Note that these figures are
theoretical and  deviate from actual values, due to the constant
computational costs incurred by GPUs and the pooling layers.

\subsection{Experiments}

Here the issue of output resolution is addressed experimentally.

\paragraph{Training with No Downsampling}

The effect of training with no-downsampling is investigated.  The same
network architecture was used but adapted to pool but not downsample, so that
the output is the same resolution as the input.  This network is compared to the
pool-and-downsample network on the Vaihingen data set, which uses bilinear interpolation
to restore the output to full resolution.  The results are shown in
Table~\ref{tab:trainNoDsResults}.  The overall improvement by training with
no-downsampling is 0.53\%, but the improvement for the car class is 10.23\%.
No-downsample training effectively produces finer sampling of the training
patches.  Since cars constitute about 1\% of the data, this class benefits
greatly from increased representation in the training sample.

\begin{figure}[htbp]
\centering \hfill
\includegraphics[width=0.24\textwidth]          {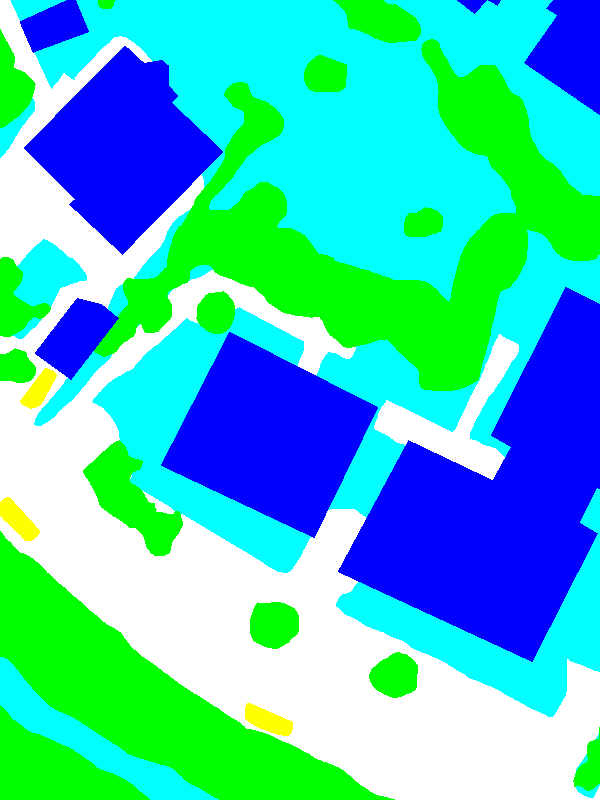} \hfill
\includegraphics[width=0.24\textwidth]         {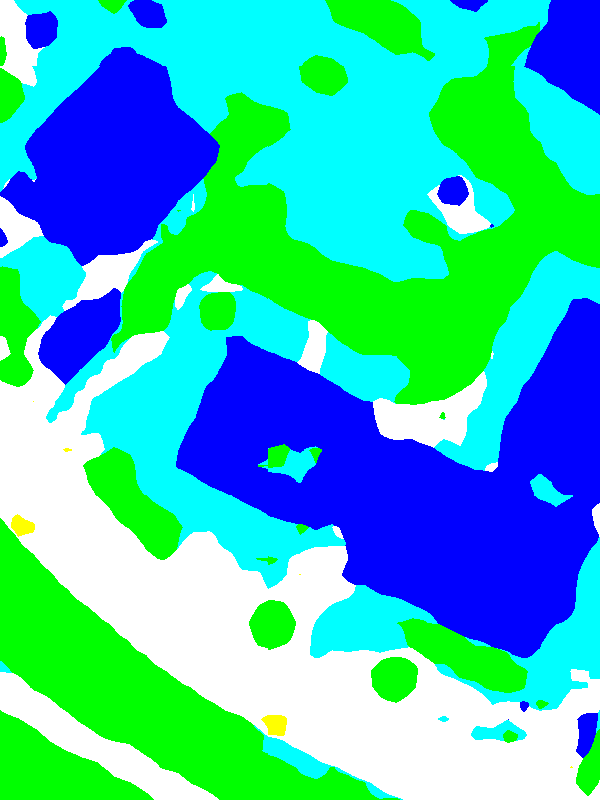} \hfill
\includegraphics[width=0.24\textwidth]     {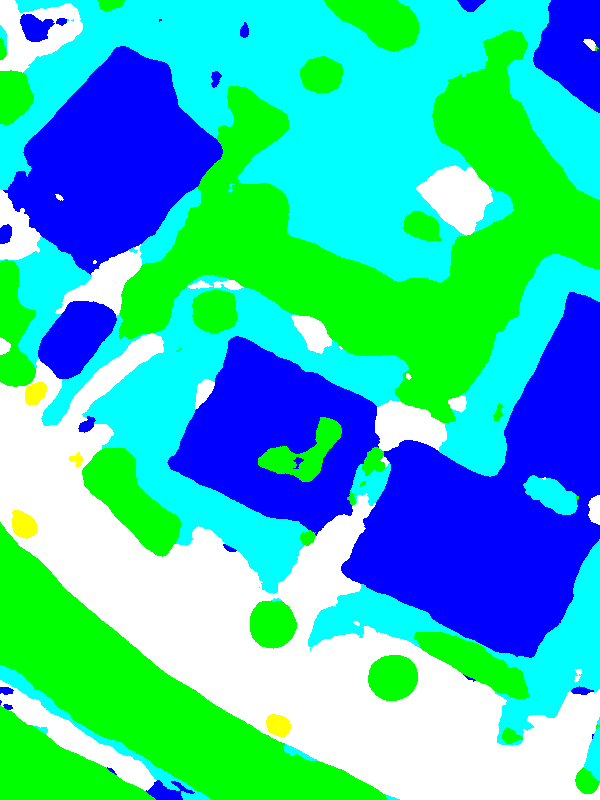} \hfill
\includegraphics[width=0.24\textwidth]{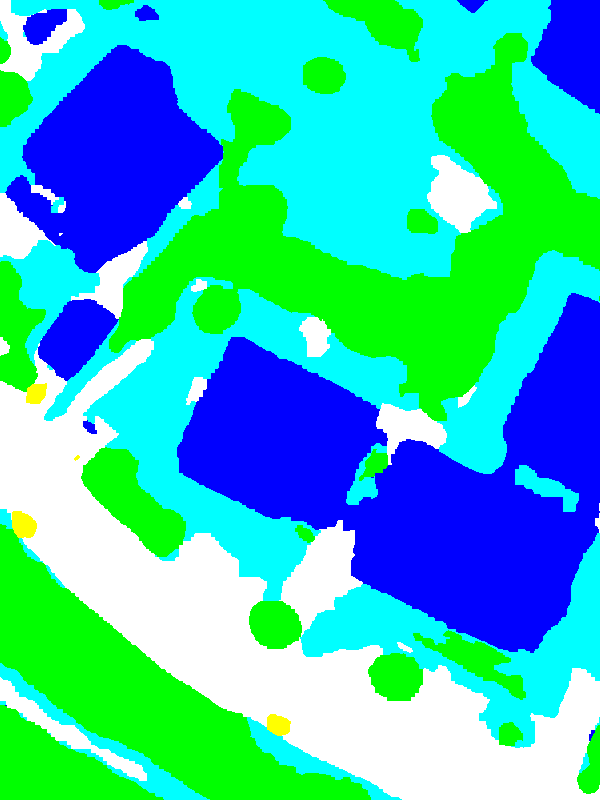} \hfill
\includegraphics[width=0.24\textwidth]          {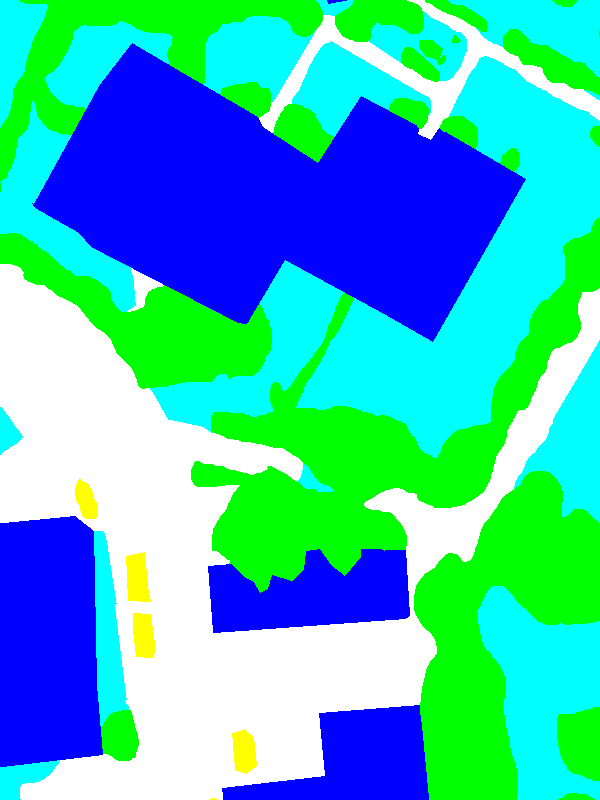} \hfill
\includegraphics[width=0.24\textwidth]         {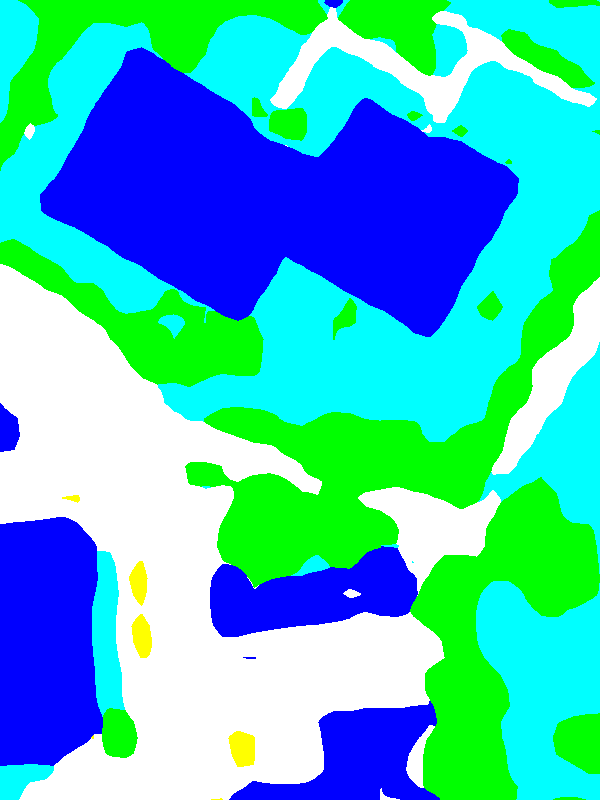} \hfill
\includegraphics[width=0.24\textwidth]     {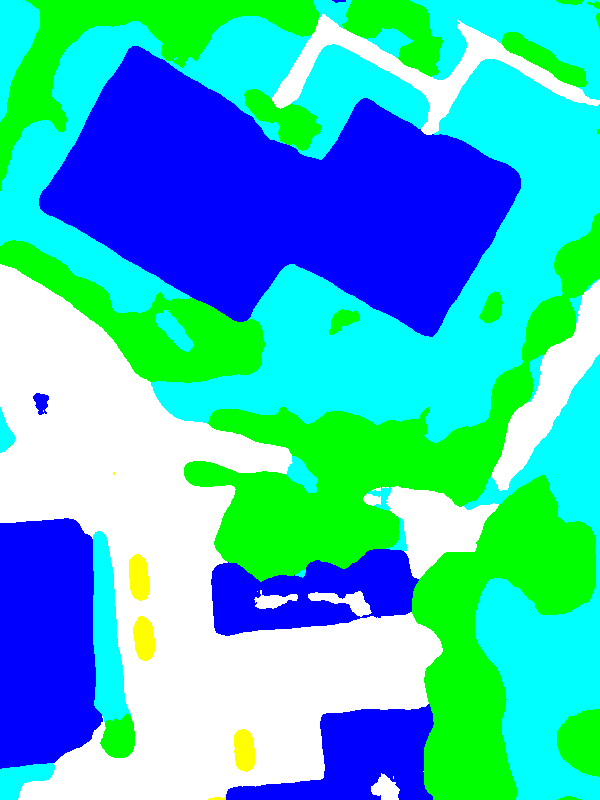} \hfill
\includegraphics[width=0.24\textwidth]{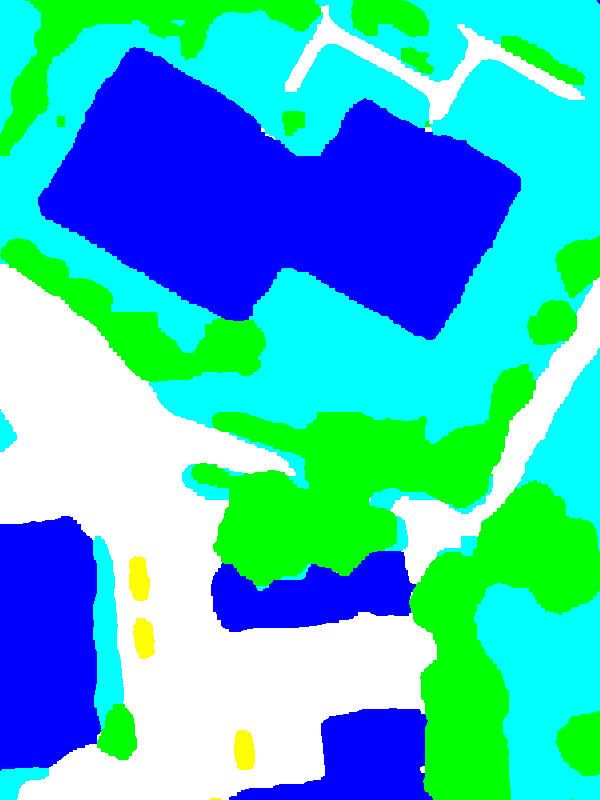} \hfill
\caption{Example semantic labelling output for Vaihingen data set using networks
  trained with no-downsampling.  From left to right: %
  ground truth labels, downsampling FCN, no-downsampling FCN, no-downsampling
  FCN with CRF. \label{fig:vNoDsResults}}
\end{figure}

In~\cite{paisitkriangkraiEtAl:cvprw2015} the labelling results from a patch-based
CNN were combined with hand-crafted features and the results smoothed with a
CRF, resulting in a 1.9\% accuracy improvement on the Vaihingen data.  We added
this strategy to the no-downsampling FCN, the results are found in
Table~\ref{tab:trainNoDsResults}.  These complimentary components only improve
the accuracy by 0.2\%.  The improved smoothness and increased effective training
size of the no-downsampling FCN has already exploited the benefits that were
previously afforded by the hand-crafted features and CRF.  Example outputs for
these networks are shown in Figure~\ref{fig:vNoDsResults}.

\begin{table*} [tb]
\scriptsize
  \centering
  \caption{ FCN semantic labelling results on the Vaihingen validation set, trained with no-downsampling.  The configurations marked ``DST\_X'' correspond to the leaderboard submissions, see Table~\ref{tab:vaihingenLeaderbordNoDS} for the hold-out test set results.}
  \label{tab:trainNoDsResults}
  {
  \begin{tabular}{p{.36\textwidth}|c|c|c|c|c|c|c}
  \hline
   & \scriptsize{Imp. surf.} & \scriptsize{Building} & \scriptsize{Low veg.} & \scriptsize{Tree} & \scriptsize{Car} & \scriptsize{Overall} & \scriptsize{Overall}\\
   &&&&&& \scriptsize{ F1} & \scriptsize{ Acc.}\\
  \hline
  \hline  

  FCN trained with downsampling   & $89.49\%$  & $94.20\%$  & $77.03\%$  & $86.99\%$  & $66.54\%$  & $82.85\%$  & $87.17\%$ \\

  FCN trained with no-downsampling (DST\_1) & $90.19\%$  & $94.49\%$  & $77.69\%$  & $87.24\%$  & $76.77\%$  & $85.28\%$  & $87.70\%$ \\

  FCN trained with no-DS + RF + CRF (DST\_2)    & $90.41\%$  & $94.73\%$  & $78.25\%$  & $87.25\%$  & $75.57\%$  & $85.24\%$  & $\bf 87.90\%$ \\

  \hline
  \end{tabular}
  }
\end{table*}

\begin{figure}[htbp]
\centering \hfill
  \includegraphics[width=0.24\textwidth]{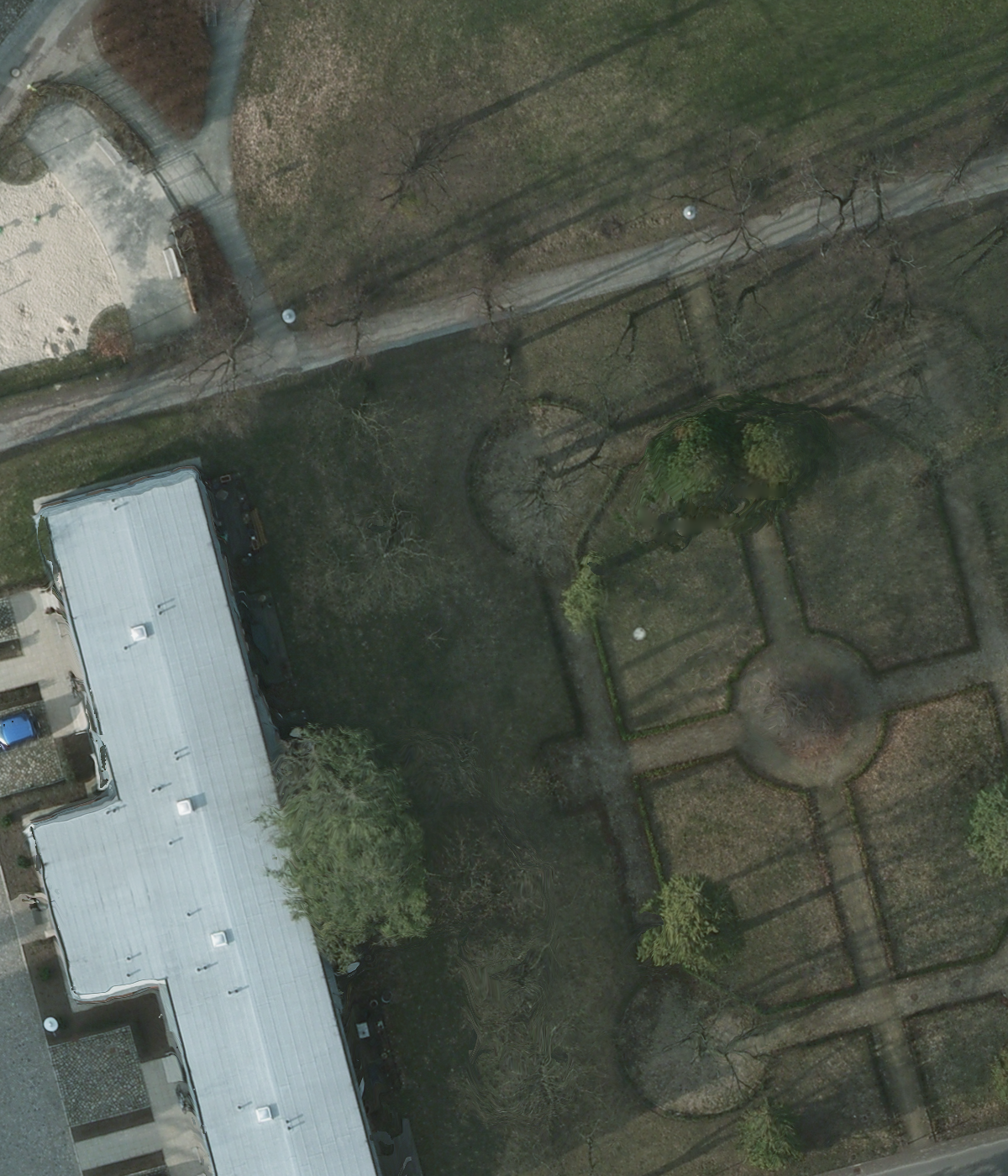} \hfill
     \includegraphics[width=0.24\textwidth]{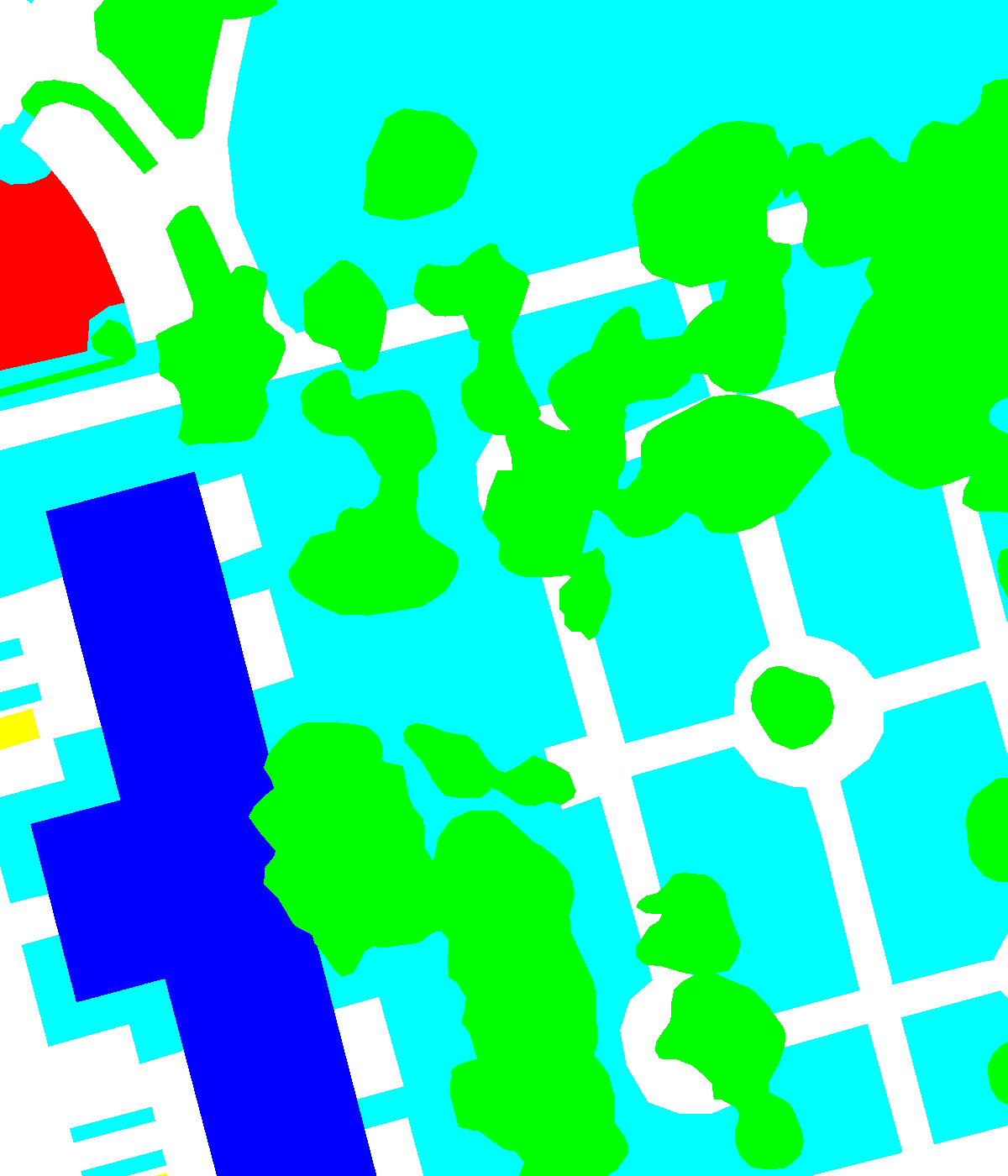} \hfill
    \includegraphics[width=0.24\textwidth]{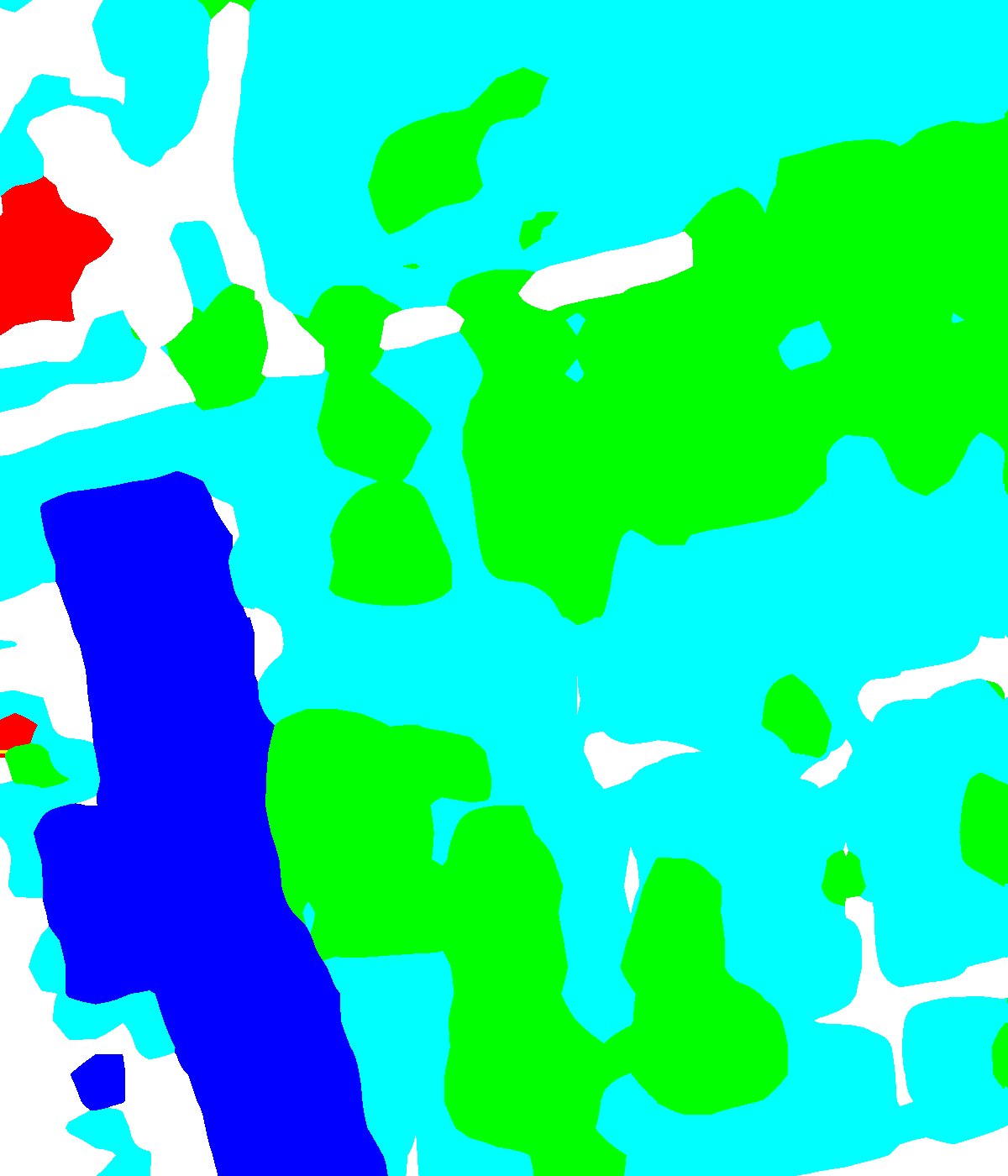} \hfill
\includegraphics[width=0.24\textwidth]{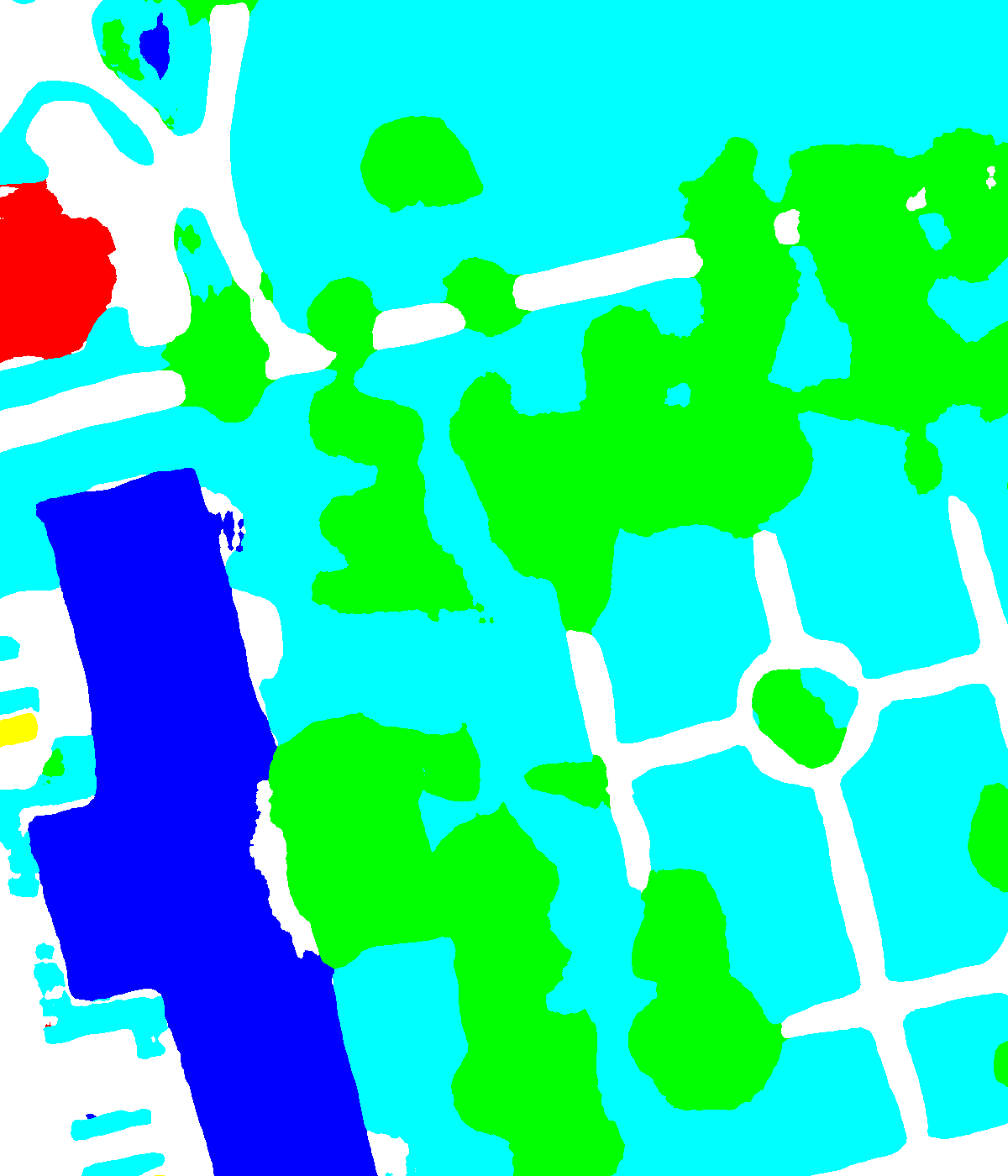} \hfill
  \includegraphics[width=0.24\textwidth]{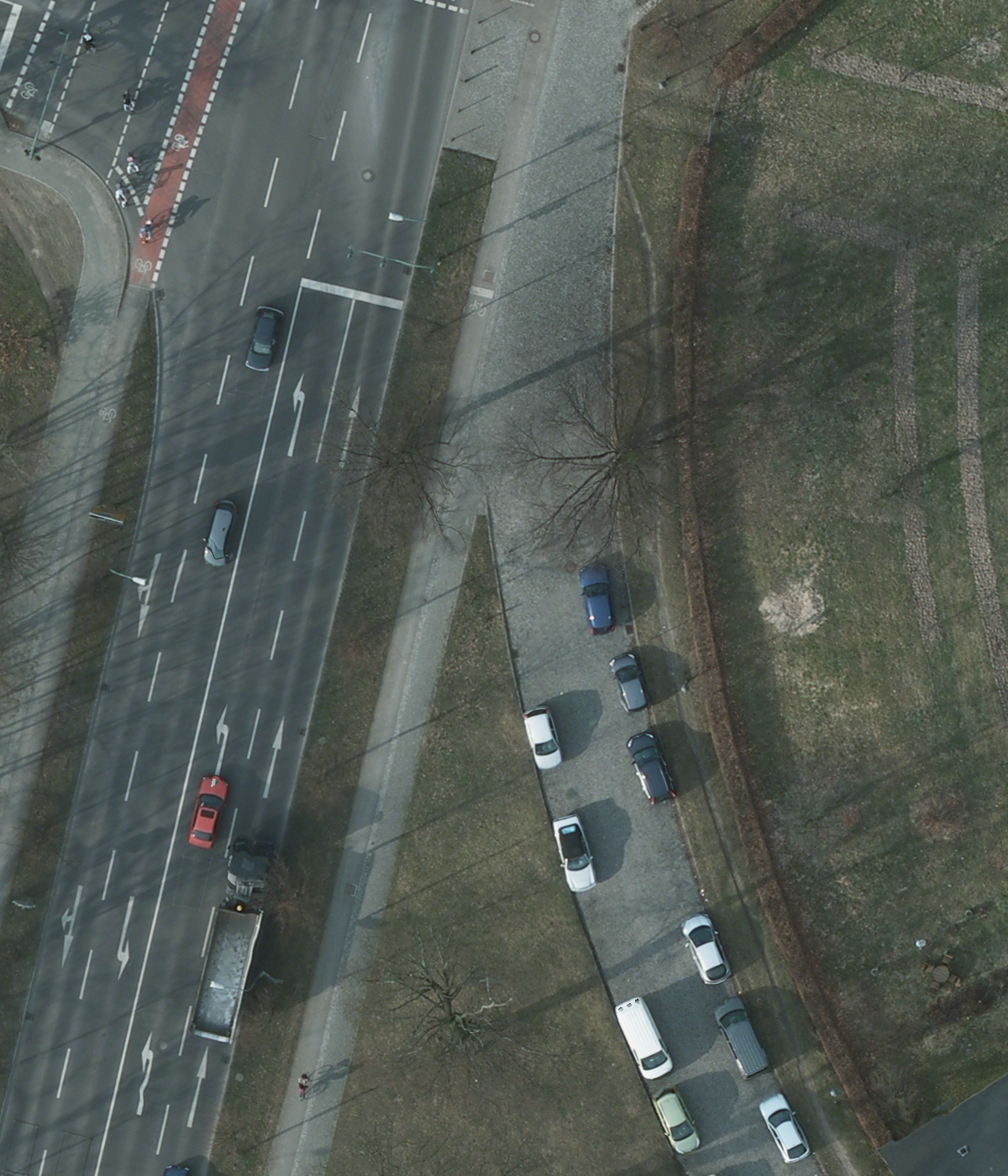} \hfill
     \includegraphics[width=0.24\textwidth]{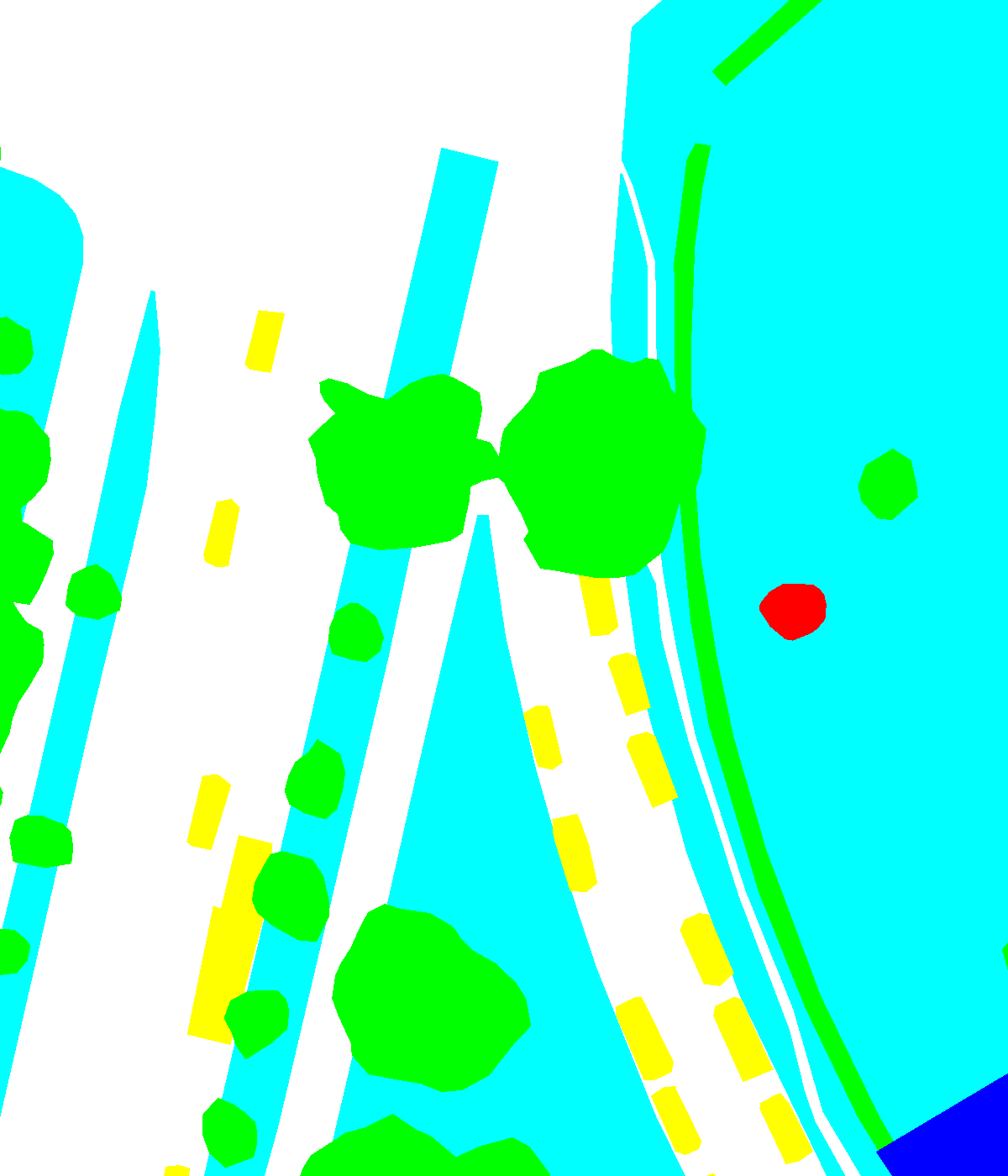} \hfill
    \includegraphics[width=0.24\textwidth]{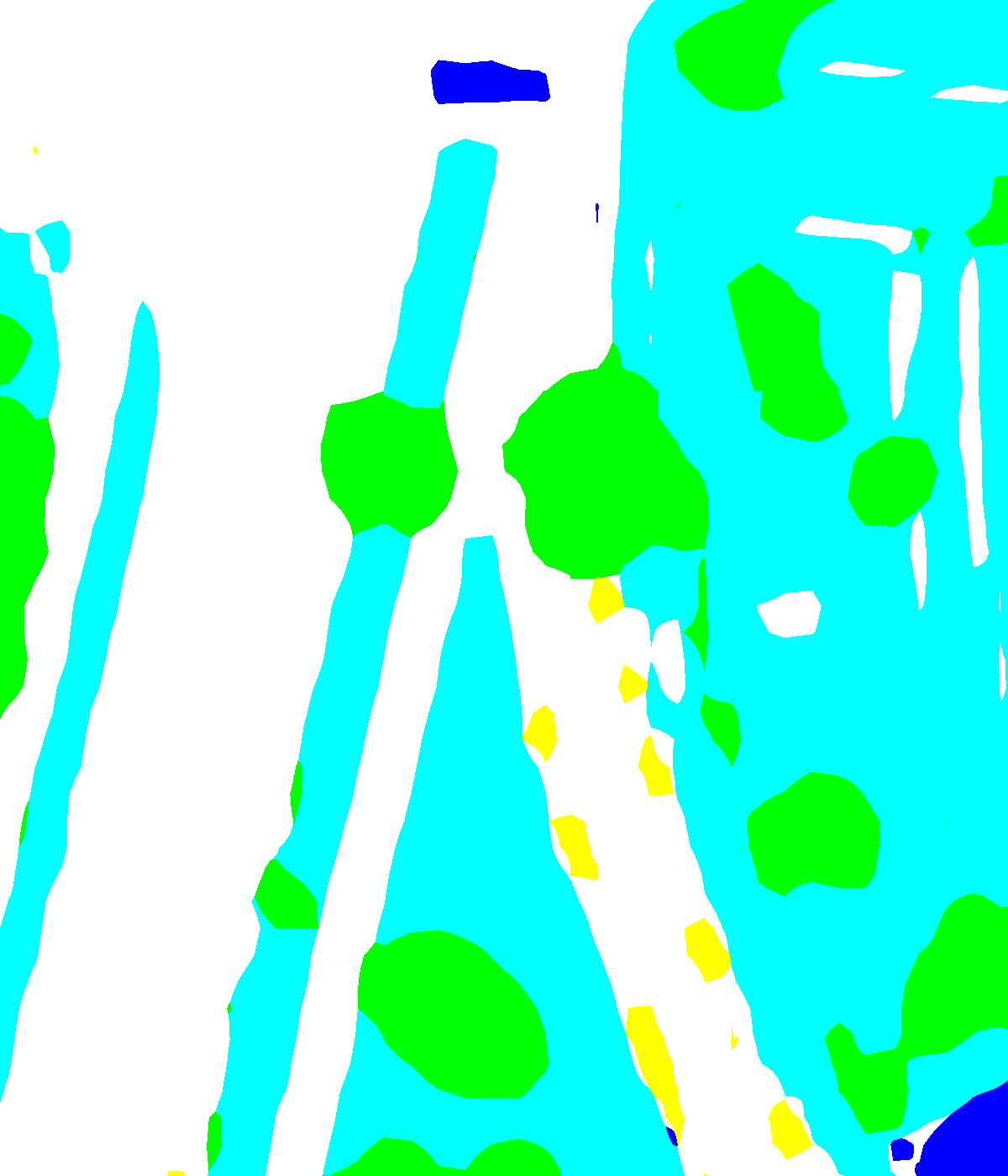} \hfill
\includegraphics[width=0.24\textwidth]{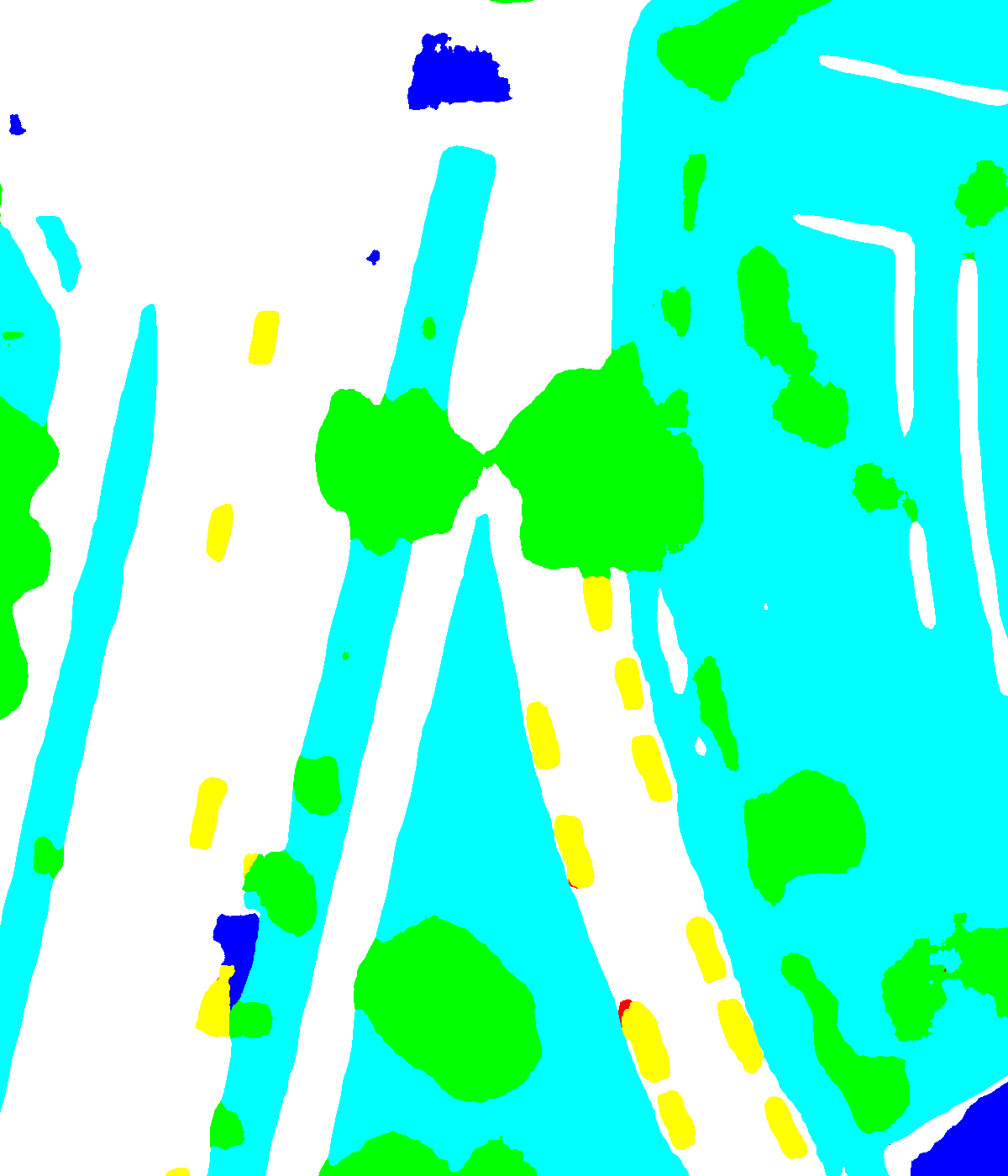} \hfill
  \includegraphics[width=0.24\textwidth]{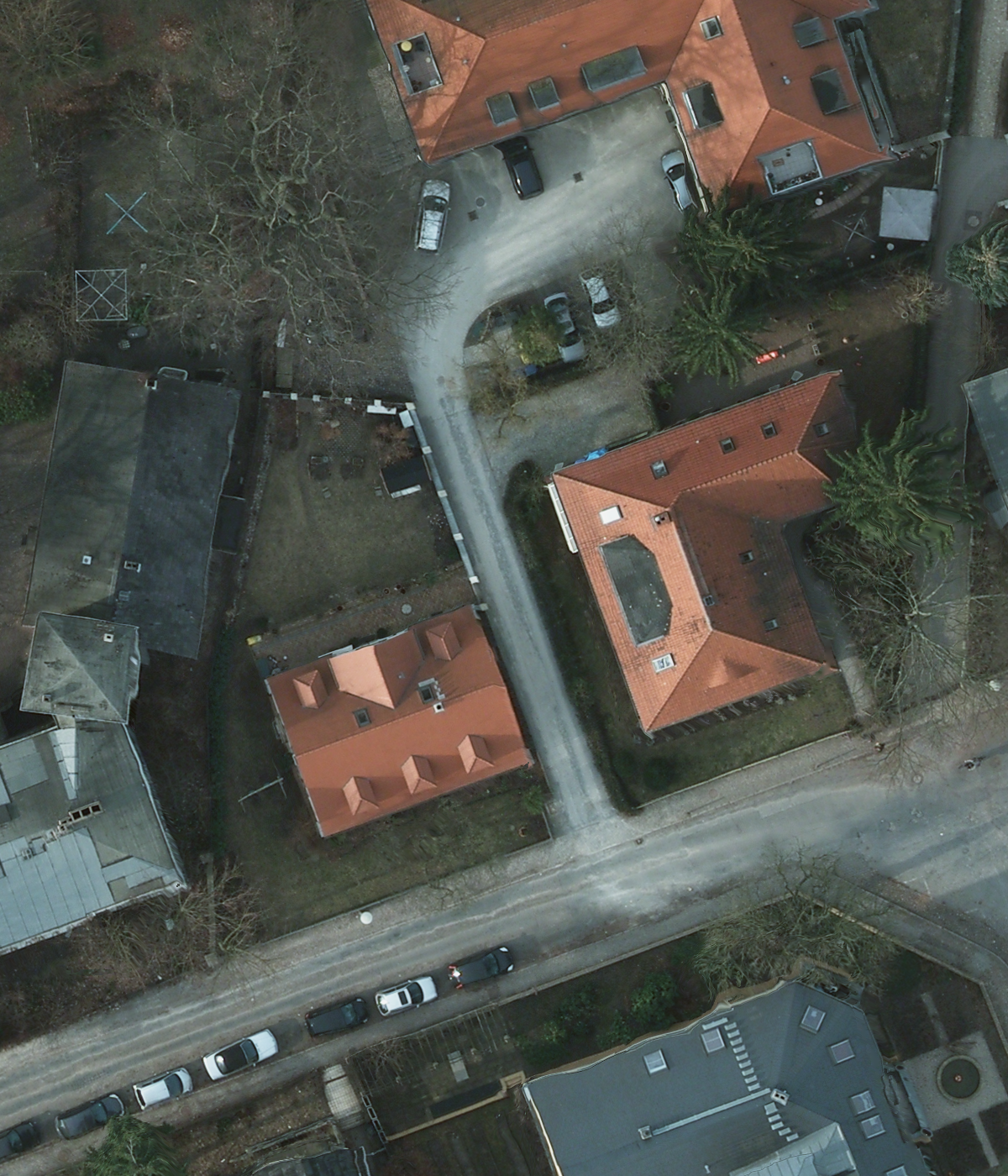} \hfill
     \includegraphics[width=0.24\textwidth]{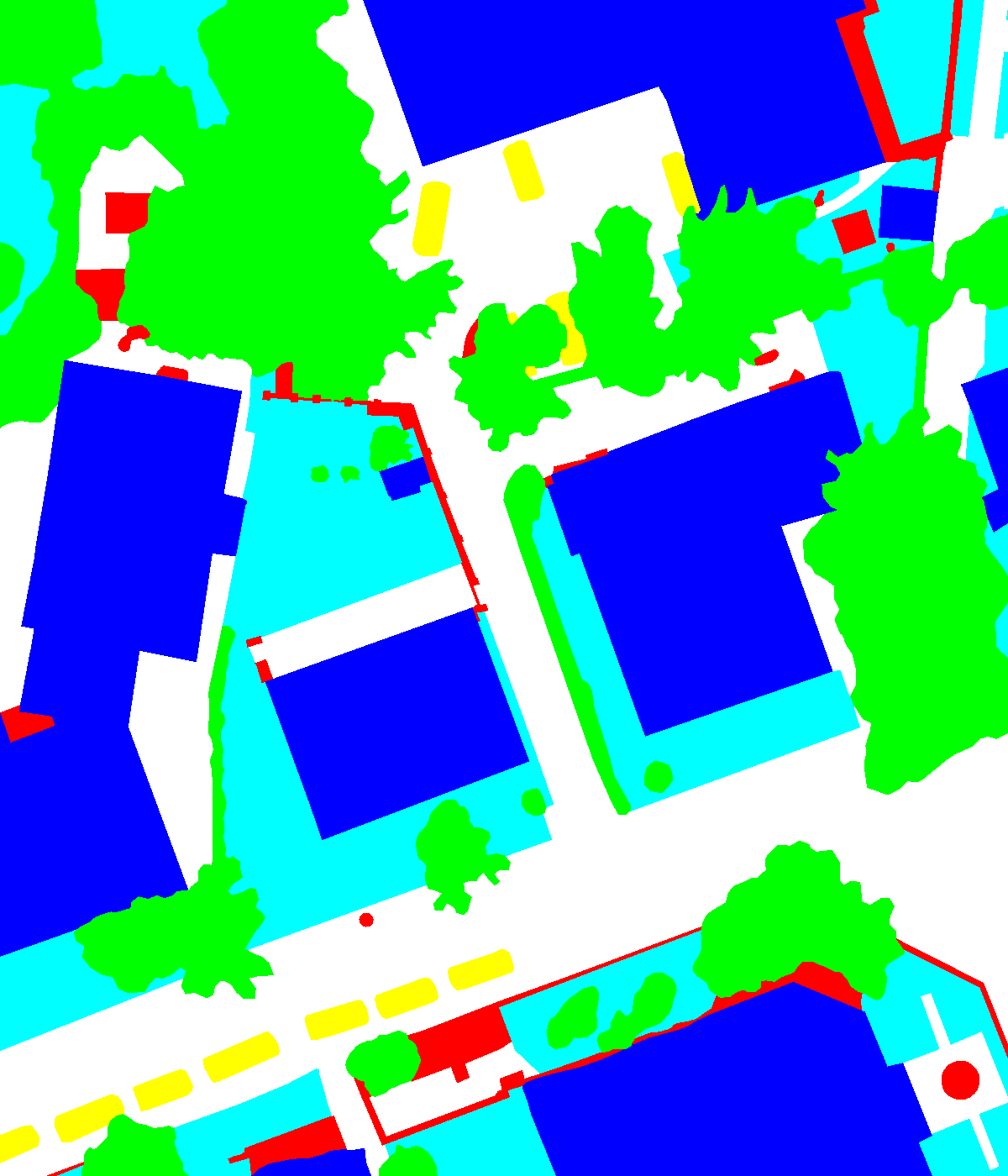} \hfill
    \includegraphics[width=0.24\textwidth]{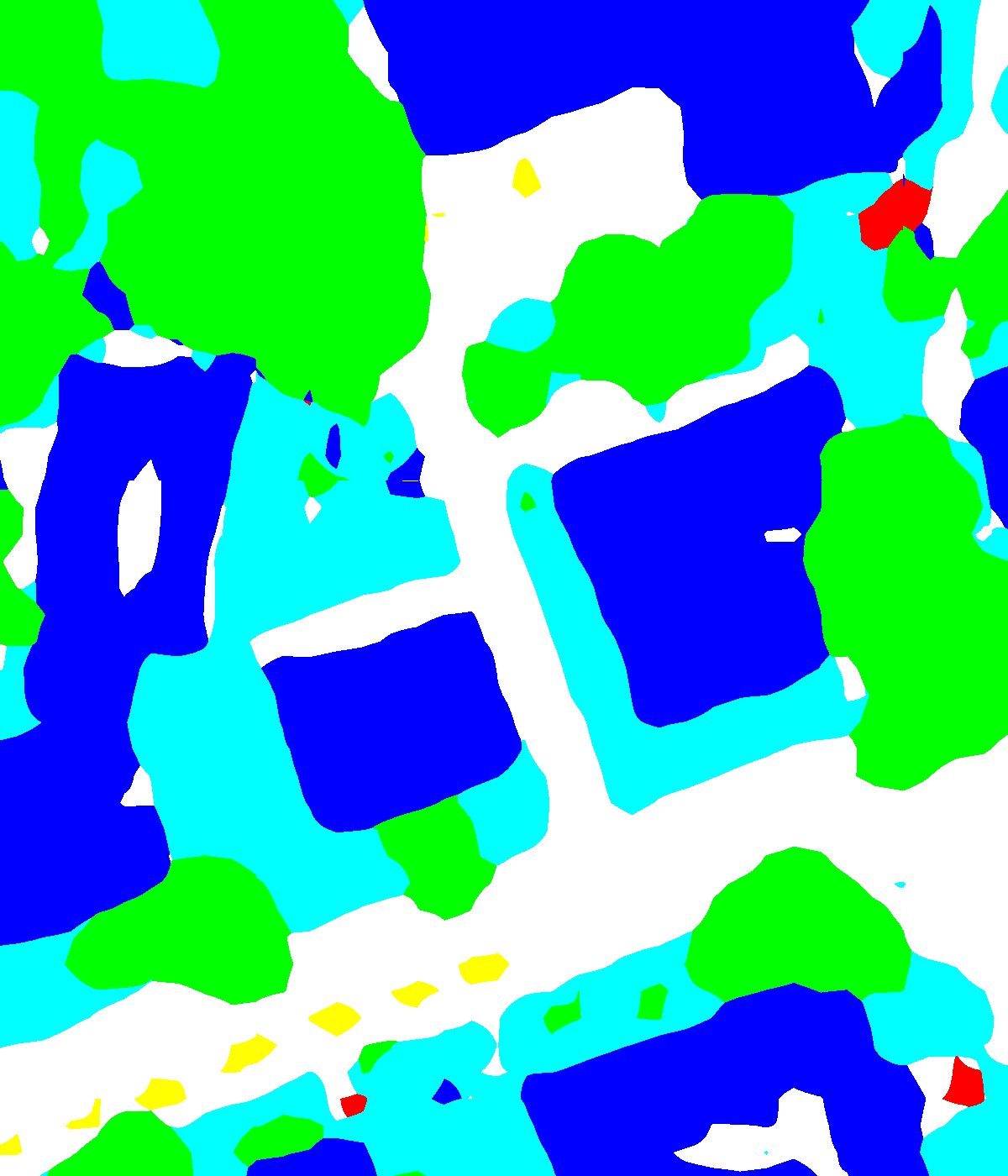} \hfill
\includegraphics[width=0.24\textwidth]{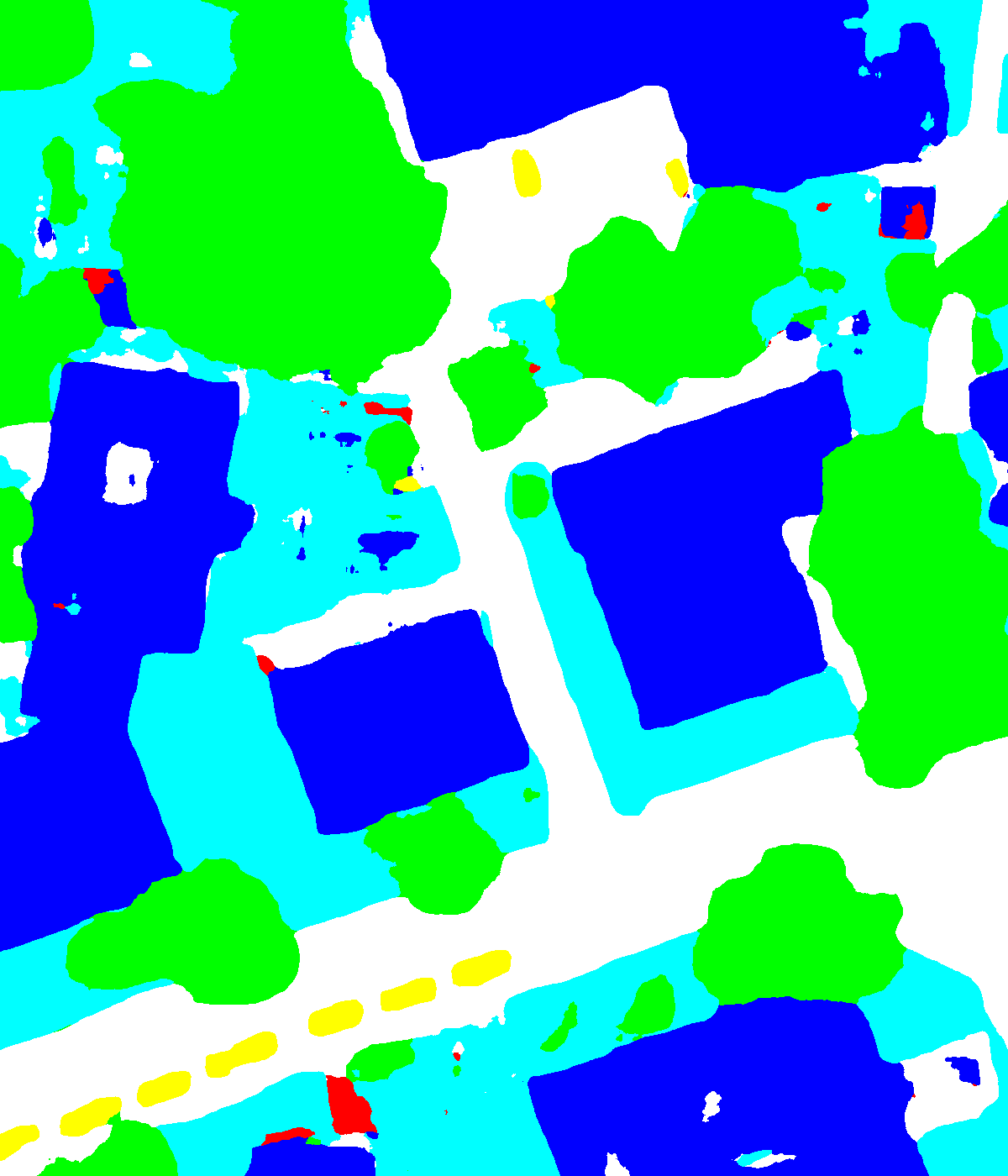} \hfill
\caption{Example semantic labelling output for Potsdam data set using networks
  trained with no-downsampling.  From left to right: input RGB image, ground
  truth labels, downsampling FCN, no-downsampling FCN. \label{fig:pNoDsResults}}
\end{figure}

The experiment is repeated for the Potsdam data set.  The same network
architecture is used, but due to its higher resolution the stride of the first
convolutional layer is set to 2.  Consequently the CNN filter stride of the
network is $s=32$.  All five channels of data are fed as input to the network:
RGB, infra-red (IR), and DSM.  The results are shown in
Table~\ref{tab:trainNoDsResultsPotsdam}.  The improvement of 1.98\% is much more
significant than for the Vaihingen data set due to the increased downsampling
rate of the network (32 for Potsdam, 16 for Vaihingen); see
Section~\ref{sec:dsAcc} for more discussion on this point.  The accuracy for
cars improved dramatically, by almost 20\%.  The accuracy on fine
structures is most severely effected by downsampling and interpolation.  These fine structures 
benefit the most from the high resolution of the no-downsampling network.  A
comparison of the labelling outputs is shown in Figure~\ref{fig:pNoDsResults}.

\begin{table*} [tb]
\scriptsize
  \centering
  \caption{ FCN semantic labelling results on the Potsdam validation set, trained with no-downsampling.  
  }
  \label{tab:trainNoDsResultsPotsdam}
  {
  \begin{tabular}{p{.28\textwidth}|c|c|c|c|c|c|c|c}
  \hline
   & \scriptsize{Imp. surf.} & \scriptsize{Building} & \scriptsize{Low veg.} & \scriptsize{Tree} & \scriptsize{Car} & \scriptsize{Unknown} & \scriptsize{Overall} & \scriptsize{Overall}\\
   &&&&&&& \scriptsize{ F1} & \scriptsize{ Acc.}\\
  \hline
  \hline  

  FCN trained with downsampling & $84.39\%$  & $89.95\%$  & $81.09\%$  & $75.71\%$  & $71.04\%$  & $67.97\%$  & $78.36\%$  & $82.16\%$ \\

  FCN trained with no-downsampling  & $86.52\%$  & $90.78\%$  & $83.01\%$  & $78.41\%$  & $90.28\%$  & $68.67\%$  & $82.94\%$  & $84.14\%$ \\

  \hline
  \end{tabular}
  }
\end{table*}

To compare the no-downsampling FCN approach to other methods, the network was
applied to the unlabelled test images of the Vaihingen data set and the results
submitted to the challenge moderator.  The network was only trained on the
training images listed in Table~\ref{tab:subTrainVal}.  The result is shown in
Table~\ref{tab:vaihingenLeaderbordNoDS} as DST\_1, giving a higher overall accuracy
than competing methods (at the time of writing) including the work
of~\cite{paisitkriangkraiEtAl:cvprw2015}.  Adding the hand-crafted features and
CRF steps to improve the labelling (entry DST\_2) results in the highest
overall accuracy of 89.1\%.

\begin{table*} [tb]
  \centering
  \caption{ The effect of no-downsampling on computation time in CNNs.  }
  \label{tab:trainNoDsTiming}
  {
  \begin{tabular}{l|c|c|c|c}
  \hline
& \multicolumn{2}{c|}{training}& \multicolumn{2}{|c}{test} \\ \hline
data set  &  downsampling & no-downsampling & downsampling & no-downsampling \\ \hline
  \hline  
Vaihingen & 1.5 hrs       & 2 days 14 hours & 1.05 s/MPix & 3.9 s/MPix \\
Potsdam   & 8 hrs 15 min  & 16 days 16 hours & 1.2 s/MPix & 4.3 s/MPix \\
  \hline
  \end{tabular}
  }
\end{table*}

\begin{table*} [tb]
  \centering
  \caption{ ISPRS Challenge Vaihingen Leaderboard Results~\cite{ISPRS}.  {DST\_1} is FCN, {DST\_2} is FCN+RF+CRF, corresponding to the validation set results in Table~\ref{tab:trainNoDsResults}. }
  \label{tab:vaihingenLeaderbordNoDS}
  {
  \begin{tabular}{l|c|c|c|c|c|c}
  \hline
   & \small{Imp. surf.} & \small{Building} & \small{Low veg.} & \small{Tree} & \small{Car} & \small{Overall Acc.}\\
  \hline
  \hline  

UT\_Mev	& $84.3\%$	& $88.7\%$	& $74.5\%$	& $82.0\%$	& $9.9\%$	& $81.8\%$ \\
SVL\_3	& $86.6\%$	& $91.0\%$	& $77.0\%$	& $85.0\%$	& $55.6\%$	& $84.8\%$ \\
HUST	& $86.9\%$	& $92.0\%$	& $78.3\%$	& $86.9\%$	& $29.0\%$	& $85.9\%$ \\
ONE\_5	& $87.8\%$	& $92.0\%$	& $77.8\%$	& $86.2\%$	& $50.7\%$	& $85.9\%$ \\
ADL\_3	& $89.5\%$	& $93.2\%$	& $82.3\%$	& $88.2\%$	& $63.3\%$	& $88.0\%$ \\
UZ\_1   & $89.2\%$      & $92.5\%$      & $81.6\%$      & $86.9\%$      & $57.3\%$      & $87.3\%$ \\
DLR\_1  & $90.2\%$      & $92.1\%$      & $82.2\%$      & $89.5\%$      & $80.5\%$      & $88.4\%$ \\
DLR\_2  & $90.3\%$      & $92.3\%$      & $82.5\%$      & $89.5\%$      & $76.3\%$      & $88.5\%$ \\
DST\_1  & $90.3\%$      & $93.5\%$      & $82.5\%$      & $88.8\%$      & $73.9\%$      & $88.7\%$ \\
{\bf DST\_2}    & $90.5\%$	& $93.7\%$	& $83.4\%$	& $89.2\%$	& $72.6\%$	& $\bf 89.1\%$ \\
  \hline
  \end{tabular}
  }
\end{table*}

\paragraph{Effect of Downsampling on Accuracy}\label{sec:dsAcc}

The effect of CNN filter stride on no-downsample training accuracy is further
examined in the following experiment.  A series of networks of increasing depth
is trained on the Vaihingen data set.  The network starts as a single
fully-connected layer turned convolutional, and is made deeper and deeper.
The 5 architectures examined are shown in Table~\ref{tab:convFtrResResults}.
The second fully-connected layer has 512 neurons.  The network is trained as an
FCN and the validation set accuracy is shown in the table.  The results get
progressively better as the network gets deeper, validating the paradigm of deep
learning.  This also decreases the spatial resolution of the last convolutional
layer, making it easier for the fully connected layers to learn spatial feature
relationships.  Consequently as the network gets deeper, the output downsampling
becomes more severe, which negatively impacts the accuracy of semantic
labelling.  Despite this effect, the accuracy still improved with depth.  The
output is restored to full resolution using bilinear interpolation.

\begin{table*} [tb]
  \centering
  \caption{ The effect of downsampling on CNN accuracy.  Five FCNs with increasing numbers of convolution, pooling and $2\times$ downsampling layers are trained on the Vaihingen data set, with and without the no-downsampling strategy.  Columns 5 and 6 show the validation set accuracy with downsampling and with no-downsampling, and the last column shows their difference.  As depth increases so does the accuracy, and the downsampling rate (convolutional filter stride).  No-downsampling has greater effect for FCNs with more downsampling.}
  \label{tab:convFtrResResults}
  {
  \begin{tabular}{l|p{.25\textwidth}|c|c|c|c|c}
  \hline

Config.  & Convolutional Layers                               & FC             &Downsample & Validation & No-DS & 	Diff \\ 
         &                                                    & Layer 1        & factor    & Accuracy   &  val. acc. & 	 \\ 
  \hline
  \hline  
1 &	n/a                                                         &    64x64x32    &	1    & 79.83 & n/a 	    &  0.00   \\ \hline
2 &	5x5x32+pool (out=30x30) 	                            &    30x30x144   &   2   & 84.25 & 83.99 	    &  -0.26  \\ \hline
3 &	5x5x32+pool; 3x3x64+pool (out=14x14) 	                    &    14x14x661   &   4   & 86.59 & 86.75 	    &  0.16   \\ \hline
4 &	5x5x32+pool; 3x3x64+pool; 3x3x96+pool (out=6x6) 	    &    6x6x3578    &   8   & 86.87 & 87.41 	    &  0.54   \\ \hline
5 &	5x5x32+pool; 3x3x64+pool; 3x3x96+pool; 3x3x128+pool (out=2x2)&	2x2x31920    & 	16   & 87.11 & (**)87.52  &   0.41  \\ 

  \hline
  \end{tabular}
  }
\end{table*}

Next each network is trained using the no-downsampling version of the network,
which has the same number of parameters.  Again, accuracy improves with depth,
and is generally more accurate than the version trained with downsampling.  We
would expect the improvement to be more pronounced for deeper networks where the
downsampling factor is higher.  This is borne out by the last column of
Table~\ref{tab:convFtrResResults}: although the improvement is not monotonic, it
is generally higher for deeper networks.
These results indicate that no-downsample training can give significant
improvements for deep networks with a high downsample factor (\ie high FCN filter
stride).

\section{Pre-Trained CNNs}\label{sec:pretrain}

Pre-trained CNN weights have been found to be effective for a variety of visual tasks~\cite{Razavian2014CNN}.
In~\cite{penattiEtAl:cvprw2015} pre-trained CNNs were applied to classification
of aerial imagery patches and found to out-perform hand-crafted features.  This
is remarkable since the network is trained on multimedia images like dogs and
cats, and generalises to overhead imagery.  Here we propose to use pre-trained
networks for overhead semantic labelling.  The first reason is that even though
we have a very large number of overlapping image patches for training, they are
highly correlated and over-fitting may occur during training.  Instead of
training from random initialisation, fine-tuning pre-trained features might
improve generalisation.  The second reason is that randomly initialised networks
might focus too much on the spectral scene properties since this information is
more readily exploitable.  However for good generalisation the network should
rely more on appearance-based features and less on spectral content.  Using
pre-trained networks should encourage the network to make use of texture
features.  

\begin{figure}[htbp]
\centering
\includegraphics[width=\textwidth]{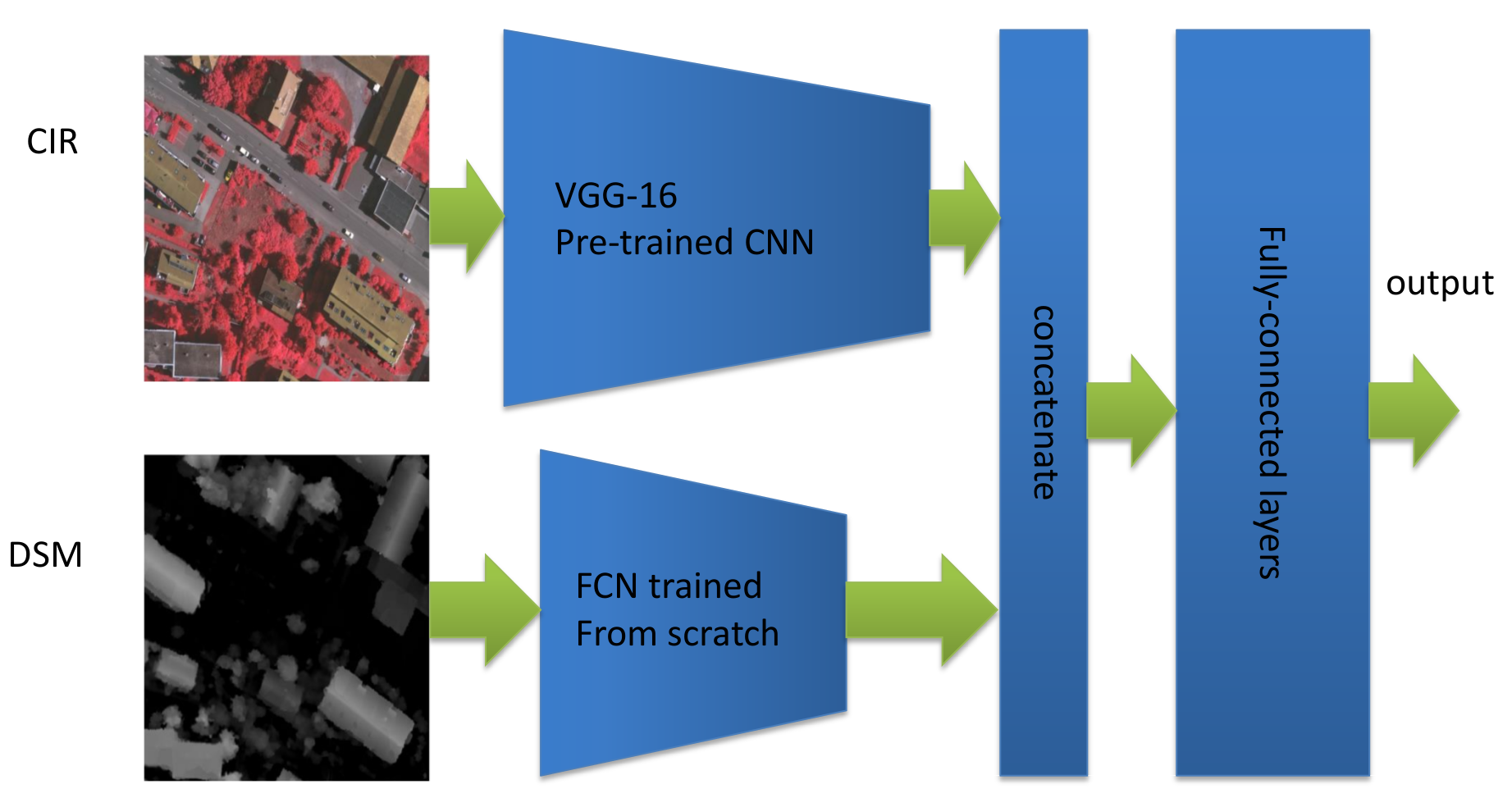}
\caption{Schematic of the hybrid network architecture, combining pre-trained
  image features with DSM features trained from scratch.}
\label{fig:hybridDiag}
\end{figure}

Fine-tuning pre-trained networks on 3-channel imagery has been performed many
times in the literature and is straightforward.  However in our experiments we
also use a normalised DSM channel, akin to depth information.  CNNs have
previously been applied to RGB-D data such as that generated by a Kinect sensor
for object recognition.  In~\cite{eitel2015multimodal} the depth channel is
converted to RGB using scene constraints and fed to a second three-channel
pre-trained network.  The features are combined with the output of an RGB
pre-trained network using a late fusion layer.  In our case those scene
constraints are not relevant.  Our proposed approach is a hybrid network: three
spectral bands are fed to the convolutional layers of a pre-trained network,
whereas the single DSM channel is fed to randomly initialised convolutional
layers.  These two sets of convolutional features are concatenated and serve as
input to two randomly initialised fully-connected layers.  The network
architecture is illustrated in Figure~\ref{fig:hybridDiag}.  The architecture of
the DSM sub-network is the same as the standard FCN, specified in
Section~\ref{sec:netDesc}.

\subsection{Experiments}

\begin{figure}[htbp]
\centering \hfill
    \includegraphics[width=0.24\textwidth]{figs/results_potsdam_gt_ex0} \hfill
\includegraphics[width=0.24\textwidth]{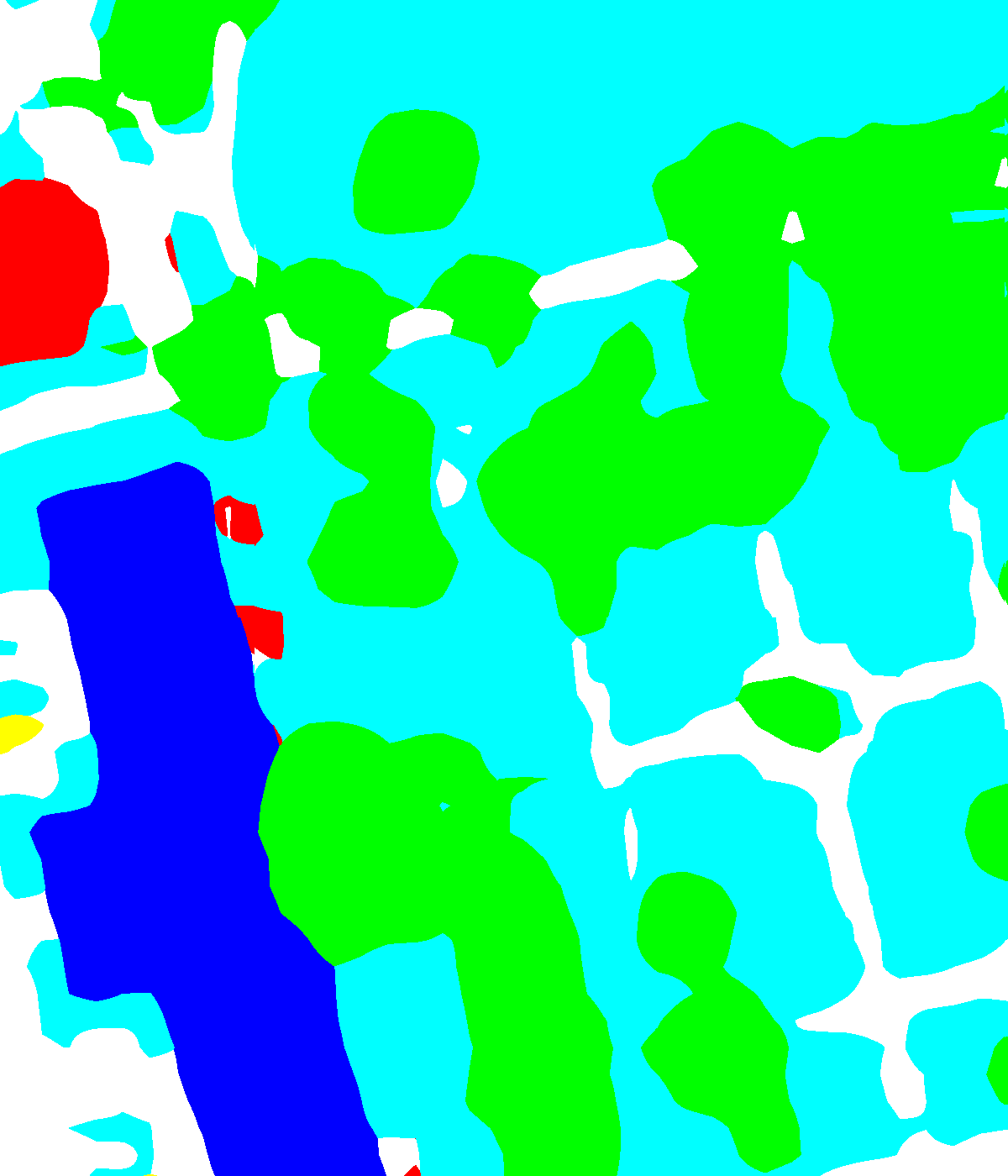} \hfill
  \includegraphics[width=0.24\textwidth]{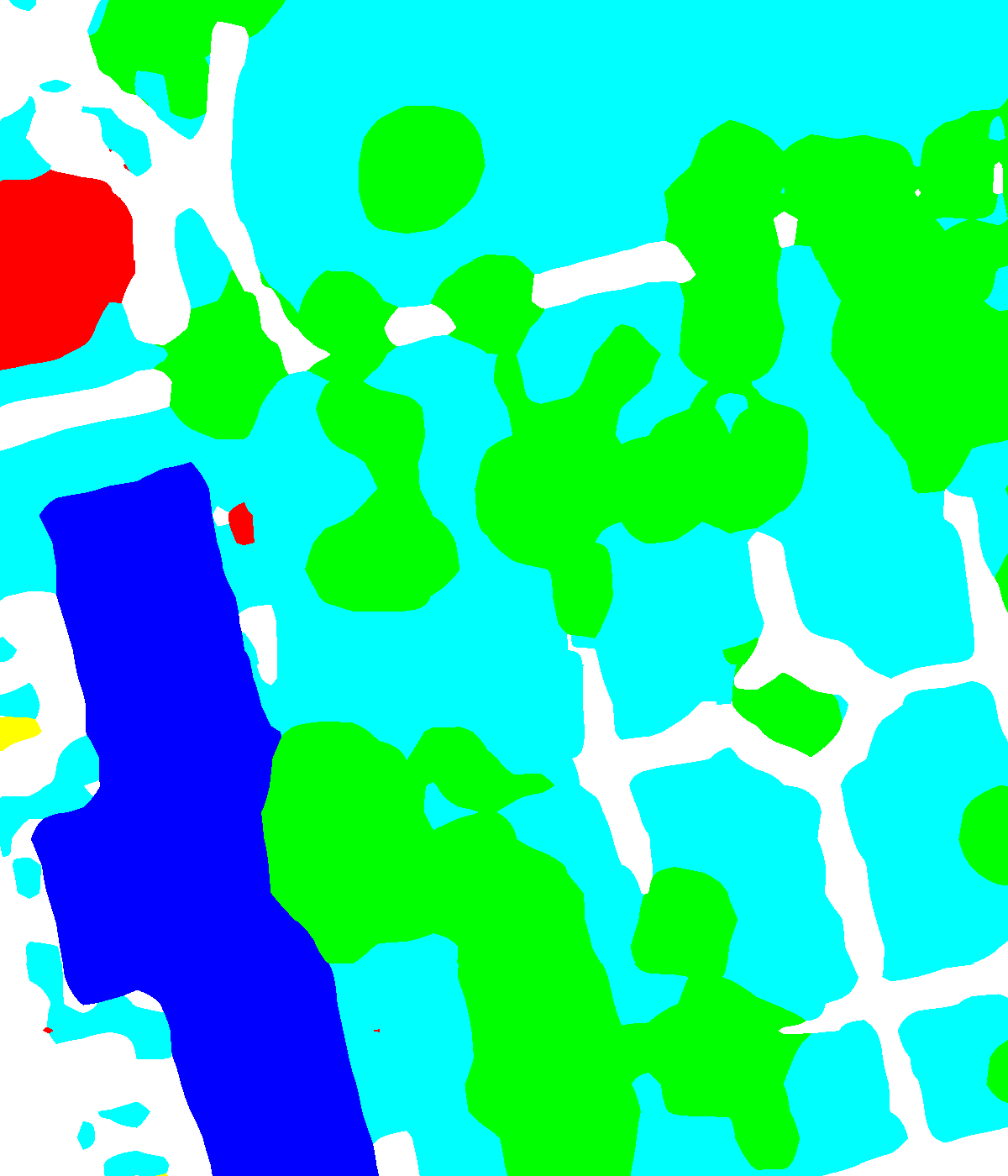} \hfill
  \includegraphics[width=0.24\textwidth]{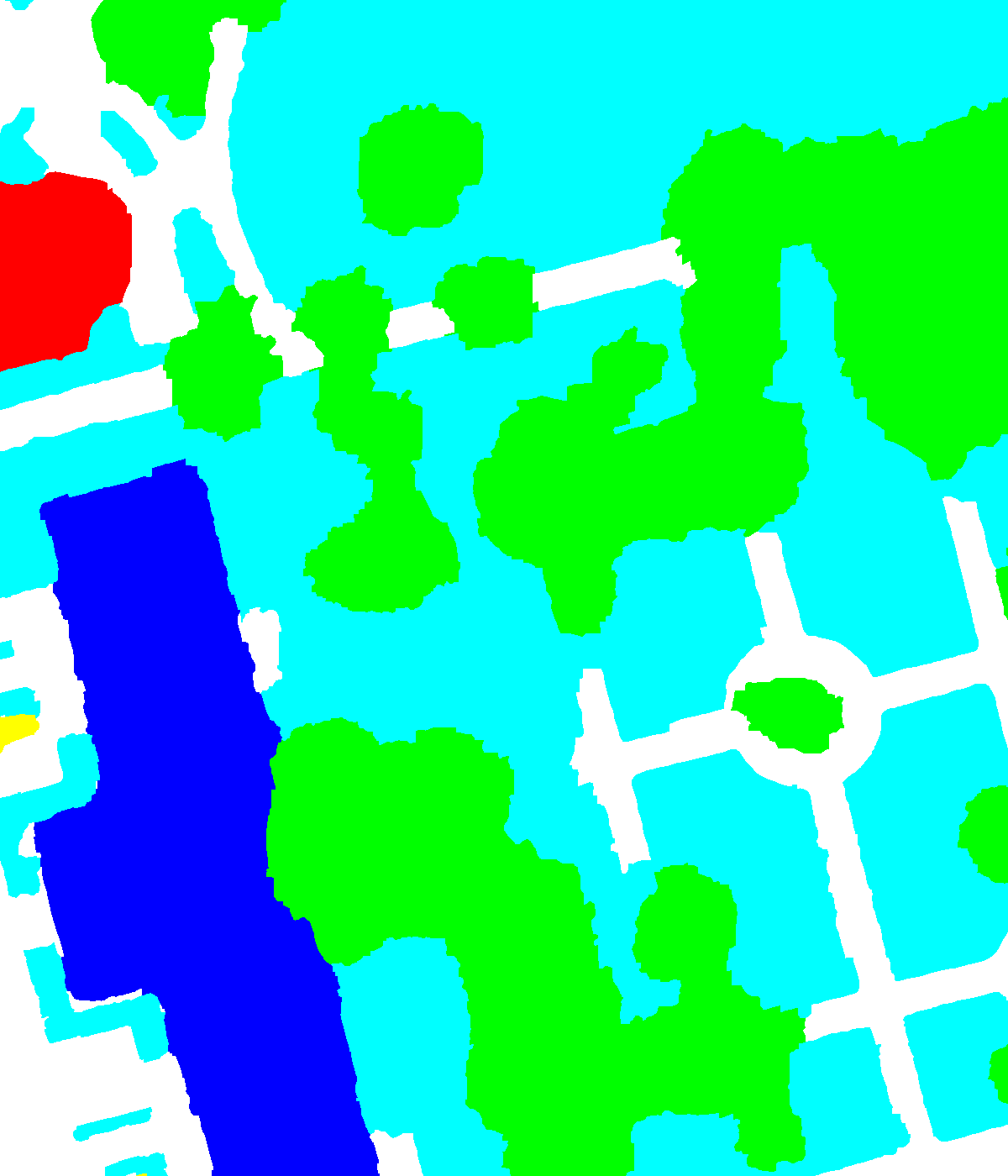} \hfill
    \includegraphics[width=0.24\textwidth]{figs/results_potsdam_gt_ex2} \hfill
\includegraphics[width=0.24\textwidth]{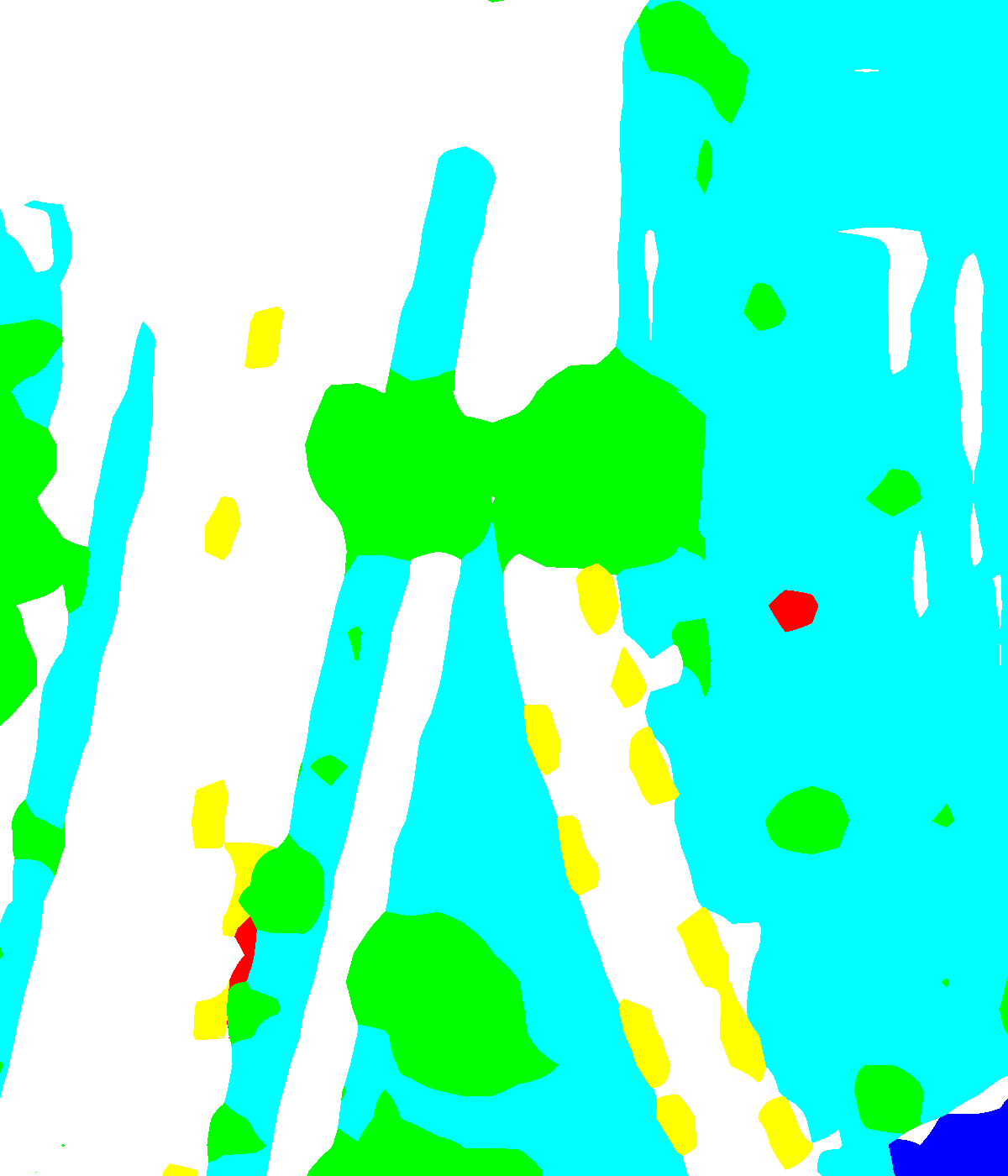} \hfill
  \includegraphics[width=0.24\textwidth]{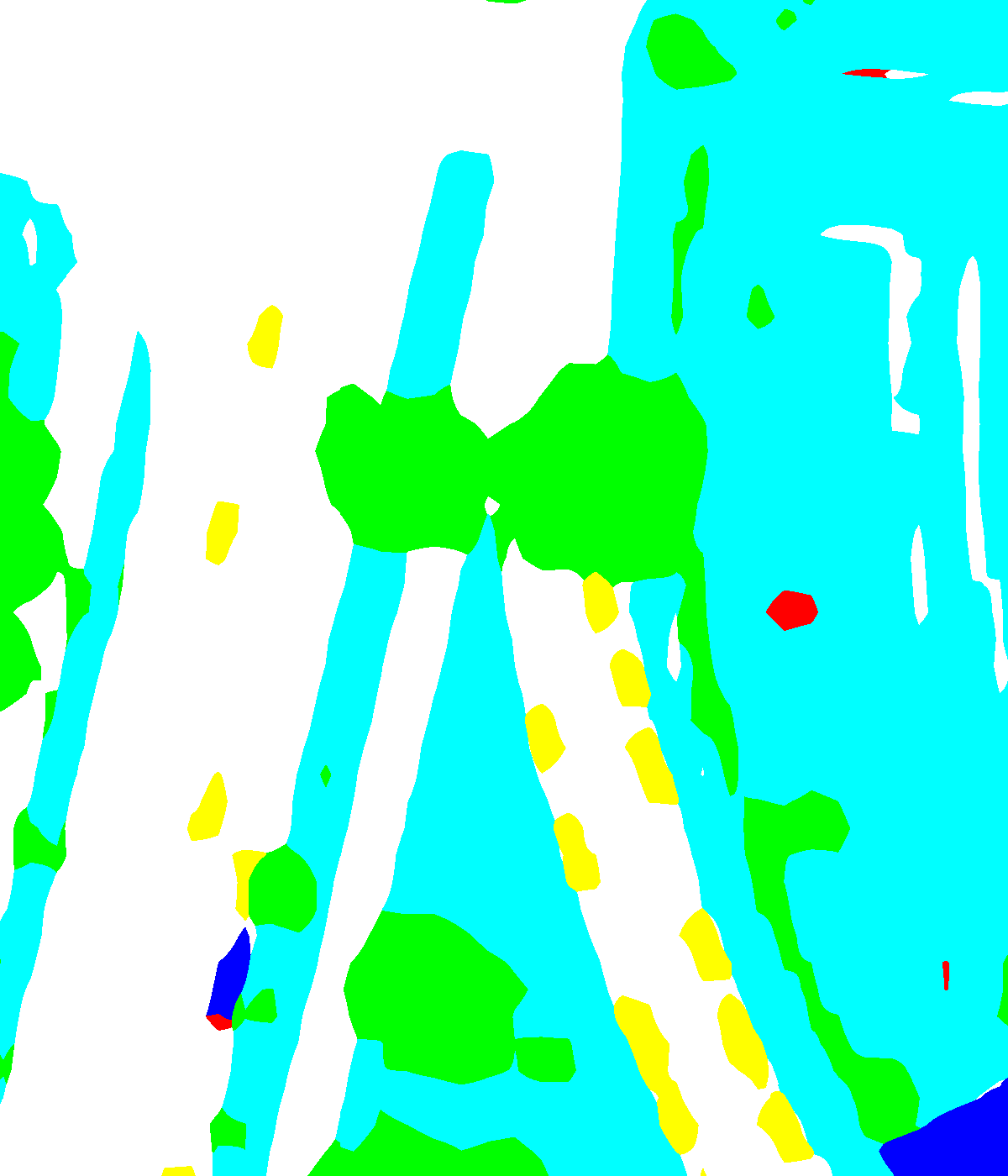} \hfill
  \includegraphics[width=0.24\textwidth]{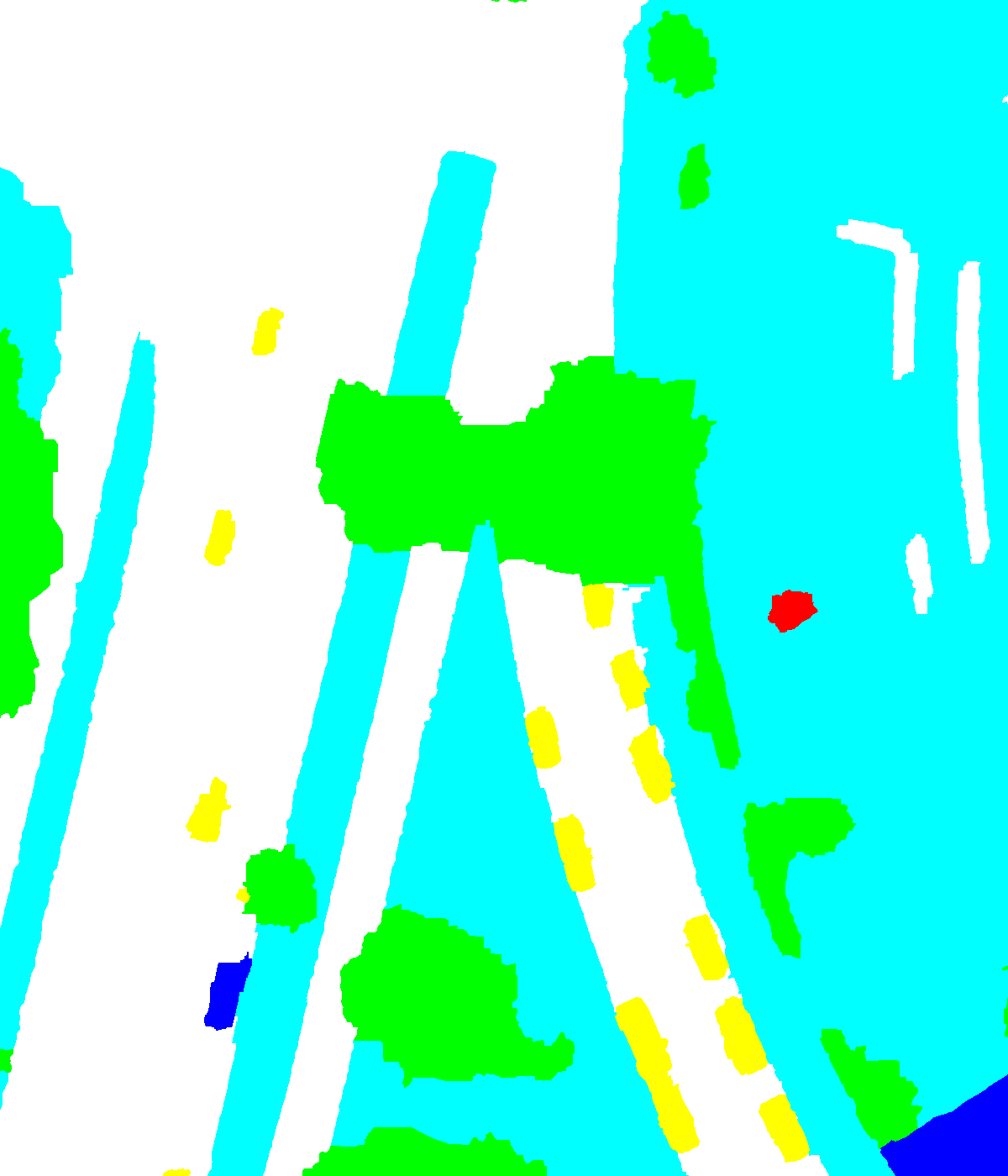} \hfill
    \includegraphics[width=0.24\textwidth]{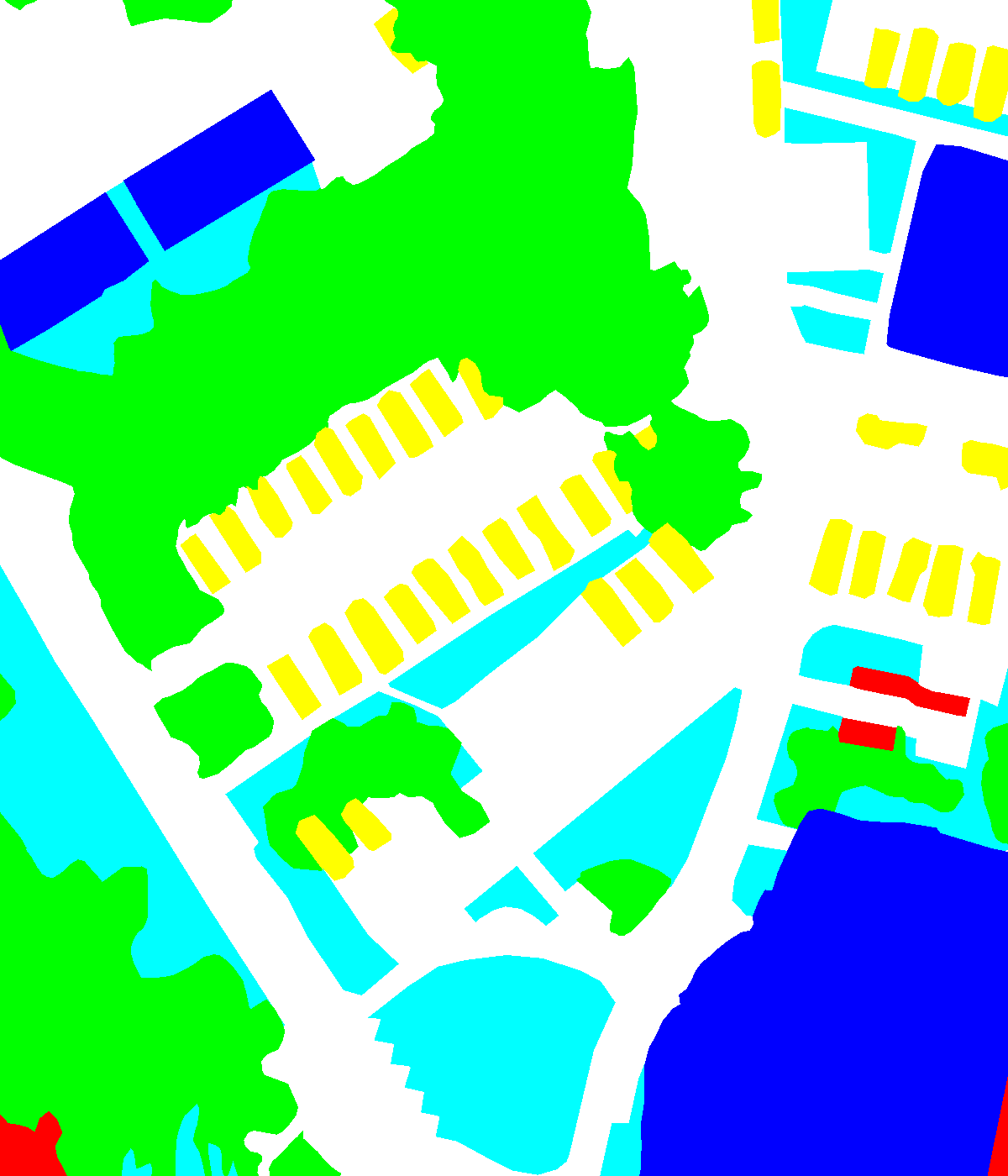} \hfill
\includegraphics[width=0.24\textwidth]{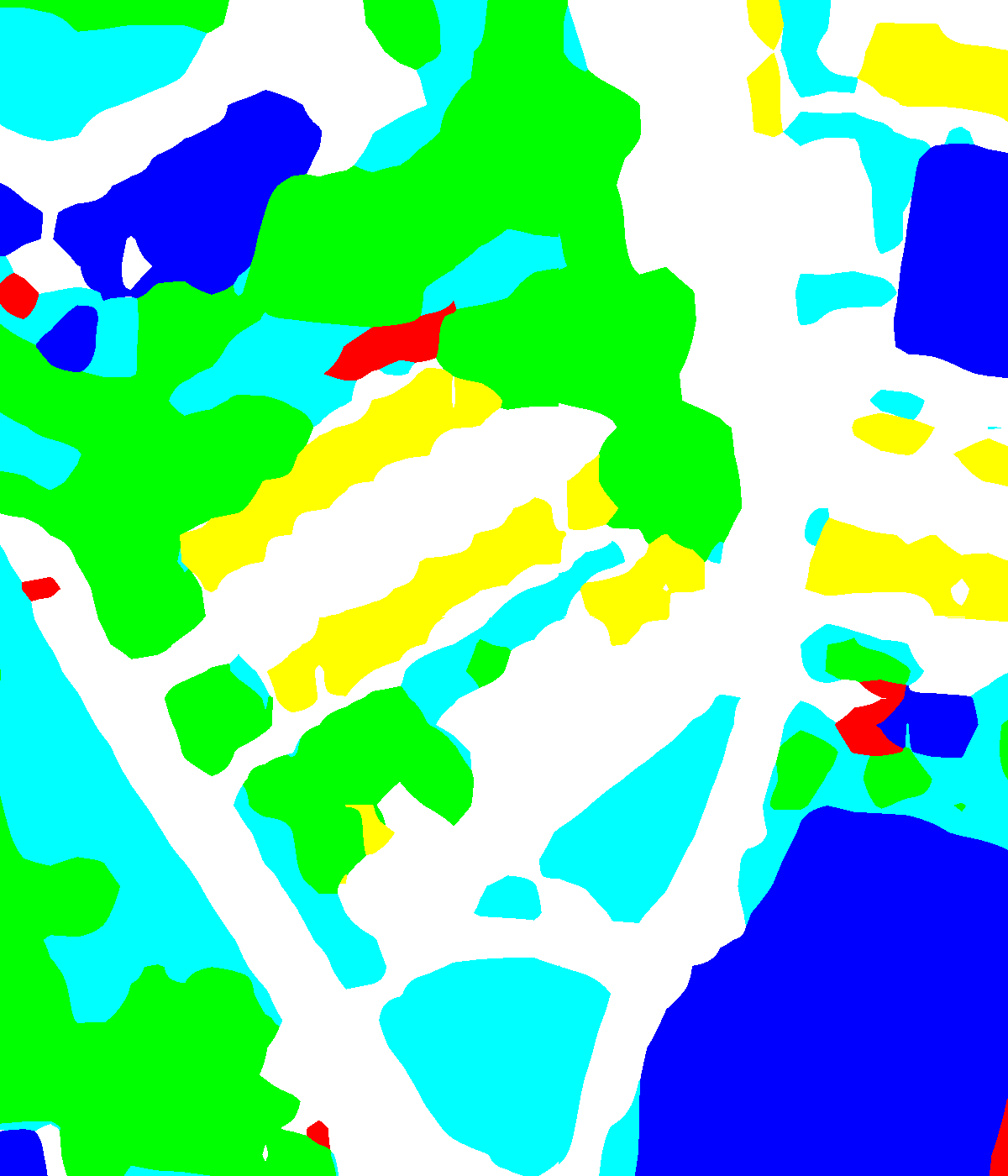} \hfill
  \includegraphics[width=0.24\textwidth]{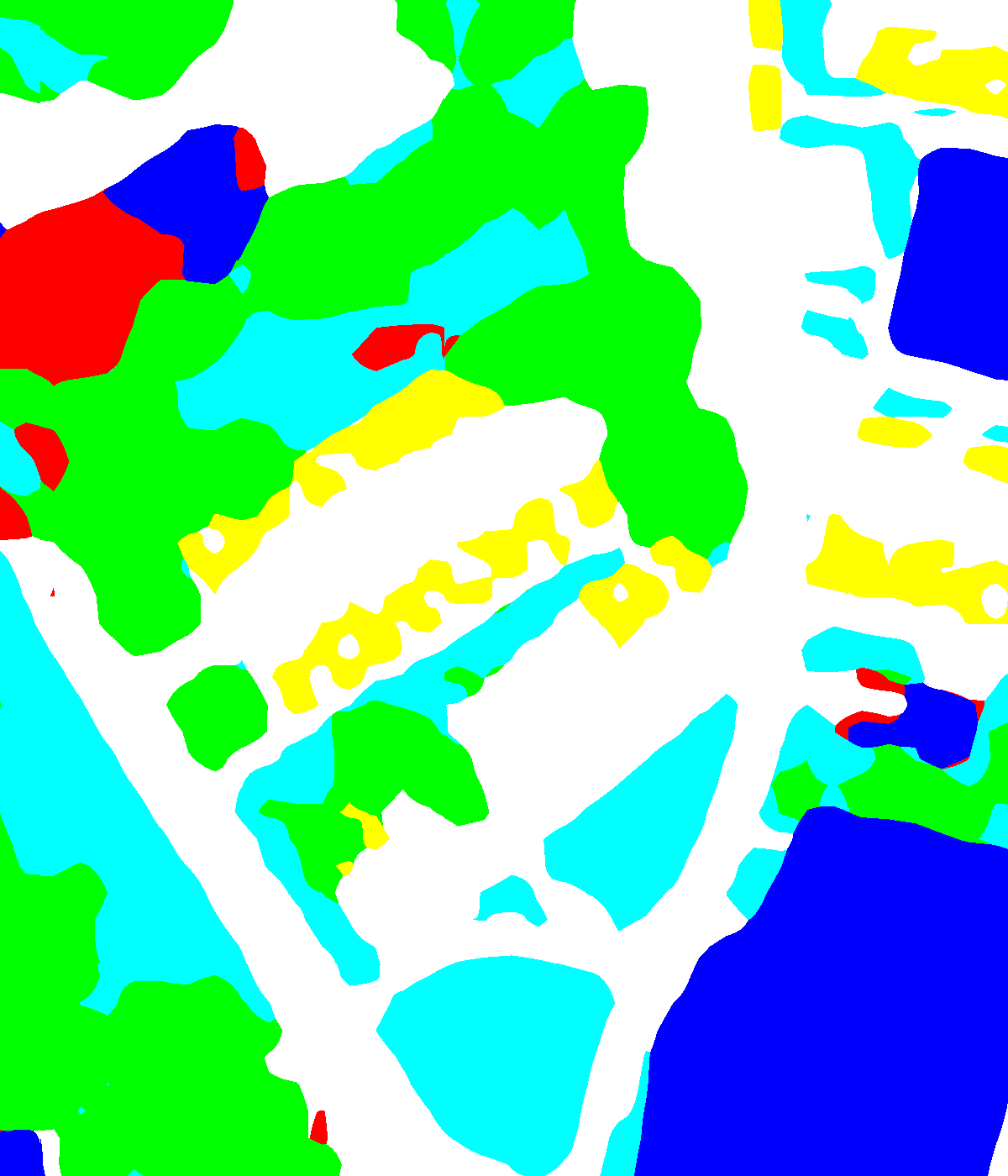} \hfill
  \includegraphics[width=0.24\textwidth]{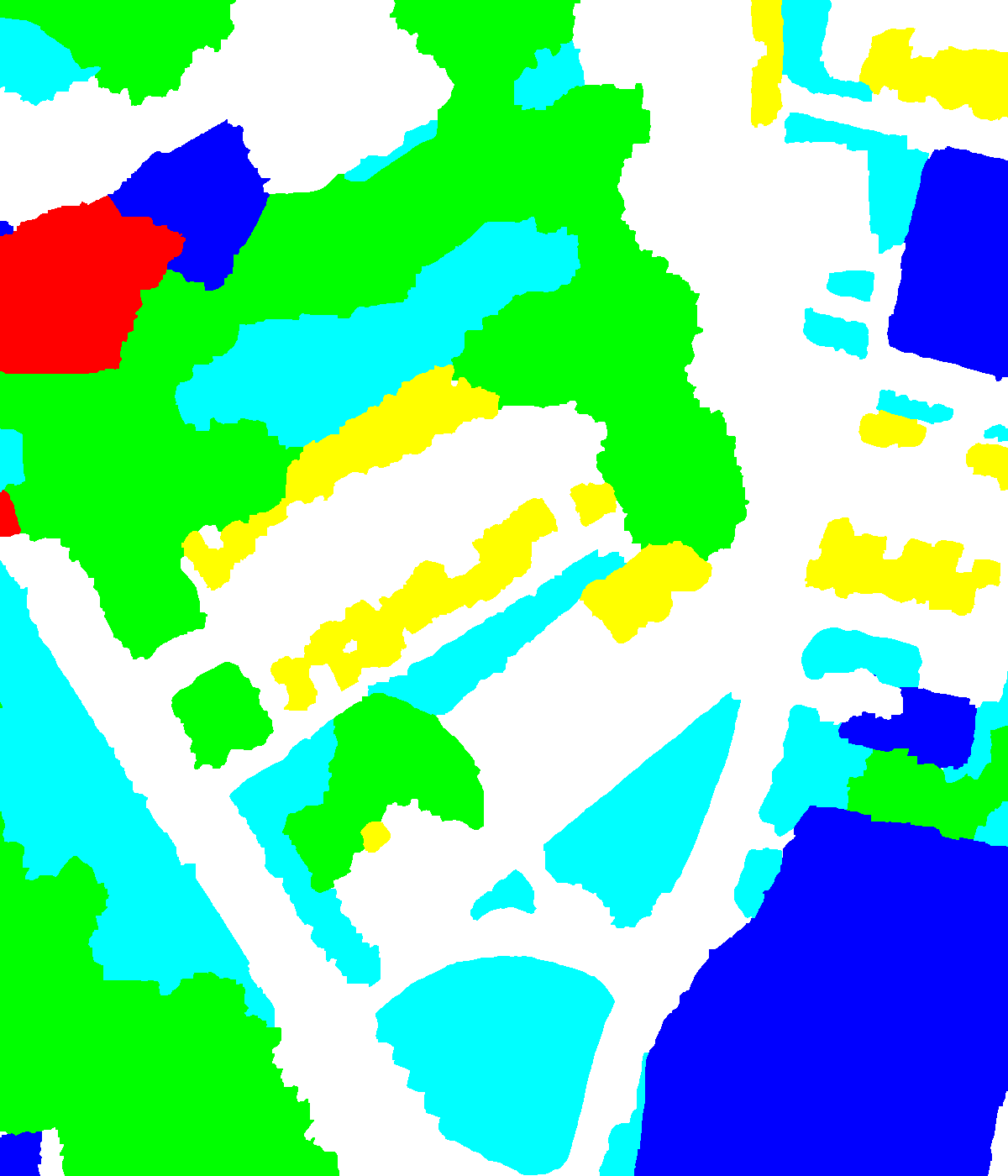} \hfill
  \caption{Comparison of pre-trained VGG FCN on different input channels for
    Potsdam data.  From left to right: ground truth, VGG on RGB, VGG on CIR
    (DST\_1), VGG on CIR + CRF (DST\_2).\label{fig:pPreTrainedResults}}
\end{figure}

The VGG-16 pre-trained network~\cite{simonyan2014very} is applied to
fully-convolutional semantic labelling of the Potsdam data.  The VGG net is
transformed to a fully-convolutional network using the net surgery approach
of~\cite{longEtAl:CVPR2015}.  While the convolutional weights are pre-trained,
the fully-connected layers are randomly initialised.  All pre-trained layers are
fine-tuned using a lower learning rate of 0.0001.  In the first experiment, the
RGB channels are fed as input to the network. The results are shown in the first
row of Table~\ref{tab:preTrainedPotsdam}.  When compared with the
randomly-initialised network that uses all five input channels there is a
considerable improvement to the overall accuracy of almost 1.91\% (see
Table~\ref{tab:trainNoDsResultsPotsdam}).  The value of pre-trained features for
analysis of this high-resolution imagery (5cm GSD) is evident.  Next the CIR
channels are used as input instead of RGB.  Row 2 of
Table~\ref{tab:preTrainedPotsdam} shows a marked increase in accuracy of
1.13\%.  The 1.82\% improvement in accuracy for low vegetation demonstrates the
discriminative power of the near infra-red band for plant-life.  Examples of the
labelling outputs are shown in Figure~\ref{fig:pPreTrainedResults}.  Comparing
columns 2 and 3, the CIR data facilitates improved delineation of vegetation.
The worst-classified class is the {\em unknown} category (shown in red), which
is any pixel not belonging to the other five classes.  The training data covers
a limited geographic area; a significantly more varied training set would be
required to accurately label the unknown class.

\begin{figure}[htbp]
\centering \hfill
  \includegraphics[width=0.24\textwidth]{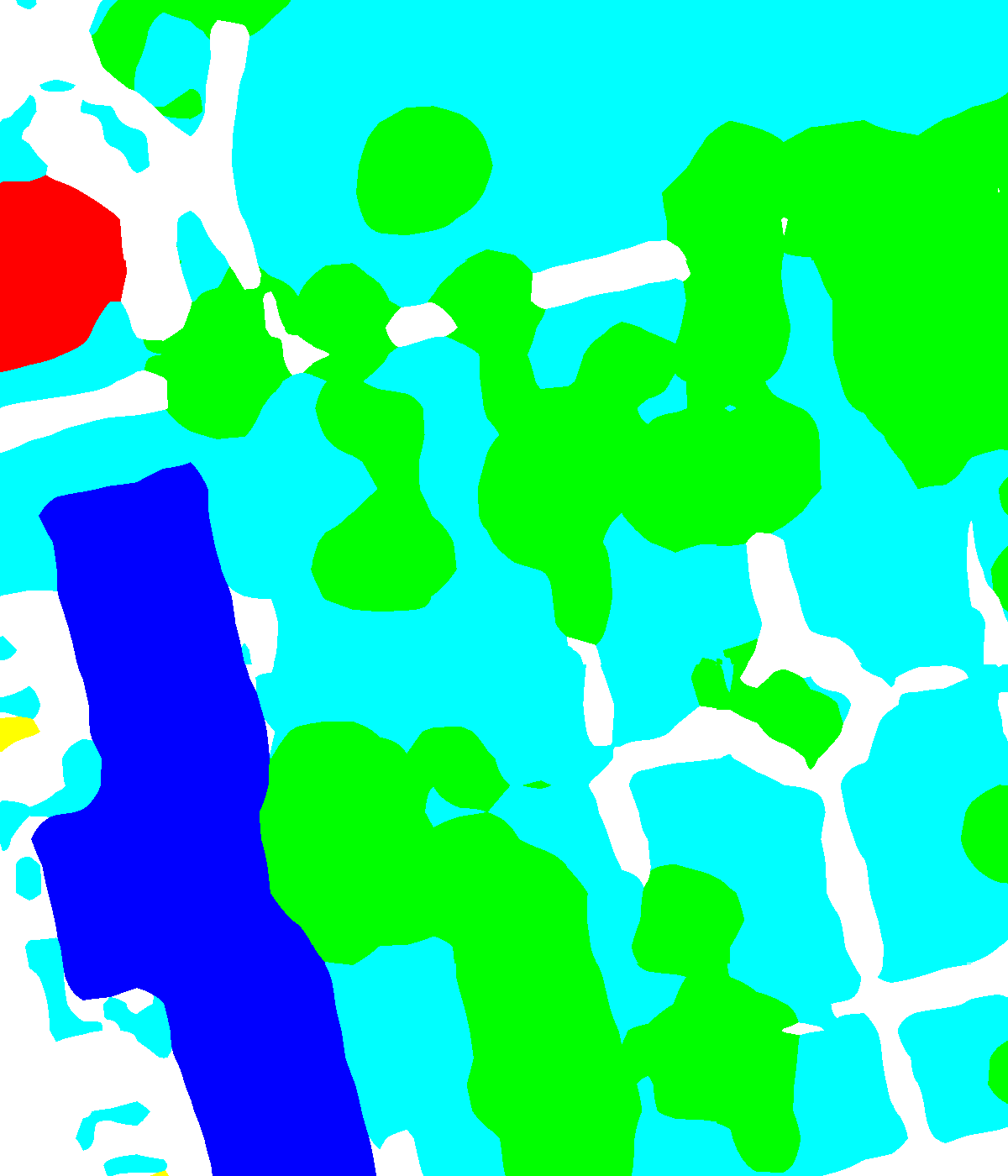} \hfill
  \includegraphics[width=0.24\textwidth]{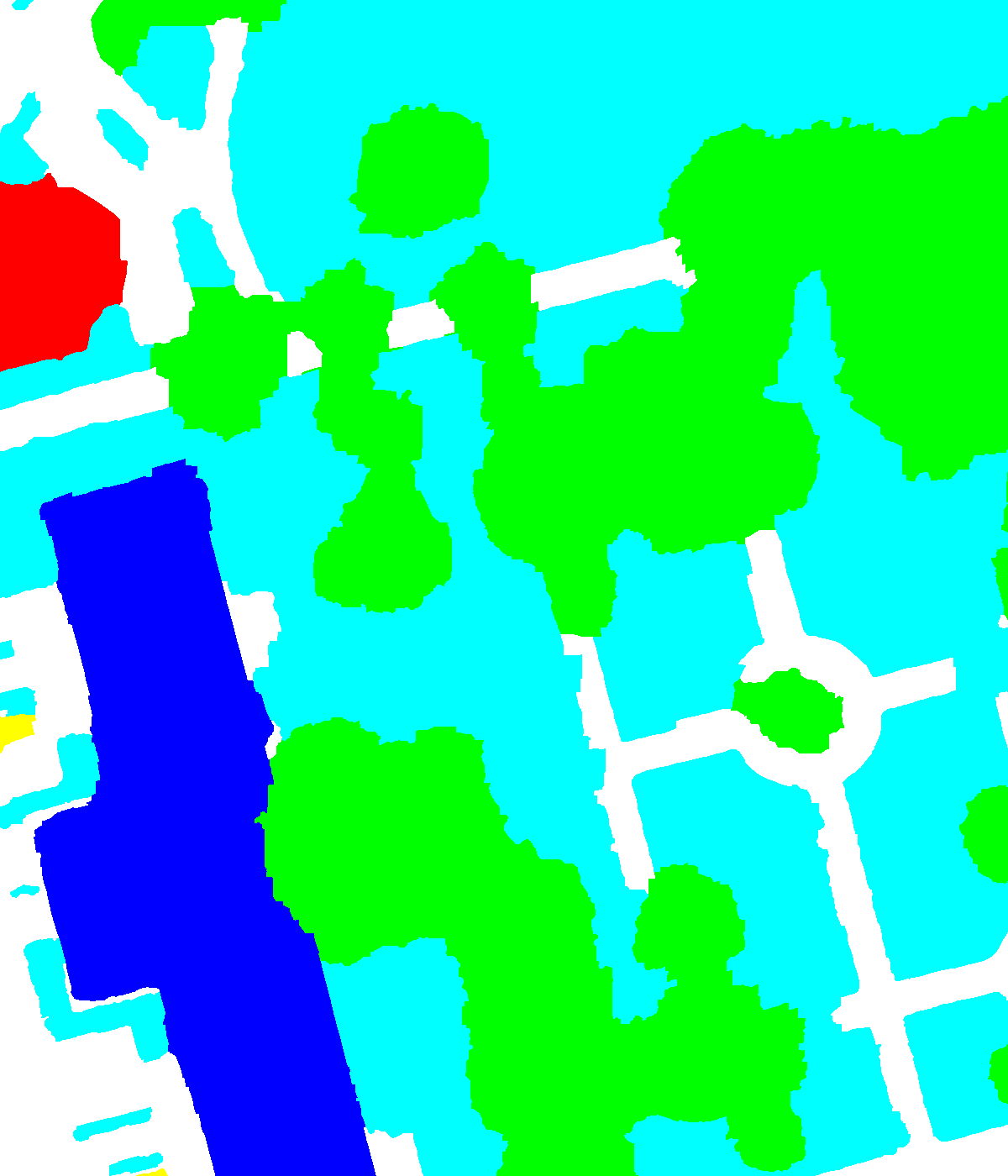} \hfill
  \includegraphics[width=0.24\textwidth]{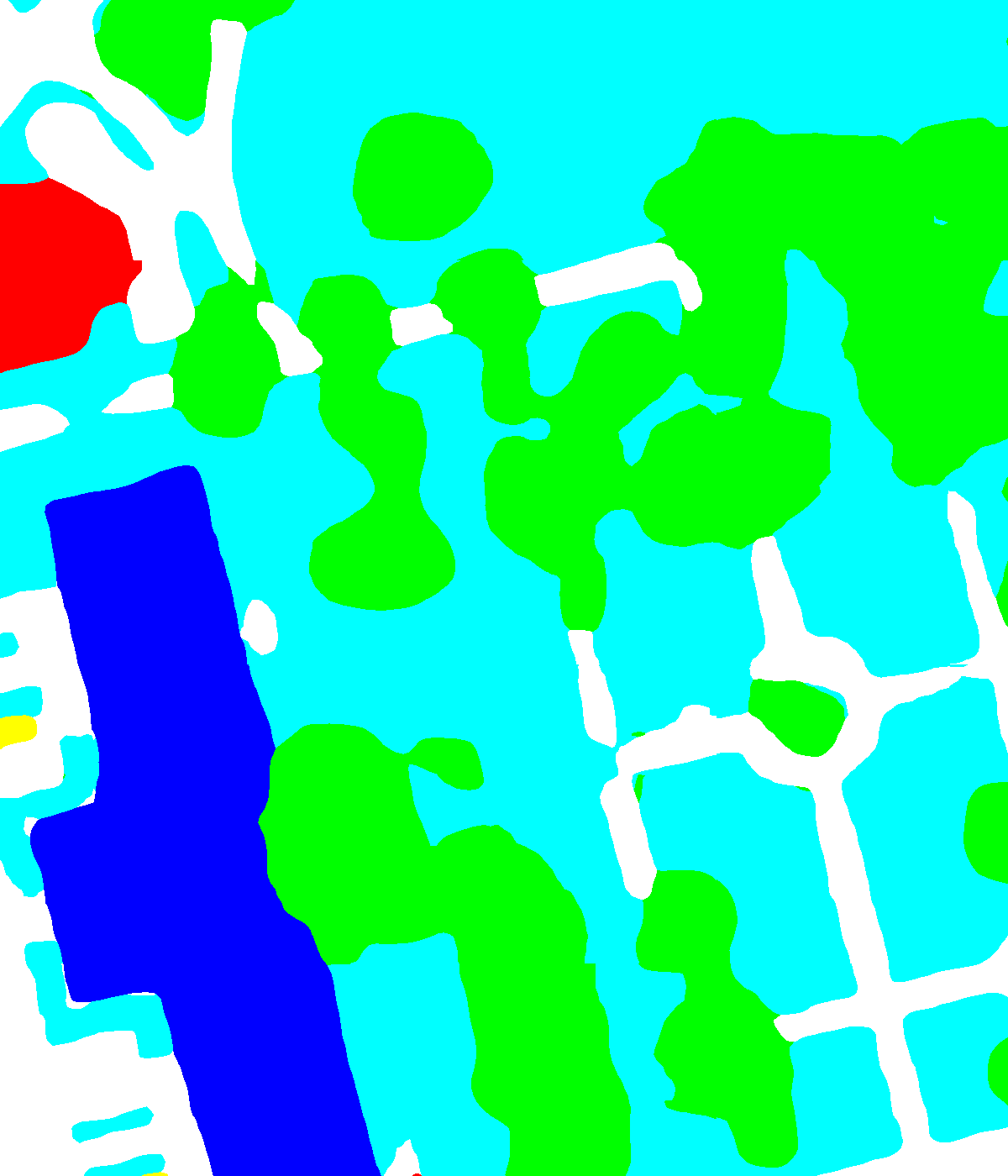} \hfill
  \includegraphics[width=0.24\textwidth]{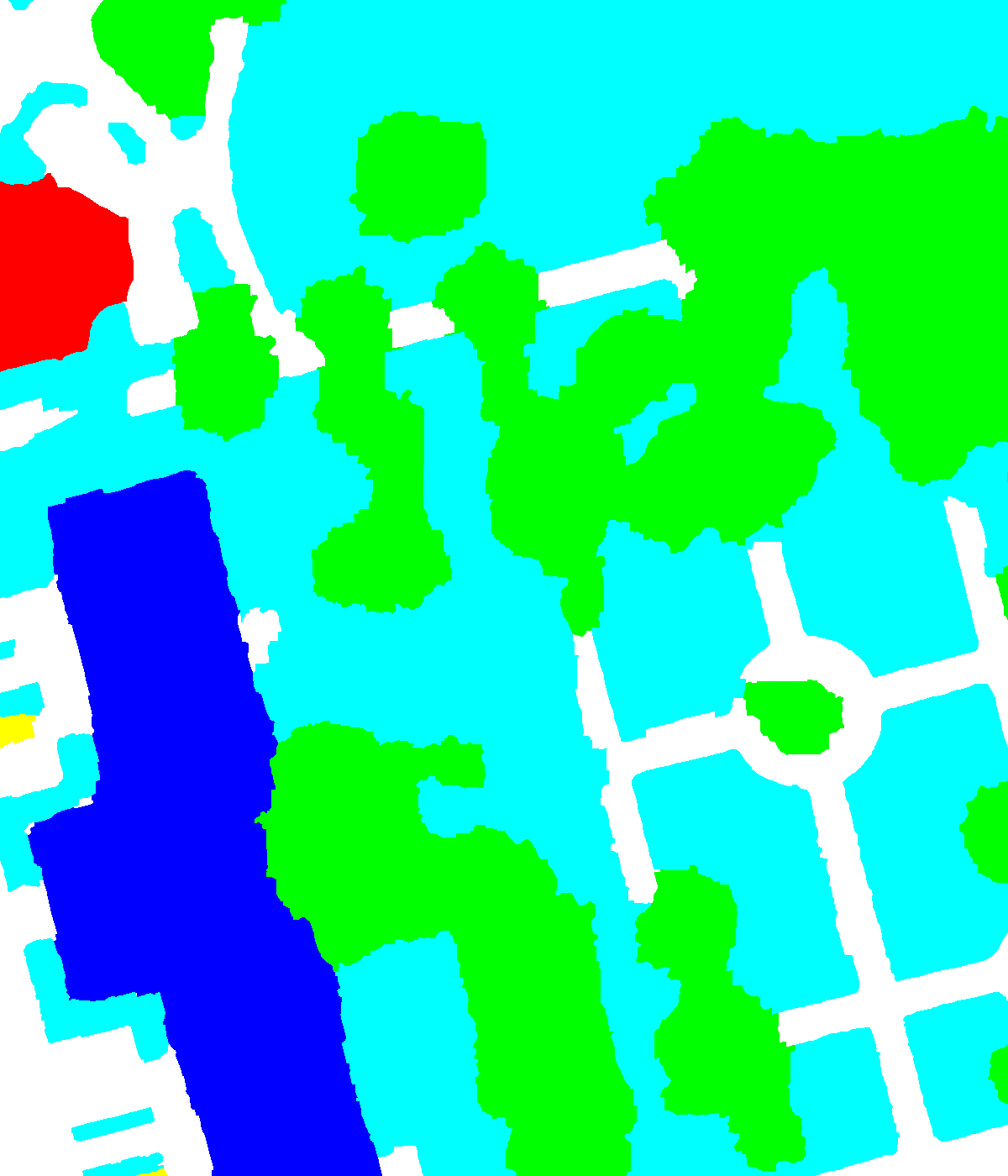} \hfill
  \includegraphics[width=0.24\textwidth]{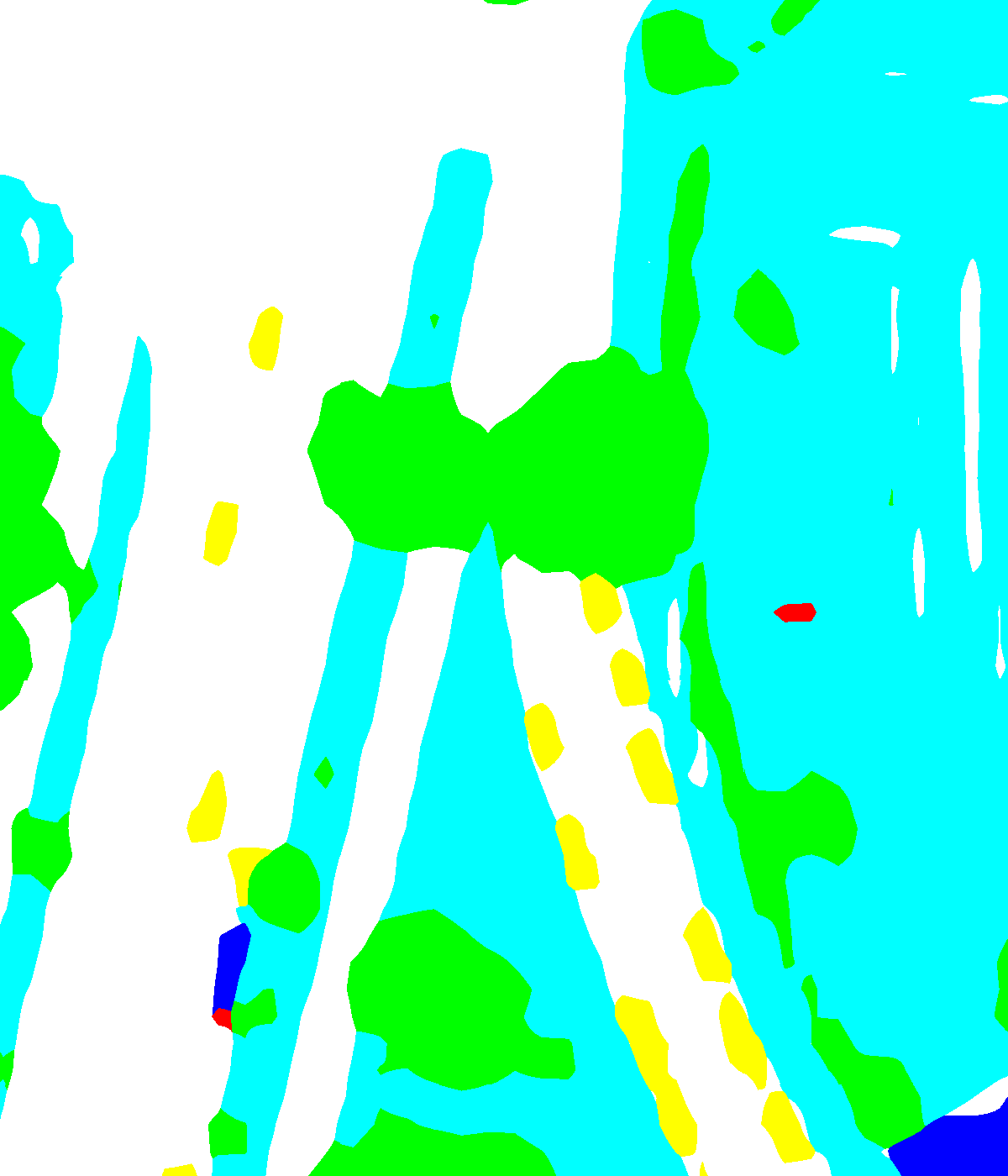} \hfill
  \includegraphics[width=0.24\textwidth]{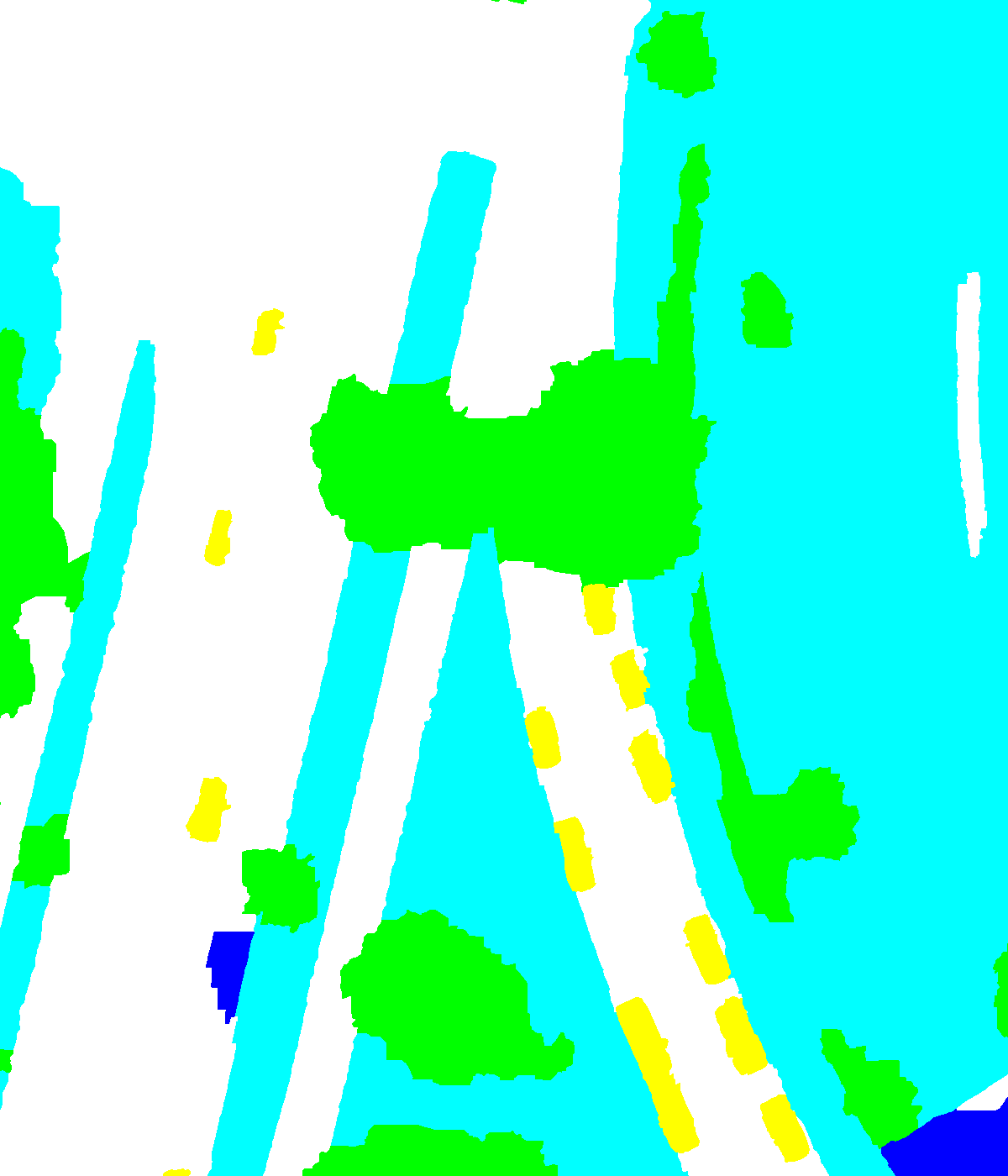} \hfill
  \includegraphics[width=0.24\textwidth]{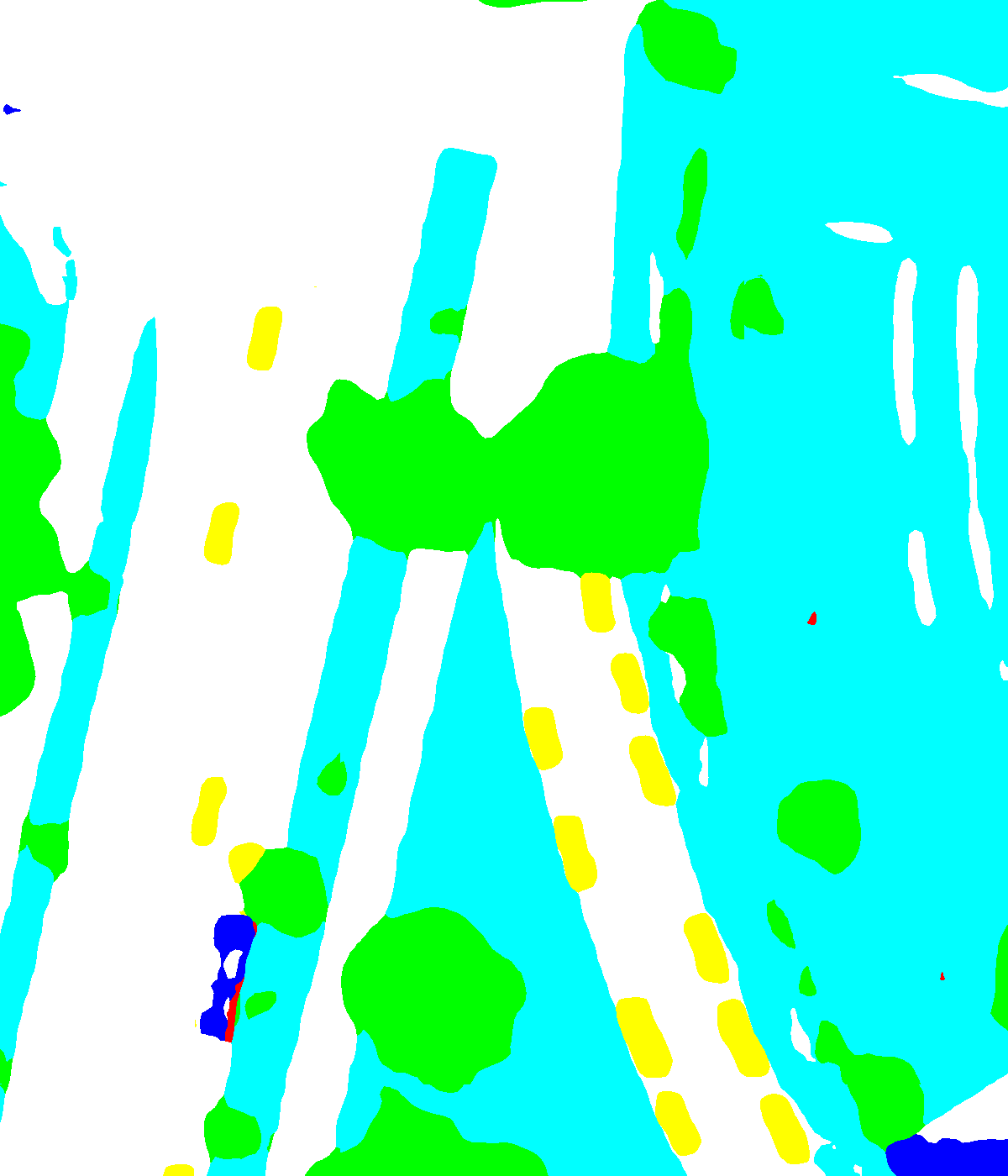} \hfill
  \includegraphics[width=0.24\textwidth]{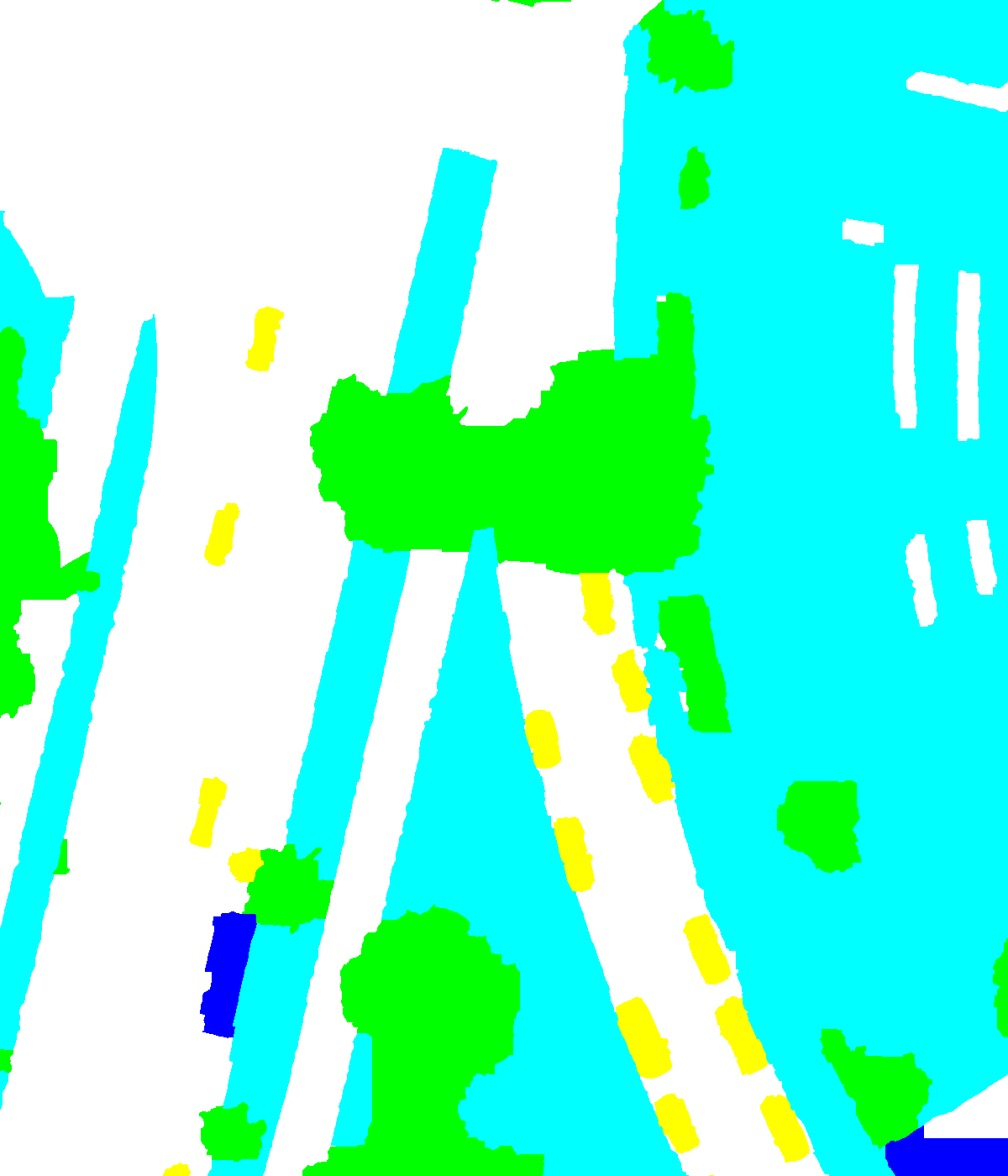} \hfill
  \includegraphics[width=0.24\textwidth]{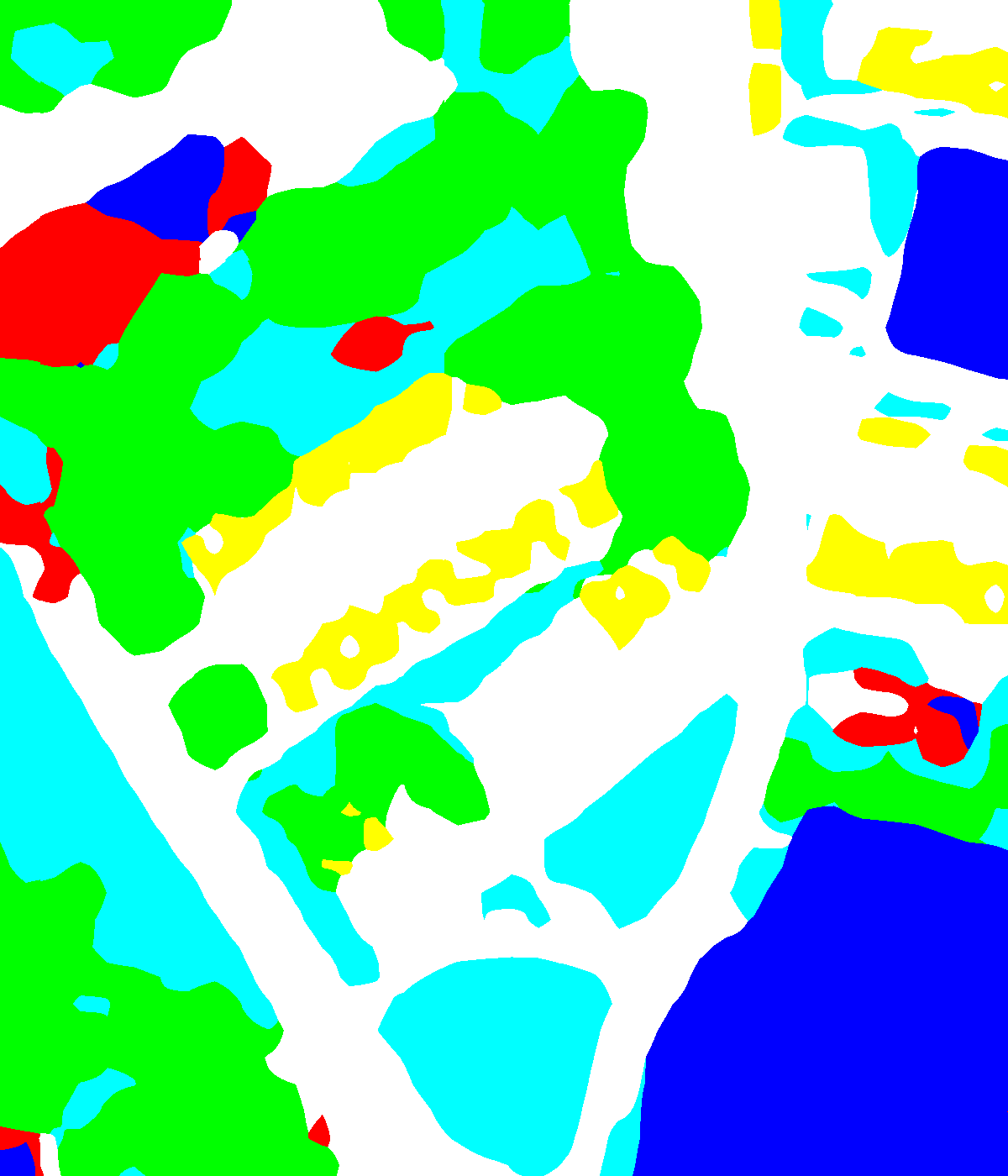} \hfill
  \includegraphics[width=0.24\textwidth]{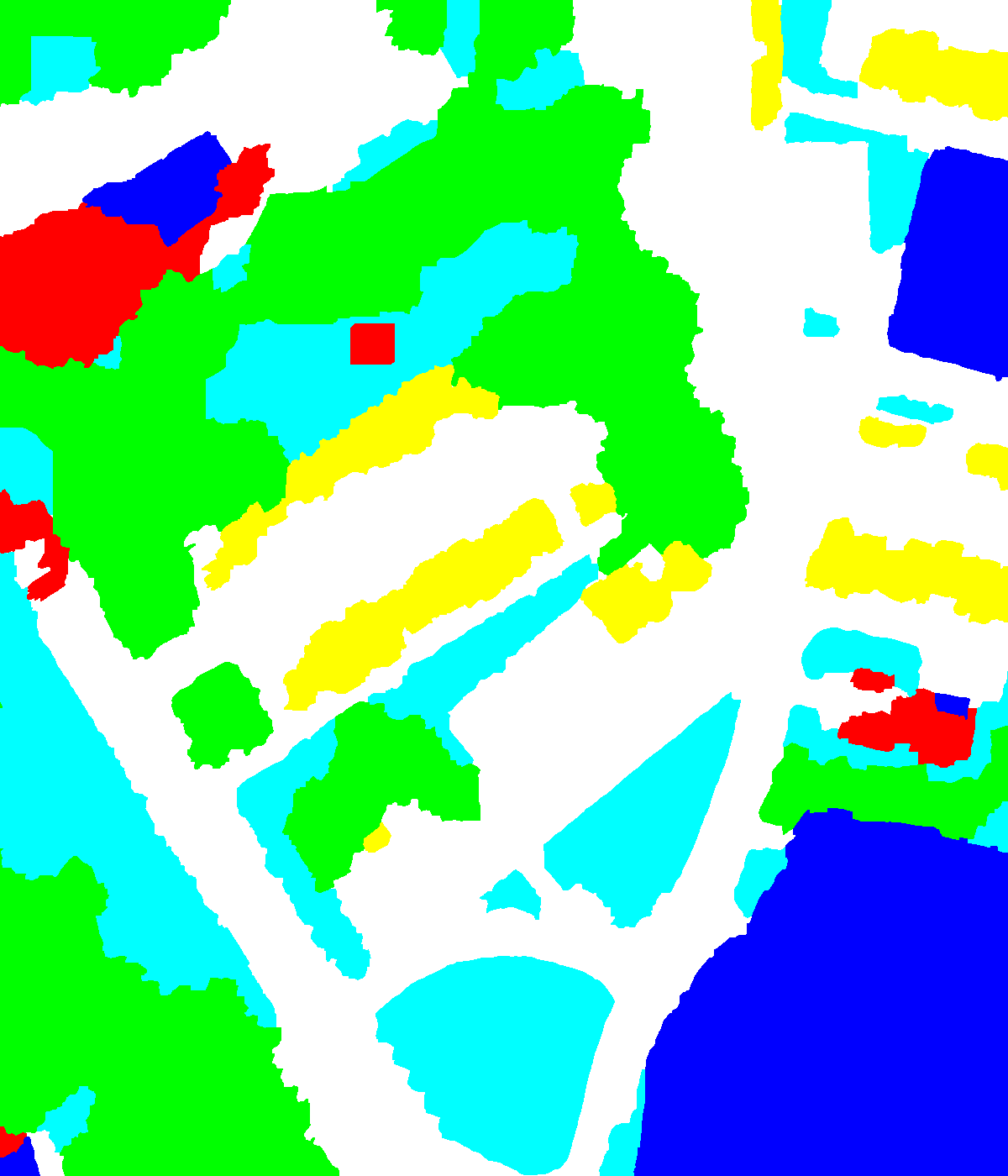} \hfill
  \includegraphics[width=0.24\textwidth]{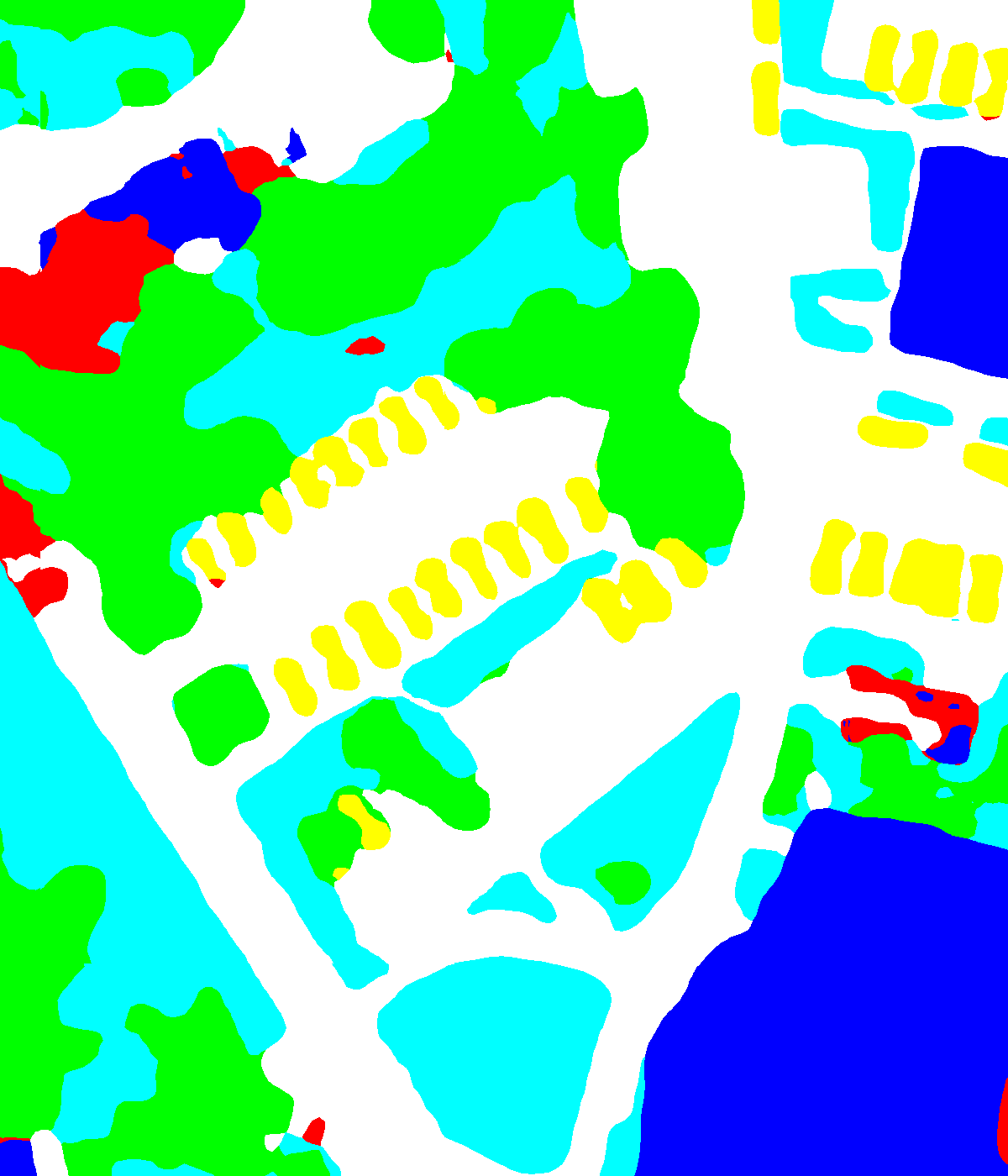} \hfill
  \includegraphics[width=0.24\textwidth]{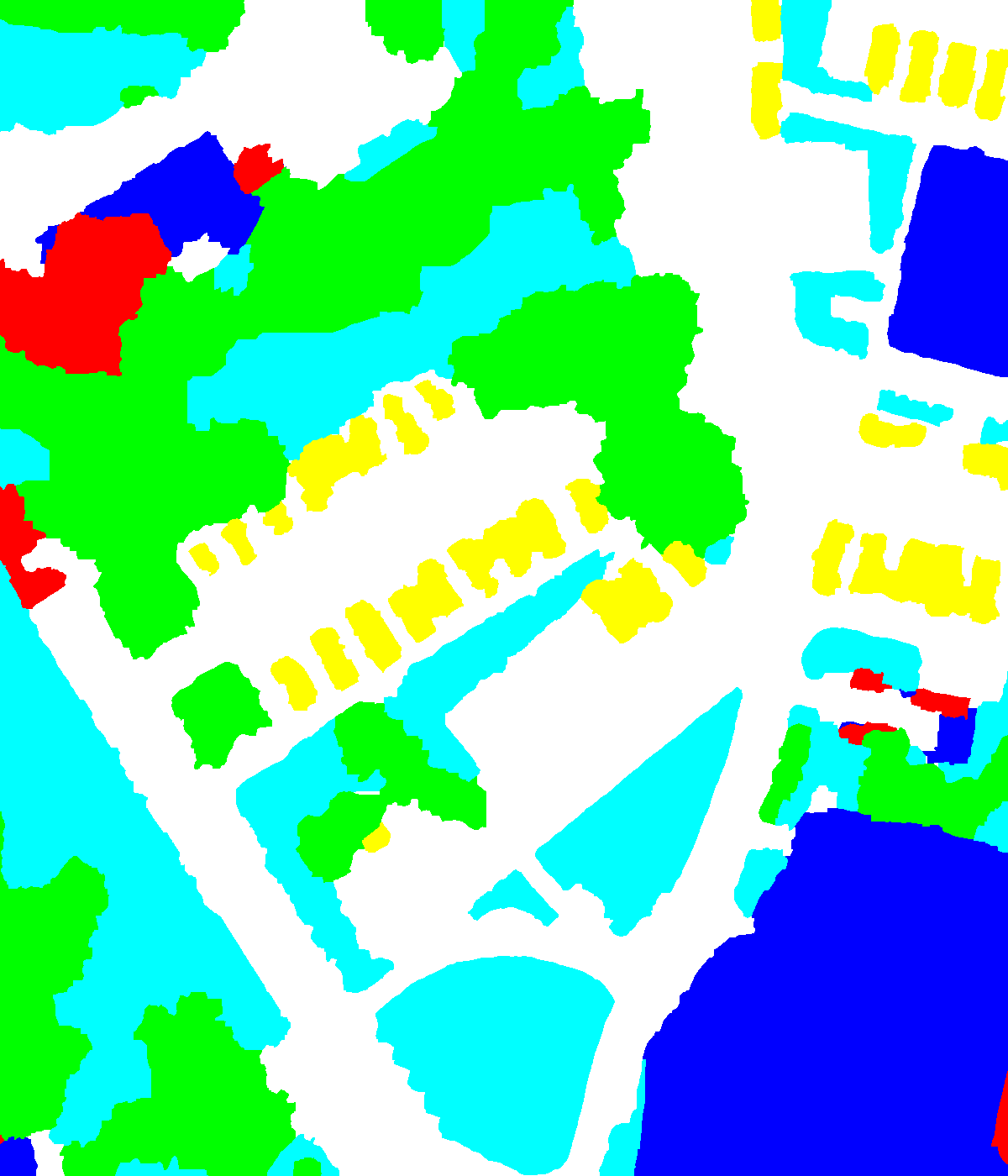} \hfill
  \caption{Results of pre-trained VGG FCN combined with DSM FCN, for Potsdam
    data set. From left to right: VGG on CIR + DSM FCN (DST\_3), DST\_3 + CRF
    (DST\_4), VGG on CIR + DSM FCN no-downsampling (DST\_5), DST\_5 + CRF
    (DST\_6) \label{fig:pPreTrainedDSMResults}}
\end{figure}

In~\cite{paisitkriangkraiEtAl:cvprw2015} it was shown that adding the DSM data
to the CNN input improved the accuracy significantly, by more than 3\% on the
Potsdam data set.  The DSM data is combined with pre-trained features computed
from CIR data using the hybrid architecture shown in
Figure~\ref{fig:hybridDiag}.  The results in row 3 of
Table~\ref{tab:preTrainedPotsdam} show only a small improvement over the
CIR-only result, indicating that late fusion does not make the most of the
height information.  Next a no-downsampling version of the hybrid network is
trained.  The network is much larger than the standard FCN used in other
experiments, and using a stride of 1 pixel in all layers exhausts the GPU's
memory.  Therefore the no-downsampling strategy is applied only partially,
resulting in a convolutional filter stride of 8 that is followed with bilinear
interpolation to produce the output.  In row 4 of
Table~\ref{tab:preTrainedPotsdam} the results show another small
improvement in accuracy due to the no-downsampling approach, but a significant
improvement for the car class (6.31\%).  Row 5 shows the result when
incorporating all five input channels in a hybrid network.  The results are
worse than CIR-only again indicating that late fusion is a sub-optimal approach.
The last three rows of the table show the previous approaches followed with a
CRF for non-linear smoothing of the labels.  In each case the CRF improves
accuracy.  The labelling outputs are exemplified in
Figure~\ref{fig:pPreTrainedDSMResults}.  Comparing columns 1 and 3, the
no-downsampling network does a better job of labelling the finely-detailed parts
of the scene.

The pre-trained and hybrid networks trained on the Potsdam data were applied to
the hold-out data set and the results submitted to the Potsdam labelling
challenge.  The leaderboard results are shown in
Table~\ref{tab:leaderboardPotsdam}.  These are the first externally-contributed
results to the Potsdam challenge (at the time of writing).  The FCN approach makes a significant
improvement over the baseline method, achieving a best accuracy of 90.3\%.

\begin{table*} [tb]
\scriptsize
  \centering
  \caption{ FCN semantic labelling results on Potsdam data set, fine-tuning of VGG-16 pre-trained model, validation set results.    The configurations marked ``DST\_X'' correspond to the leaderboard submissions, see Table~\ref{tab:leaderboardPotsdam} for the hold-out test set results. }
  \label{tab:preTrainedPotsdam}
  {
  \begin{tabular}{p{.31\textwidth}|c|c|c|c|c|c|c|c}
  \hline
   & \scriptsize{Imp.} & \scriptsize{Building} & \scriptsize{Low} & \scriptsize{Tree} & \scriptsize{Car} & \scriptsize{Unknown} & \scriptsize{Overall} & \scriptsize{Overall}\\
   &\scriptsize{surf.}  &                       & \scriptsize{ veg.} &&&& \scriptsize{ F1} & \scriptsize{ Acc.}\\
  \hline
  \hline  

VGG as FCN, RGB, pre-trained & $88.96\%$  & $92.49\%$  & $83.84\%$  & $82.11\%$  & $86.13\%$  & $73.09\%$  & $84.44\%$  & $86.05\%$ \\

VGG as FCN, CIR, pre-trained (DST\_1)  & $89.77\%$  & $92.97\%$  & $85.66\%$  & $83.55\%$  & $87.85\%$  & $74.27\%$  & $85.68\%$  & $87.28\%$ \\

VGG as FCN, CIR, pre-trained + DSM FCN rand. init. (DST\_3) & $89.84\%$  & $93.80\%$  & $85.43\%$  & $83.61\%$  & $88.00\%$  & $74.48\%$  & $85.86\%$  & $87.42\%$ \\

VGG as FCN, CIR, pre-trained + DSM FCN rand. init., no-DS (DST\_5)    & $89.95\%$  & $93.73\%$  & $85.91\%$  & $83.86\%$  & $94.31\%$  & $74.62\%$  & $87.06\%$  & $87.69\%$ \\

VGG as FCN, RGB, pre-trained + IR-DSM FCN rand. init. & $89.78\%$  & $93.85\%$  & $84.55\%$  & $82.40\%$  & $87.74\%$  & $74.09\%$  & $85.40\%$  & $86.92\%$ \\

DST\_1 + CRF   (DST\_2)   & $89.98\%$  & $93.05\%$  & $85.86\%$  & $83.21\%$  & $87.76\%$  & $75.28\%$  & $85.86\%$  & $87.43\%$ \\

DST\_3 + CRF   (DST\_4)   & $90.04\%$  & $93.92\%$  & $85.73\%$  & $83.35\%$  & $87.85\%$  & $75.59\%$  & $86.08\%$  & $87.63\%$ \\

DST\_5 + CRF (DST\_6) & $90.01\%$  & $93.83\%$  & $86.15\%$  & $83.59\%$  & $92.97\%$  & $75.87\%$  & $87.07\%$  & $\bf 87.84\%$ \\

  \hline
  \end{tabular}
  }
\end{table*}

\begin{table*} [tb]
  \centering
  \caption{ ISPRS Challenge Potsdam Leaderboard Results.  See Table~\ref{tab:preTrainedPotsdam} for configurations corresponding to DST\_x submissions. } 
  \label{tab:leaderboardPotsdam}
  {
  \begin{tabular}{l|c|c|c|c|c|c}
  \hline
   & \small{Imp. surf.} & \small{Building} & \small{Low veg.} & \small{Tree} & \small{Car} & \small{Overall Acc.}\\
  \hline
  \hline  
SVL\_1	& $83.5\%$	& $91.7\%$	& $72.2\%$	& $63.2\%$	& $62.2\%$	& $77.8\%$ \\
DST\_1  & $91.4\%$	& $95.3\%$	& $85.1\%$	& $87.3\%$	& $88.7\%$	& $89.1\%$ \\
DST\_2	& $91.8\%$	& $95.9\%$	& $86.3\%$	& $87.7\%$	& $89.2\%$	& $89.7\%$ \\
DST\_3	& $92.0\%$	& $96.2\%$	& $86.4\%$	& $88.3\%$	& $89.5\%$	& $90.1\%$ \\
DST\_4	& $92.0\%$	& $96.3\%$	& $86.5\%$	& $88.1\%$	& $89.5\%$	& $90.1\%$ \\
DST\_5	& $92.5\%$	& $96.4\%$	& $86.7\%$	& $88.0\%$	& $94.7\%$	& $\bf90.3\%$ \\
DST\_6	& $92.4\%$	& $96.4\%$	& $86.8\%$	& $87.7\%$	& $93.4\%$	& $90.2\%$ \\

  \hline
  \end{tabular}
  }
\end{table*}

These experiments focus on the Potsdam data due to its higher GSD.  Pre-trained
features were found to not improve the labelling of the Vaihingen data compared
with the standard FCN.  The Vaihingen data is lower resolution than the Potsdam
data, and does not benefit as much from the textural features.  In this case
early fusion of the DSM information is more beneficial than the addition of
pre-trained features.

\section{Conclusion}\label{sec:conclusion}

In this paper the benefits of fully-convolutional networks for semantic
labelling have been shown to extend to high-resolution aerial imagery.  A novel
no-downsampling approach to FCNs was introduced that preserves the full input
image resolution at every layer.  Training with no downsampling takes
considerable computational resources, however for detailed high-resolution
imagery the labelling accuracy is significantly improved.  The extra
computational cost of no-downsampling at test time is modest ($\approx 4
\times$) when compared with the equivalent shift-and-stitch approach to
interpolation ($\approx 20-70 \times$).  In~\cite{longEtAl:CVPR2015} the
benefits of shift-and-stitch were found to not outweigh the computational cost;
perhaps by using a no-downsampling network a different conclusion would be
reached.  Although the no-downsampling network produces dense outputs, the
spatial feature resolution still decreases with layer depth.  To improve the
detail in the labelling results this work should be extended to combine features
from multiple scales, see for example~\cite{linEtAl:cvpr2016}.

The same pre-trained CNN features that have been applied successfully to so many
other computer vision problems were also found to benefit semantic labelling of
the higher-resolution Potsdam aerial imagery.  However it is not straightforward
to incorporate extra bands of information, namely the DSM.  A hybrid network was
proposed to combine a pre-trained FCN with a DSM-only FCN trained from scratch
using late fusion in the fully-connected layers.  Whilst this did improve
labelling accuracy somewhat for the higher-resolution Potsdam data, input-level
fusion of the DSM without pre-trained features gave a bigger boost to the
Vaihingen data set results.  Determining how pre-trained features can be fused
at a lower level would be a beneficial topic of future work.

The proposed semantic labelling methods have been applied to two
publicly-available benchmark data sets and, at the time of writing, achieved
state-of-the-art accuracy.  In order for this method to work in practice, the
key issue is generalisation - extending the labelling to a wider variety of
scenes.  To achieve this a much larger labelled training set is needed.  The
labels provided with the challenge data were painstakingly created by hand, and
repeating this effort across the globe is not feasible.  Automatic generation of
labels or pseudo-labels, for instance from OpenStreetmap, is the most promising
way forward.  Learning would need to take into account noise in these
less-accurate labels~\cite{Minh2013Machine}.  Another factor important for
generalisation is extension to different input bands, in particular the DSM
might not be available in many scenarios.  This paper has shown that high
accuracy results can be achieved without elevation data, relying on image
features only (89.7\% on Potsdam).  With improved generalisation and a more
complete set of class labels, semantic labelling of aerial imagery would provide
a semantic image layer that would benefit a great many geospatial application.

\section{Acknowledgements}

Thanks to Robert Christie for his software development work.  Thanks to Markus
Gerke and the other organisers of the challenge for the normalised DSMs and
support relating to submissions.  Thanks to the ISPRS for providing the research
community with the wonderful challenge data sets.

\bibliographystyle{plain}
\bibliography{sherrah_fcn_overhead_arxiv,draft}

\begin{appendices}

\section{Experimental Setup}\label{apdx:experiments}

In all experiments the FCN filter support is 64x64 pixels for the Vaihingen
data, 128x128 pixels for Potsdam, and 224x224 pixels in the case of pre-trained
networks.  For 64x64 input, non-overlapping tiles of size 128x128 pixels are
cropped out of the input images and ground truth labels.  It was found that
larger tiles give equivalent accuracy. In all other cases the tile size is
256x256 pixels.  In general a batch size of 2 tiles is used, with larger batch
sizes resulting in poor convergence and over-fitting.  For pre-trained nets a
larger batch size is beneficial, but due to GPU RAM limitations 6 tiles are
used per batch.  The input data is augmented by flipping and rotating each
input image before tiling.  Images are rotated in 10 degree increments,
resulting in 72 augmentations per image (36 angles by 2 flips). Only the largest
rectangle fitting inside the valid area of the rotated image is used for
training.  With augmentations the Vaihingen data set has 123,330 tiles and
Potsdam has 417,114 tiles.

Unless otherwise stated, the input image for Vaihingen has 4 layers (infra-red,
red, green, normalised DSM) and for Potsdam 5 layers
(red,green,blue,infra-red,normalised DSM).  Each input channel has the training
sample mean subtracted.  The following standard parameters are used when
training the networks: learning rate = 0.001, momentum = 0.9, weight decay =
0.0005, initial weights drawn from a Normal distribution with standard deviation =
0.01.  In drop-out layers the ratio is 0.5.  Networks are trained with 150,000 %
iterations (weight updates).  The learning rate is decreased by a factor of 10
two-thirds of the way through training.

A modified version of Caffe~\cite{jia2014caffe} is used for all experiments, in
conjunction with NVIDIA Titan X and Tesla K80 GPUs.  Due to the large size of
overhead images, the amount of RAM on the GPU is a limiting factor.  Each of the
graphics cards used has 12 GB of RAM.

Trained FCNs are applied to the large overhead images by tiling the image,
applying the FCN to each tile, and assembling the tiled results back into a
full-sized image.  To avoid boundary artefacts the tiles are overlapped by half
of the FCN input size.  These tiles are generally 512x512 pixels but sometimes
have to be made smaller to fit into GPU RAM, depending on the network.  Whenever
the output resolution is lower than the input resolution, the class
probabilities are bilinearly interpolated to the full resolution and the class
with the highest probability used as the output label.

The protocol for accuracy assessment on the Vaihingen and Potsdam data sets is
described on the challenge website~\cite{ISPRS}.  Overall accuracy -- the percentage
of correctly labelled pixels -- can be considered the main accuracy measure.  F1
scores for each of the classes are also presented.  For the Vaihingen data set
our classifiers do not produce the ``unknown'' class due to its scarcity, and
the unknown class is not included in validation set metrics.  The unknown class
is included fully for the Potsdam data set.  A three-pixel boundary between
ground truth regions with different labels is ignored in the accuracy assessment
to allow for human error in the ground truth.

In some experiments a pixel-wise 4-connected conditional random field (CRF) is
used to non-linearly smooth the output labels.  The CNN output probabilities are
used as unary potentials and the binary potentials are based on Canny edges
derived from the input data.  The method is described in more detail
in~\cite{paisitkriangkraiEtAl:cvprw2015}.  In the case of the Potsdam data, the
binary edge mask is the combination of CIR, RGB and NDVI edges.

\end{appendices}

\end{document}